\documentclass{article}

    \PassOptionsToPackage{numbers, compress}{natbib}

    \usepackage[final]{neurips_2024}

\usepackage[utf8]{inputenc} 
\usepackage[T1]{fontenc}    
\usepackage{hyperref}       
\usepackage{url}            
\usepackage{booktabs}       
\usepackage{amsfonts}       
\usepackage{nicefrac}       
\usepackage{microtype}      
\usepackage{xcolor}         

\usepackage{eccvabbrv}
\usepackage{subcaption}
\usepackage{amsmath}
\usepackage{multirow}
\usepackage{utfsym}
\usepackage{csquotes}
\usepackage{tabularx}
\usepackage{verbatim} 
\usepackage{rotating}
\usepackage{adjustbox}
\usepackage{dashrule}
\usepackage{makecell}
\usepackage{color, colortbl}
\usepackage{algorithm}
\usepackage{algpseudocode}
\usepackage{tcolorbox}

\title{Incorporating Test-Time Optimization into Training with Dual Networks for Human Mesh Recovery}

\author{
    Yongwei Nie$^{1}$, Mingxian Fan$^{1}$, Chengjiang Long$^{2}$, Qing Zhang$^{3}$, Jian Zhu$^{4}$,
    Xuemiao Xu$^{1}$\thanks{Corresponding author.} \\
    $^{1}$South China University of Technology, China \\
    $^{2}$Meta Reality Labs, USA \\
    $^{3}$Sun Yat-sen University, China \\
    $^{4}$Guangdong University of Technology, China \\
    \texttt{\{nieyongwei, xuemx\}@scut.edu.cn, fanmingxian123@gmail.com} \\
    \texttt{cjfykx@gmail.com, zhangq93@mail.sysu.edu.cn, rockeyzhu@163.com}
}

\begin{document}

\maketitle

\begin{abstract}
Human Mesh Recovery (HMR) is the task of estimating a parameterized 3D human mesh from an image. There is a kind of methods first training a regression model for this problem, then further optimizing the pretrained regression model for any specific sample individually at test time. However, the pretrained model may not provide an ideal optimization starting point for the test-time optimization. Inspired by meta-learning, we incorporate the test-time optimization into training, performing a step of test-time optimization for each sample in the training batch before really conducting the training optimization over all the training samples. In this way, we obtain a meta-model, the meta-parameter of which is friendly to the test-time optimization. At test time, after several test-time optimization steps starting from the meta-parameter, we obtain much higher HMR accuracy than the test-time optimization starting from the simply pretrained regression model. Furthermore, we find test-time HMR objectives are different from training-time objectives, which reduces the effectiveness of the learning of the meta-model. To solve this problem, we propose a dual-network architecture that unifies the training-time and test-time objectives. Our method, armed with meta-learning and the dual networks, outperforms state-of-the-art regression-based and optimization-based HMR approaches, as validated by the extensive experiments. The codes are available at \href{https://github.com/fmx789/Meta-HMR}{https://github.com/fmx789/Meta-HMR}.

\end{abstract}

\section{Introduction}

Human mesh recovery (HMR) from a single image is of great importance to human-related applications, such as action capture without MoCap device, action transfer with vision-based system, and VR/AR entertainments, etc. This topic has received extensive research during past years, for which most of previous approaches represent a 3D human mesh by the parametric human model SMPL \cite{loper2015smpl} with parameters $\Theta=(\theta,\beta)$, where $\theta$ encodes the pose of the mesh and $\beta$ describes the body shape. The aim is thus to estimate $\Theta$ of a human in a given image.

Originally, the problem is solved by optimizing a standard human mesh so that its 2D projection matches the 2D joints of the target human (e.g., SMPLify \cite{bogo2016keep}). Later, works of \cite{kanazawa2018end,li2022cliff, wang2023zolly, lin2021end, cho2022cross,yoshiyasu2023deformable,zheng2023potter,xue20223d} propose end-to-end networks trained on large datasets to directly output a 3D SMPL mesh given an input image. In SPIN~\cite{kolotouros2019learning}, the regression-based approach \cite{kanazawa2018end} and optimization-based approach SMPLify~\cite{bogo2016keep} are combined together by interleaved training, where the regression method provides an initial solution for optimization and then the optimization method provides supervision for regression. Different from SMPLify \cite{bogo2016keep} that directly optimizes a 3D mesh, works of EFT \cite{joo2021exemplar} and BOA~\cite{guan2021bilevel} finetune a pretrained regression model on each test sample, which indirectly optimize the target human mesh (i.e., the the outcome of the regression models) . 



In this paper, we are particularly interested in the test-time optimization-based approaches of EFT \cite{joo2021exemplar} and BOA \cite{guan2021bilevel}. Since BOA is developed for video leveraging consistency properties between frames, we mainly discuss and compare with EFT that performs HMR for images, like ours. Our observation is that the test-time optimization is analogous to one-shot learning. That is, it finetunes a pretrained model on a specific sample before really applying the model to solve the human mesh recovery task for that sample. However, the pretrained model, which is not specially tailored for the one-shot learning problem, may not be so effective for the test-time adaptation as desired. 

Based on the above analysis and inspired by Kim et al. \cite{kim2022meta}, we incorporate the test-time optimization into the training process, re-formulating the test-time optimization from the perspective of learning to learn, i.e., meta learning \cite{finn2017model}.
Specifically, given a batch of training samples (or saying a set of tasks), our method first performs test-time optimization on each sample to update the regression network parameters temporally for that sample. Then, based on all pieces of regression parameters after test-time optimizations, we further optimize the training-time objectives over the whole batch of training samples. By performing training-time optimization after test-time optimization, we imagine that the training optimization works as a faithful supervision to correct the wrong optimization directions of the test-time optimizations. After the training, the obtained parameters of the regression network can be viewed as meta-parameters which will be instantiated to parameters actually used for human mesh recovery through several test-time optimization steps.

We find that the test-time objectives for the human mesh recovery task are different from training-time objectives, because we sometimes have ground-truth human meshes at training time but forever not at test time. This may produce an obstacle in the meta-learning process, since the optimization directions of the test-time and training optimizations are not identical. To alleviate this problem, we design a dual-network structure to implement our method, which owns a main regression network and an auxiliary network. The auxiliary network provides the main network with pseudo ground-truth SMPL meshes, by which we unify the training and test-time objectives elegantly. 


We demonstrate through extensive experiments that our method equipped with meta-learning and the dual networks greatly outperforms state-of-the art approaches. To summarize, our main contributions are three-fold: (1) We propose a novel dual-network HMR framework with test-time optimization involved into the training procedure, which improves the effectiveness of the test-time optimizations. (2) We ensure the test-time objectives identical to the training objectives, further facilitating the joint-training of the test-time and training-time optimizations. (3) Extensive experiments validate that our results outperform those of previous approaches both quantitatively and qualitatively.

\section{Related Work}

\textbf{Regression-based HMR methods} typically employ neural networks to regress the human body mesh representation from images. Methods of \cite{kocabas2021pare, pavlakos2018learning, omran2018neural, zhang2021pymaf, kanazawa2018end, moon2022accurate, xu20203d, zanfir2020weakly, tan2017indirect, zhang2023pymaf, fang2023learning, lin2024mpt, xuan2024mh, liao2024instahmr, liao2024progressive} choose to  regress parametric human body model, i.e., SMPL \cite{loper2015smpl}. HMR \cite{kanazawa2018end} was the first employing CNN \cite{he2016identity} to extract features and MLP layers to output 3D mesh parameters. Later, sophisticated networks were proposed for improving the reasoning accuracy. For example, PyMAF \cite{zhang2021pymaf,zhang2023pymaf} extracted features in a pyramid structure and iteratively aligned 3D vertices with human body in the image. Xue et al. \cite{xue20223d} used a learnable mask to automatically identify the most discriminative features related to 3D mesh recovery. Works of \cite{kocabas2021spec, li2022cliff, wang2023zolly, nie2024multiple} observed that prior works overlooked the importance of camera parameters. Among them, CLIFF \cite{li2022cliff} innovatively considered using the cropping bounding boxes as input to reduce the ambiguity of reprojection loss. Zolly \cite{wang2023zolly} considered the camera distortion produced by perspective projection. Recently, Nie et al. \cite{nie2024multiple} proposed a RoI-aware feature extraction and fusion network, guided by camera consistency and contrastive loss functions tailored to the multi-RoI setting.

There are also non-parameterized methods directly regressing mesh vertices. For example, METRO \cite{lin2021end} utilized Transformer to model the global relationship between human keypoints and mesh vertices. 
FastMETRO \cite{cho2022cross} separated backbone features from the features corresponding to keypoints and vertices. Recently, \cite{yoshiyasu2023deformable} combined pyramid structure in PyMAF \cite{zhang2021pymaf,zhang2023pymaf} with the Transformer-based HMR regression method, further improving the regression accuracy.

The regression model is a key component in our method. In this paper, We test HMR \cite{kanazawa2018end} and CLIFF \cite{li2022cliff} as the regression network in our method. 

\textbf{Optimization-based HMR methods} \cite{bogo2016keep, pavlakos2019expressive, lassner2017unite, guan2009estimating, hasler2010multilinear, zhou2010parametric, bualan2008naked, joo2021exemplar, sigal2007combined} usually attempt to estimate a 3D body mesh consistent with 2D image cues. Bogo et al. \cite{bogo2016keep} proposed an approach called SMPLify, which iteratively adjusts SMPL parameters to fit detected 2D keypoints. SPIN \cite{kolotouros2019learning} combined regression-based methods with SMPLify in an interleaved training strategy. 
CycleAdapt \cite{nam2023cyclic} alternately trained a HMR network and a motion denoising network to enhance each other. Unlike SMPLify, EFT \cite{joo2021exemplar} and BOA \cite{guan2021bilevel} fine-tuned a pretrained regression network via 2D reprojection loss or temporal consistency loss at test phase, updating the SMPL parameters indirectly. 
Some approaches proposed learning stronger 3D priors \cite{kolotouros2021probabilistic, pavlakos2019expressive, davydov2022adversarial, nam2023cyclic} or utilizing trainable neural networks to update parameters in lieu of gradient updates \cite{zanfir2021neural, choutas2022learning, song2020human}. Inverse kinematics (IK) has also been explored. These methods address IK problems by decomposing relative rotations \cite{li2021hybrik}, designing networks that integrate forward and inverse kinematics \cite{li2023niki}, or incorporating UV position maps \cite{shetty2023pliks}. Different from all the above, our method integrates test-time optimization into the training process, obtaining a meta-model and meta-parameters. 

\textbf{Meta Learning} Our method is most related to the model-agnostic meta-learning (MAML, or more precisely FOMAML) \cite{finn2017model}. MAML first samples a number of tasks, then performs local optimization on each task, and finally conducts a global optimization to update the parameters of the original network. Similarly, our method first executes test-time optimization on each training sample and then performs training optimization on a batch of training samples. Although our method is inspired by MAML and its extensions \cite{antoniou2018how,Raghu2020Rapid,rajeswaran2019meta}, our goal is fundamentally different from theirs. The goal of MAML is usually for few-shot learning or domain adaptation, which assumes there is ground-truth labeled data in the target domain. In contrast, our goal is to adapt the network to a single test sample which is free of ground-truth human mesh. We have noted MAML has been proven effective in various domains, such as talking head generation \cite{zhang2023metaportrait}, SVBRDF recovery \cite{zhou2022look, fischer2022metappearance}, image super resolution \cite{park2020fast}, etc. As far as we know, the work of Kim et al. \cite{kim2022meta} is the first that applies meta learning to HMR. The key difference is that that Kim et al. \cite{kim2022meta} used the 2D reprojection loss only (please see Eq. 1 in their paper) in both inner and outer loops of meta learning, while we incorporate the ground-truth 3D SMPLs into the outer loop of meta learning and additionally generate pseudo SMPLs and incorporate them into the inner loop of the metal learning. The utilization of the GT and pseudo SMPLs greatly improves the results of our method upon the method of \cite{kim2022meta}.

\begin{figure*}[t]
    \centering
    \includegraphics[width=1.0\textwidth]{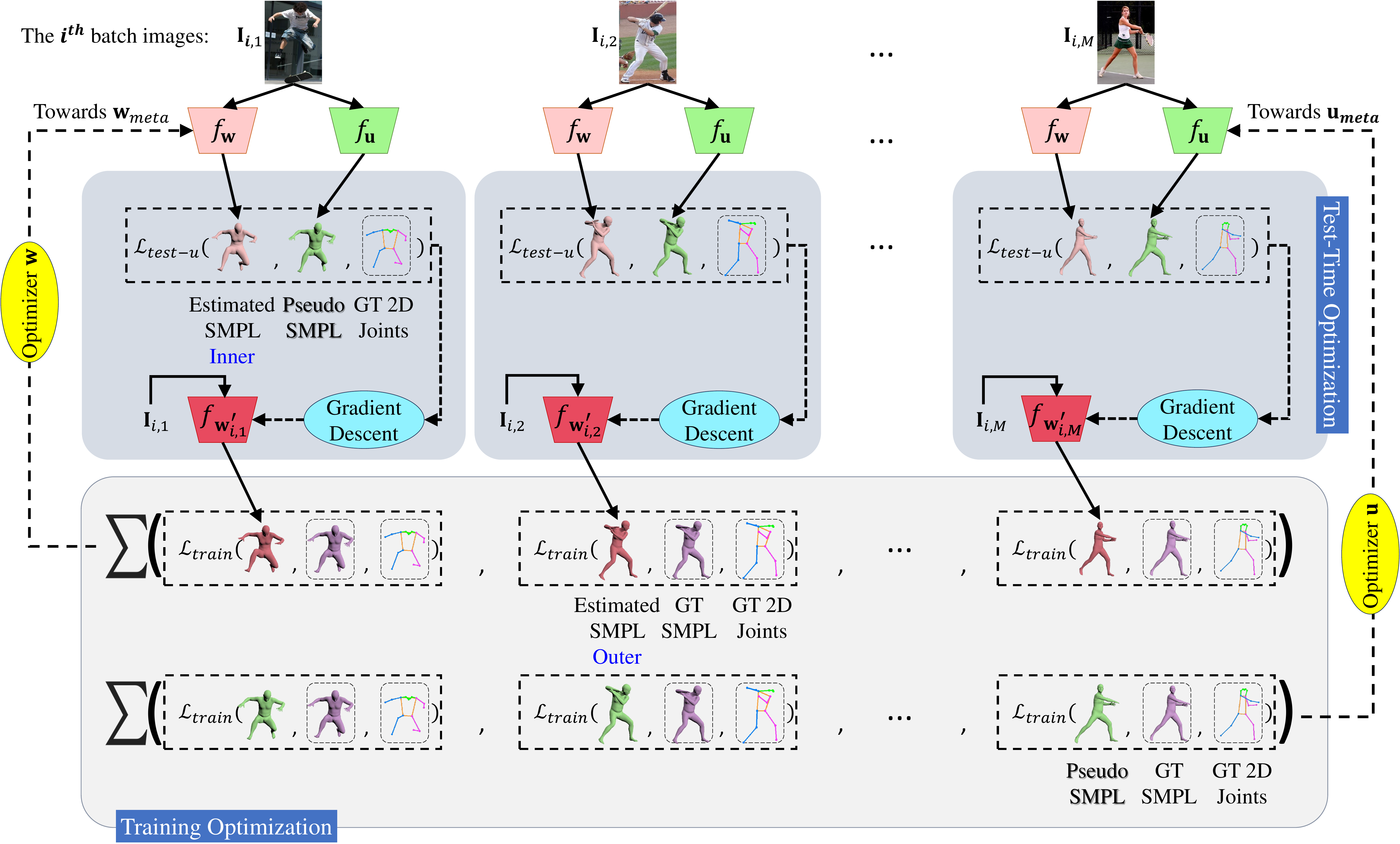}
    \caption{\textbf{Overview of the dual-network meta-learning HMR method,} composed of a main HMR regression network $f_{\mathbf{w}}$ and an auxiliary network $f_{\mathbf{u}}$. Both networks have the same architecture but different parameters. Given $i^{th}$ batch of images, test-time optimization is first executed for each training image $\mathbf{I}_{i,j}$ in the batch individually, updating $f_{\mathbf{w}}$ to $f_{\mathbf{w}'_{i,j}}$ by performing a gradient descent step w.r.t. the test-time loss function $\mathcal{L}_{test-u}$. Then based on $\{f_{\mathbf{w}'_{i,j}}|j\in[1,M]\}$ ($M$ is the batch size), the training optimization is executed to update the parameters of both main and auxiliary networks by $\mathcal{L}_{train}$ with different arguments respectively. $\mathbf{w}_{meta}$ and $\mathbf{u}_{meta}$ are the finally generated meta-parameters. $f_{\mathbf{u}}$ generates ``Pseudo SMPLs'' that are used in the test-time loss to supervise the learning of the ``Estimated SMPL \textcolor{black}{Inner}''. GT SMPLs are used in the training loss to supervise the learning of ``Estimated SMPL \textcolor{black}{Outer}'' and the Pseudo SMPLs.}
    \label{fig:overview-train}
\end{figure*}

\section{Our Method}
\label{sec:Our Method}
Our goal is to estimate a 3D SMPL human mesh parameterized by ${\Theta}=(\theta,\beta)$ together with a camera $\pi$ from a given image $\mathbf{I}$ of a person, where $\theta\in \mathbb{R}^{24\times3}$ and $\beta\in \mathbb{R}^{10}$ are pose and shape parameters of the SMPL human model \cite{loper2015smpl}, respectively. Let $\{\mathbf{I}_{i,j}, \hat{\Theta}_{i,j}, \hat{\mathbf{J}}_{i,j}\}_{i=1,j=1}^{B,M}$ be a training dataset, where $\mathbf{I}_{i,j}$ is a training image, $\hat{\Theta}_{i,j}$ is the ground-truth (GT) human mesh, $ \hat{\mathbf{J}}_{i,j}$ is the 2D GT joints (or joints detected by such as OpenPose \cite{cao2017realtime}) of the human in the input image, $B$ is the number of batches, and $M$ is the batchsize. End-to-end HMR regression approaches usually train a neural network $f_\mathbf{w}: \mathbf{I}_{i,j} \rightarrow (\Theta_{i,j},\pi_{i,j})$ by minimizing the following training-time loss function $\mathcal{L}_{train}$:
\begin{equation}
    \mathbf{w}_{pre} = \mathop{\arg\min}_{\mathbf{w}}\sum_{i=1}^B\sum_{j=1}^M \mathcal{L}_{train}(f_{\mathbf{w}}( \mathbf{I}_{i,j}),\hat{\Theta}_{i,j}, \hat{\mathbf{J}}_{i,j}),
    \label{eq:regression-optimization}
\end{equation}
where 
\begin{equation}
    \mathcal{L}_{train}(f_{\mathbf{w}}( \mathbf{I}_{i,j}),\hat{\Theta}_{i,j}, \hat{\mathbf{J}}_{i,j}) = \mathcal{L}_{2D}(f_{\mathbf{w}}( \mathbf{I}_{i,j}), \hat{\mathbf{J}}_{i,j}) + \mathcal{L}_{3D}(f_{\mathbf{w}}( \mathbf{I}_{i,j}),\hat{\Theta}_{i,j}),
    \label{eq:l-train}
\end{equation}
with 
\begin{equation}
  \mathcal{L}_{2D}(f_{\mathbf{w}}( \mathbf{I}_{i,j}), \hat{\mathbf{J}}_{i,j}) = \|\pi(\Theta_{i,j})-\hat{\mathbf{J}}_{i,j}\|_2^2, \;\;\; \mathcal{L}_{3D}(f_{\mathbf{w}}( \mathbf{I}_{i,j}),\hat{\Theta}_{i,j})= \|X(\Theta_{i,j}) - X(\hat{\Theta}_{i,j})\|_2^2. \\
\end{equation}
As seen, $\mathcal{L}_{train}$ is composed of a 2D reprojection loss $\mathcal{L}_{2D}$ and a 3D loss $\mathcal{L}_{3D}$. The 2D loss first projects mesh $\Theta_{i,j}$ to the 2D plane by the camera $\pi$ and then computes difference between the projected 2D joints and the given 2D joints $\hat{\mathbf{J}}_{i,j}$. The 3D loss computes difference between 3D meshes, where $X$ can be an identity transformation or transformations computing 3D human joints or mesh vertices from the mesh parameters. 



\subsection{Test-time Optimization}
Exemplar-Fine-Tuning (EFT) \cite{joo2021exemplar} was the first work proposing test-time optimization for HMR. 
In particular, the pretrained network $\mathbf{w}_{pre}$ is further finetuned by performing the following test-time optimization loss function on a specific test sample $\mathbf{I}_{i,j}$:
\begin{equation}
    \mathbf{w}^{*}_{i,j} = \mathop{\arg\min}_{\mathbf{w}_{i,j}} \mathcal{L}_{test} (f_{\mathbf{w}_{i,j}}(\mathbf{I}_{i,j}), \hat{\mathbf{J}}_{i,j} ), {\rm \;initially \; \; } \mathbf{w}_{i,j}=\mathbf{w}_{pre},
    \label{eq:exemplar-optimization}
\end{equation}
with 
\begin{equation}
    \mathcal{L}_{test} (f_{\mathbf{w}}(\mathbf{I}_{i,j}), \hat{\mathbf{J}}_{i,j}) = \mathcal{L}_{2D}(f_{\mathbf{w}}( \mathbf{I}_{i,j}), \hat{\mathbf{J}}_{i,j}).
    \label{eq:l-test}
\end{equation}
The test-time optimization starts from the initial solution $\mathbf{w}_{pre}$ provided by the pretrained model. It resembles one-shot learning but actually does not, because the pretrained model is obtained using normal supervised learning techniques while not introducing any strategy for guaranteeing the properties of one-shot learning. The parameters of the pretrained model may be not ideal as the starting point for the test-time optimization. 


\begin{algorithm}
\caption{Meta learning of Dual networks for 3D Human Recovery}
\begin{algorithmic}[1]
\Statex \textbf{\Large \textbullet} \textbf{The stage of training}
\Require Training dataset $\{\mathbf{I}_{i,j}, \hat{\Theta}_{i,j}, \hat{\mathbf{J}}_{i,j}\}_{i=1,j=1}^{B,M}$
\Require $f_{\mathbf{w}}, f_{\mathbf{u}}$: main and auxiliary networks, randomly initialized
\Require $\alpha, \beta$: step size hyperparameters
\While{not done}
    \State Sample a batch of images $\{\mathbf{I}_{i,j}\mid j \in [1, M]\}$, $i \sim \mathcal{U}(1,B)$ \Comment{$\mathcal{U}$ is uniform distribution}
    \For{all $\mathbf{I}_{i,j}$}
        \State Compute SMPL meshes by $f_{\mathbf{w}}(\mathbf{I}_{i,j})$ and $f_{\mathbf{u}}(\mathbf{I}_{i,j})$, respectively
        \State Compute $L_{\text{test-u}}$ in Eq. \ref{eq:one-step-update2}, evaluate $\nabla_{\mathbf{w}}L_{\text{test-u}}$, and update $\mathbf{w}_{i,j}' \leftarrow \mathbf{w} - \alpha \nabla_{\mathbf{w}} L_{\text{test-u}}$
    \EndFor
    \State Compute $L_{\text{train}}^1$ in Eq. \ref{eq:combined}, evaluate $\nabla_{\mathbf{w}}L_{\text{train}}^1$, and update $\mathbf{w} \leftarrow \mathbf{w} - \beta \sum_j \nabla_{\mathbf{w}_{i,j}} L_{\text{train}}^1$
    
    \Comment{$\mathbf{w}_{i,j}'$ is used in $L_{\text{train}}^1$}
    \State Compute $L_{\text{train}}^2$ in Eq. \ref{eq:combined}, evaluate $\nabla_{\mathbf{u}}L_{\text{train}}^2$, and update $\mathbf{u} \leftarrow \mathbf{u} - \beta \sum_j \nabla_{\mathbf{u}} L_{\text{train}}^2$
\EndWhile
\State $\mathbf{w}_{\text{meta}} \leftarrow \mathbf{w}, \;\; \mathbf{u}_{\text{meta}} \leftarrow \mathbf{u}$
\Statex \setcounter{ALG@line}{0} 
\Statex \textbf{\Large \textbullet} \textbf{The stage of testing}
\Require Input image $\mathbf{I}$
\Require Main and auxiliary networks with meta parameters $\mathbf{w}_{\text{meta}}$ and $\mathbf{u}_{\text{meta}}$
\State $\mathbf{w} = \mathbf{w}_{\text{meta}}$
\For{$i = 1$ to $m$}  \Comment{Iterate test-time optimization $m$ times}
    \State Pseudo GT mesh $\gets f_{\mathbf{u}_{\text{meta}}}(\mathbf{I})$  \Comment{Using frozen auxiliary network}
    \State Compute $L_{\text{test-u}}$ using Eq. \ref{eq:one-step-update2} \Comment{Given input $\mathbf{I}$, the pseudo GT mesh, and 2D joints}
    \State Evaluate $\nabla_{\mathbf{w}}L_{\text{test-u}}$ 
    \State Update $\mathbf{w} \leftarrow \mathbf{w} - \alpha \nabla_{\mathbf{w}} L_{\text{test-u}}$
\EndFor
\State $\mathbf{w}_{\text{final}} \leftarrow \mathbf{w}$
\State Compute SMPL mesh by $f_{\mathbf{w}_{\text{final}}}(\mathbf{I})$
\end{algorithmic}
\label{alg:1}
\end{algorithm}

\subsection{Incorporating Test-time Optimization into Training}

To solve the above problem, we propose to integrate the test-time optimization into the training procedure as shown in Figure~\ref{fig:overview-train}, inspired by optimization-based meta-learning \cite{finn2017model}. 
Specifically, for each sample in a batch, we first perform test-time optimization on that sample to update the parameters of the regression network temporally corresponding to the sample. After that, we perform training optimization over all $M$ training samples, based on the $M$ temporally updated regression networks. This process is formulated as:
\begin{equation}
    \mathbf{w}_{meta} = \mathop{\arg\min}_{\mathbf{w}} \sum_{i=1}^{B}\sum_{j=1}^{M} \mathcal{L}_{train}( f_{\mathbf{w}'_{ij}}(\mathbf{I}_{ij}), \hat{\Theta}_{ij}, \hat{\mathbf{J}}_{ij}),
    \label{eq:our-objective}
\end{equation}
where,
\begin{equation}
    \mathbf{w}'_{ij} = \mathbf{w} - \alpha \nabla_{\mathbf{w}}\mathcal{L}_{test}( f_{\mathbf{w}}(\mathbf{I}_{ij}), \hat{\mathbf{J}}_{ij} ).
    \label{eq:one-step-update}
\end{equation}

The difference between Eq.~\ref{eq:our-objective} and Eq.~\ref{eq:regression-optimization} is in the parameters of $f$ to be optimized. Instead of directly optimizing the current parameters $\mathbf{w}$ of $f$ using the training objective $\mathcal{L}_{train}$, we first perform a step of test-time optimization using Eq.~\ref{eq:one-step-update} on each sample $\mathbf{I}_{ij}$ to obtain network parameters $\mathbf{w}'_{ij}$ specific to that sample. Then, $\{\mathbf{w}'_{ij}|j\in[1,M]\}$ over all training samples in a batch are in turn used in Eq.~\ref{eq:our-objective} to evaluate the training objective. In Eq.~\ref{eq:one-step-update}, $\alpha$ is the learning rate of the test-time optimization.


By performing test-time optimization before training optimization in Eq.~\ref{eq:our-objective} and~\ref{eq:one-step-update}, we take test-time optimization into consideration in the training procedure. That means, the ``test-time optimization'' is trained on the training dataset, thus having better generalization ability to test samples. The proposed method resembles the optimization-based meta-learning \cite{finn2017model} and we call the obtained parameters $\mathbf{w}_{meta}$ meta-parameters.

\subsection{Unifying Training and Test-time Optimization Objectives with Dual Networks}

There are ground-truth human meshes at training time while not at test time, causing the difference between $\mathcal{L}_{train}$ (see Eq. \ref{eq:l-train})) and $\mathcal{L}_{test}$ (see Eq. \ref{eq:l-test}). Since both the test-time and training optimizations update parameters of the same network, the difference between the two optimization objectives yields different gradient descent directions, causing potential conflicts that reduce the effectiveness of the training (see ablation studies in Section~\ref{sec:ablation}). 

To make the test-time optimization more compatible with the training objective, we propose a method that unifies the training and test-time optimization objectives by introducing an auxiliary regression network $f_{\mathbf{u}}$ parameterized by $\mathbf{u}$ which is trained together with the main network:
\begin{equation}
\small
\begin{aligned}
    \mathbf{w}_{meta}, \mathbf{u}_{meta} = \mathop{\arg\min}_{\mathbf{w},\mathbf{u}} \sum_{i=1}^B\sum_{j=1}^M (\mathcal{L}^1_{train}( f_{\mathbf{w}'_{i,j}}(\mathbf{I}_{i,j}), \hat{\Theta}_{i,j}, \hat{\mathbf{J}}_{i,j}   ) + \mathcal{L}^2_{train}( f_{\mathbf{u}}(\mathbf{I}_{i,j}), \hat{\Theta}_{i,j}, \hat{\mathbf{J}}_{i,j})).
\end{aligned}
\label{eq:combined}
\end{equation}
The above equation is a combination of Eq.~\ref{eq:regression-optimization} and Eq.~\ref{eq:our-objective} (superscript 1 and 2 are used to denote the first and second term respectively), with Eq.~\ref{eq:regression-optimization} applied to the auxiliary network $f_{\mathbf{u}}$, and Eq.~\ref{eq:our-objective} applied to $f_\mathbf{w}$. We use $f_{\mathbf{u}}$ to generate a pseudo GT mesh $\hat{\Theta}_{{i,j}}^{u}$ for a training image $\mathbf{I}_{i,j}$, i.e., $\hat{\Theta}_{{i,j}}^{u} = f_{\mathbf{u}}(\mathbf{I}_{i,j})$, and use the pseudo label to supervise the gradient descent in the test-time optimization:
\begin{equation}
    \mathbf{w}'_{i,j} = \mathbf{w} - \alpha \nabla_{\mathbf{w}}\mathcal{L}_{test-u}( f_{\mathbf{w}}(\mathbf{I}_{i,j}), \hat{\Theta}_{{i,j}}^{u}, \hat{\mathbf{J}}_{i,j} ).
    \label{eq:one-step-update2}
\end{equation}
Please compare between Eq. \ref{eq:one-step-update2} and Eq.~\ref{eq:one-step-update}. The difference is that there is an additional input $\hat{\Theta}_{{i,j}}^{u}$ to $\mathcal{L}_{test-u}$ in Eq. \ref{eq:one-step-update2}, and note that this new form of $\mathcal{L}_{test-u}$ is identical to the form of $\mathcal{L}_{train}$.

\subsection{Inference with Dual Networks}
\label{sec:test-time-fine-tuning-inference}
Both the training and testing pseudo codes of our method are given in Algorithm \ref{alg:1}. The training process is fully elaborated in the above sections. Now we introduce how to perform inference at test time.
For each test sample $\mathbf{I}$, at our hand are two networks $f_{\mathbf{w}}$ and $f_{\mathbf{u}}$ with $\mathbf{w}=\mathbf{w}_{meta}$ and $\mathbf{u}=\mathbf{u}_{meta}$, respectively. We freeze the parameters of the auxiliary network, and use it to compute the pseudo GT human mesh for the test image $\mathbf{I}$. Then, under the supervision of the pseudo mesh, we compute $\mathcal{L}_{test-u}$ and use Eq.~\ref{eq:one-step-update2} to iteratively update the parameters $\mathbf{w}$ of $f$ from $\mathbf{w}_{meta}$ to $\mathbf{w}_{final}$. We run at most $m=14$ iterations, and automatically stop the iteration if losses of two consecutive iterations are close enough. We finally use $f_{\mathbf{w}_{final}}$ to estimate the human mesh for image $\mathbf{I}$.

\subsection{Implementation Details}

We implement the main network $f_{\mathbf{w}}$ and auxiliary network $f_{\mathbf{u}}$ with the same network architecture but different parameters. Specifically, we use HMR \cite{kanazawa2018end} or CLIFF \cite{li2022cliff} as $f$ due to their simplicity. The two methods and many other approaches \cite{lin2021end, cho2022cross, xue20223d, zhang2021pymaf, zhang2023pymaf, kocabas2021pare, yoshiyasu2023deformable} adopt ResNet-50 \cite{he2016identity} or HRNet-W48 \cite{sun2019deep} to extract features from the input image, and estimate human mesh based on the features. We provide results of both kinds of backbones. 

We implement our method in PyTorch using the Adam optimizer \cite{kinga2015method} with $\beta_1=0.9$ and $\beta_2=0.999$. The batchsize for ResNet backbone is 40, and for HRNet backbone is 30. The number of training epochs for ResNet backbone is 65, and for HRNet backbone is 25. The learning rate $\alpha$ used in the test-time optimization is 1e-5, and the learning rate $\beta$ (see Algorithm \ref{alg:1}) for the training optimization is 1e-4. 
Our method takes about 3 days training on a single NVIDIA RTX3090 GPU.

\section{Experiments}
\label{sec:Experiments}
\subsection{Datasets}
Following previous work \cite{li2022cliff,kocabas2021pare,zheng2023potter}, we employ the following datasets in our experiments: (1) \textbf{Human3.6M} \cite{ionescu2013human3}, an indoor dataset with precise GT human mesh and 2D joints captured through MoCap devices. (2) \textbf{MPI-INF-3DHP} \cite{mehta2017monocular}, another widely used indoor dataset whose GT human meshes are obtained through multi-view reconstruction. (3) \textbf{COCO} \cite{lin2014microsoft} and (4) \textbf{MPII} \cite{andriluka20142d}, two in-the-wild outdoor datasets with human annotated 2D joints for which we use the pseudo GT mesh provided by \cite{li2022cliff}. (5) \textbf{3DPW} \cite{von2018recovering}, a challenging in-the-wild dataset providing accurate human mesh fitted from IMU sensor data. 

\renewcommand\thefootnote{}
\footnotetext{The authors Yongwei Nie and Mingxian Fan signed the license and produced all the experimental results in this paper. Meta did not have access to the datasets.}

\subsection{Training, Testing and Metrics}

Following prior arts \cite{lin2021end, xue20223d, cho2022cross, black2023bedlam}, we first train our method on a mixture of four datasets, including Human3.6M \cite{ionescu2013human3}, MPI-INF-3DHP \cite{mehta2017monocular}, COCO \cite{lin2014microsoft}, and MPII \cite{andriluka20142d}, and then test our method on the test dataset of Human3.6M \cite{ionescu2013human3}. After that, we further fine-tune our model for 5 epochs by introducing the training dataset of 3DPW \cite{von2018recovering}, and then evaluate our method on the test dataset of 3DPW \cite{von2018recovering}. We use \textbf{MPJPE} (Mean Per Joint Position Error), \textbf{PA-MPJPE} (Procrustes-aligned MPJPE), \textbf{PVE} (Mean Per-vertex Error) as the metrics to evaluate our method. 

\begin{table}[!h] \scriptsize
\captionsetup{font=normalsize}
\caption{\textbf{Quantitative comparison with state-of-the-art methods} on 3DPW \cite{von2018recovering} and Human3.6M \cite{ionescu2013human3}. \enquote{$\dagger$}: using 2D joints detected by OpenPose \cite{cao2017realtime}, \enquote{$\ast$}: using 2D joints detected by RSN \cite{cai2020learning}.}
\centering
\tabcolsep=0.1cm
\begin{tabular*}{\columnwidth}{p{0.08cm}<{\centering}p{2.95cm}<{\raggedright}p{1.8cm}<{\centering}p{1.4cm}<{\centering}p{2.0cm}<{\centering}p{0.8cm}<{\centering}p{1.4cm}<{\centering}p{2.0cm}<{\centering}}
\toprule
 & \multirow{2.5}{*}{\ \ Method} & \multirow{2.5}{*}{Backbone} & \multicolumn{3}{c}{3DPW} & \multicolumn{2}{c}{Human3.6M} \\ \cmidrule(lr){4-6} \cmidrule(lr){7-8}
 &     &  & MPJPE$\downarrow$    & PA-MPJPE$\downarrow$    & PVE$\downarrow$   & MPJPE$\downarrow$    & PA-MPJPE$\downarrow$    \\ 
\midrule
\multirow{11}{*}{\rotatebox{90}{Regression-based}}
&\ \ HMR \cite{kanazawa2018end}'18 & Res-50 & 130.0  &  81.3 & - & 88.0  & 56.8  \\
&\ \ PARE \cite{kocabas2021pare}'21 & HR-W32 & 74.5   &   46.5   &  88.6  &   - &  -   \\
&\ \ ROMP \cite{sun2021monocular}'21 & HR-W32 & 76.7  &  47.3  &   93.4  &  -  &    -  \\
&\ \ PyMAF \cite{zhang2021pymaf}'21 & HR-W48 & 74.2  &  45.3  &   87.0  &  54.2  &    37.2  \\
&\ \ METRO \cite{lin2021end}'21 & HR-W64 & 77.1  &  47.9 & 88.2 &   54.0  &   36.7  \\
&\ \ FastMETRO \cite{cho2022cross}'22 & HR-W64 & 73.5  &  44.6 & 84.1 &  52.2  &   33.7  \\
&\ \ CLIFF \cite{li2022cliff}'22 & HR-W48 & 69.0 &  43.0 & 81.2 & 47.1 &   32.7  \\
&\ \ LearnSample \cite{xue20223d}'22 & HR-W32 & 70.5 &  43.3 & 82.7 & 45.9 &   33.5  \\
&\ \ ProPose \cite{fang2023learning}'23 & HR-W48 & 68.3   &  40.6 & 79.4  &   45.7   & \textbf{29.1} \\
&\ \ POTTER \cite{zheng2023potter}'23 & ViT & 75.0   &  44.8 & 87.4  &   56.5   & 35.1 \\
&\ \ DeFormer \cite{yoshiyasu2023deformable}'23 & HR-W48 & 72.9 &  44.3 &  82.6  & 44.8 &   31.6  \\
\midrule
\multirow{10}{*}{\rotatebox{90}{Optimization-based}}
&\ \ LearnedGD \cite{song2020human}'20 & - & - &  55.9 &  - & -  & 56.4  \\
&\ \ HUND \cite{zanfir2021neural}'21 & Res-50 & 81.4 &  57.5 &  - & 69.5  & 52.6  \\
&\ \ SPIN \cite{kolotouros2019learning}'21 & Res-50 & 96.9 &  59.2 &  116.4 & 62.5  & 41.1  \\
&\ \ EFT \cite{joo2021exemplar}'21 & Res-50 & 85.1 & 52.2 & 98.7 & 63.2 & 43.8 \\
&\ \ HybrIK \cite{li2021hybrik}'21 & Res-34 &74.1& 45.0& 86.5&55.4 &33.6 \\
&\ \  NIKI \cite{li2023niki}'23 & HR-W48 &71.3 &40.6& 86.6&-&- \\
&\ \ ReFit \cite{wang2023refit}'23 & HR-W48 & 65.8 & 41.0 & - & 48.4 & 32.2 \\
&\ \ PLIKS \cite{shetty2023pliks}'23 & HR-W48 &66.9& 42.8 & 82.6 & 49.3 & 34.7\\ 
\cmidrule(lr{2pt}){2-8}
&\ \ Ours$_{\rm CLIFF}$$\dagger $ & HR-W48   & 62.9 &  39.7  &  80.1  &  43.9  & 30.3 \\
&\ \ Ours$_{\rm CLIFF}$$\ast $ & HR-W48 & \textbf{62.4} & \textbf{39.5} & \textbf{78.1} &  \textbf{42.0}  & \textbf{29.1} \\
\bottomrule
\end{tabular*}

\label{tab:overall}
\end{table}

\subsection{Comparison with Previous Approaches}

\textbf{Quantitative results.} We present accuracy comparison with SOTA methods in Table~\ref{tab:overall}, including regression-based approaches \cite{kanazawa2018end, kocabas2021pare, sun2021monocular, lin2021end, li2022cliff, fang2023learning, zheng2023potter, yoshiyasu2023deformable, zhang2021pymaf, cho2022cross, xue20223d} and optimization-based approaches \cite{kolotouros2019learning, joo2021exemplar, li2021hybrik, wang2023refit, li2023niki, shetty2023pliks, song2020human, zanfir2021neural}. For all the compared approaches, we report the best results their papers provide. For our method, we adopt CLIFF \cite{li2022cliff} as $f$ and HRNet-W48 as the backbone network. Since our method needs 2D joints for test-time optimization, we report results using joints estimated by OpenPose \cite{cao2017realtime} (denoted by $\dagger $) and RSN \cite{cai2020learning} (denoted by $\ast $).

Please compare ``Ours$_{\rm CLIFF}\dagger$ (HR-W48)'' in Table~\ref{tab:overall} with SOTA approaches that also use HRNet-W48 as backbone. Our method outperforms most of previous approaches. Compared with \cite{li2022cliff}, we improve it from 69.0 to 62.9 taking MPJPE of 3DPW as an example, which is a large margin. If using RSN joints for the test-time optimization, our method can further improve the metrics.

\begin{figure}[!t]
    \centering  
    \begin{minipage}{0.49\linewidth}
        \centering
        \includegraphics[width=0.187\linewidth]{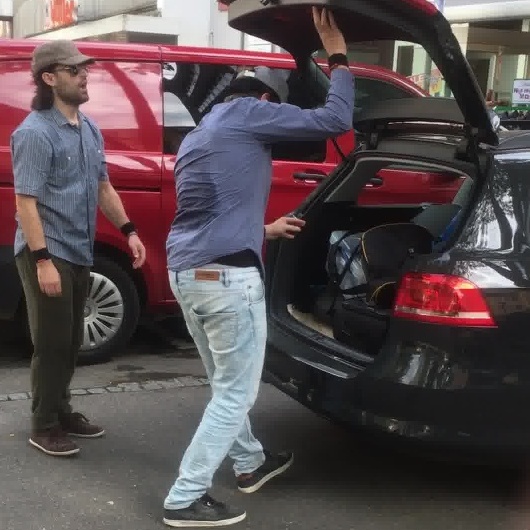}
        \includegraphics[width=0.187\linewidth]{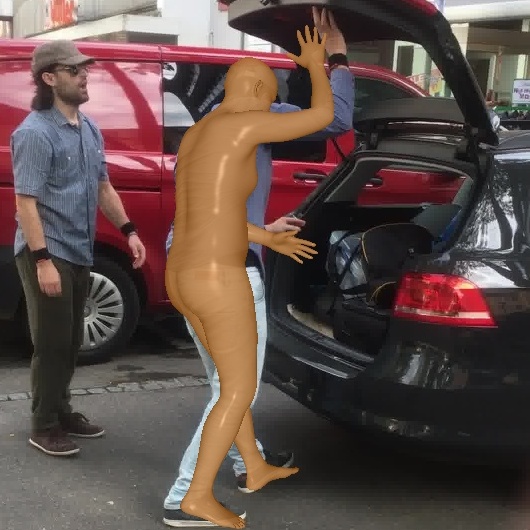}
        \includegraphics[width=0.187\linewidth]{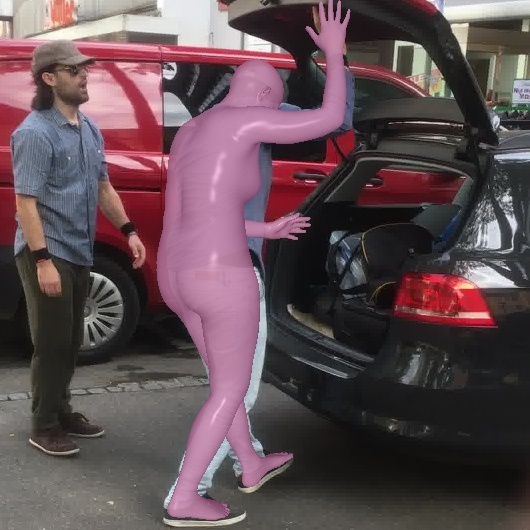}
        \includegraphics[width=0.187\linewidth]{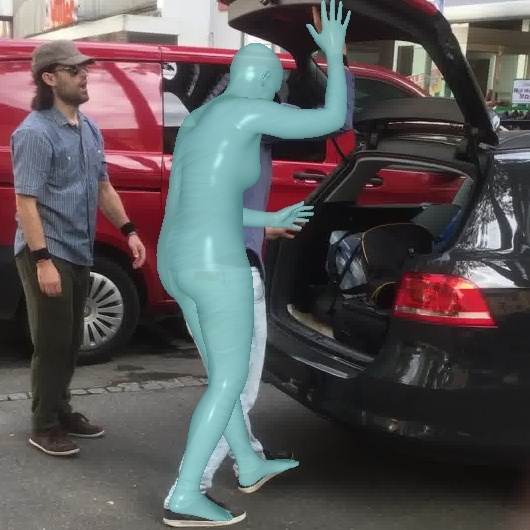}
        \includegraphics[width=0.187\linewidth]{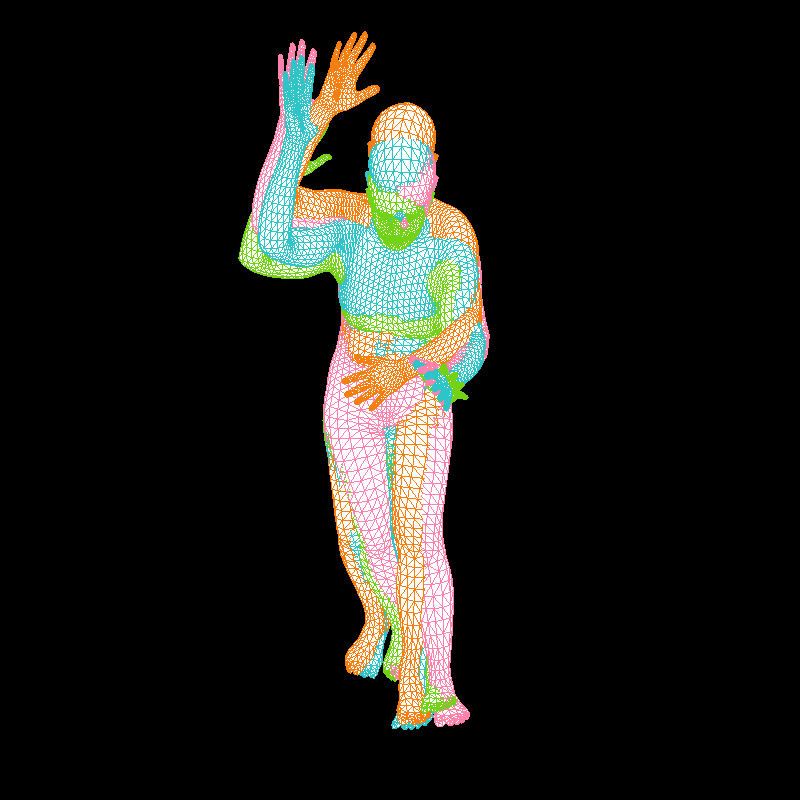}\\
        \includegraphics[width=0.187\linewidth]{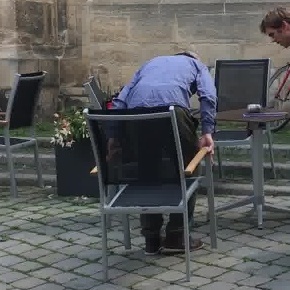}
        \includegraphics[width=0.187\linewidth]{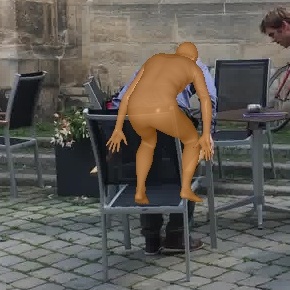}
        \includegraphics[width=0.187\linewidth]{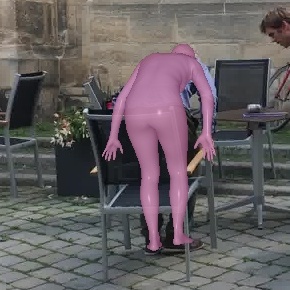}
        \includegraphics[width=0.187\linewidth]{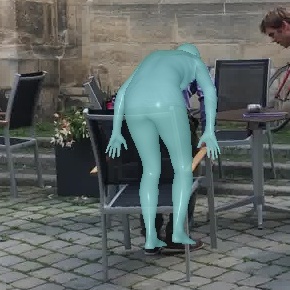}
        \includegraphics[width=0.187\linewidth]{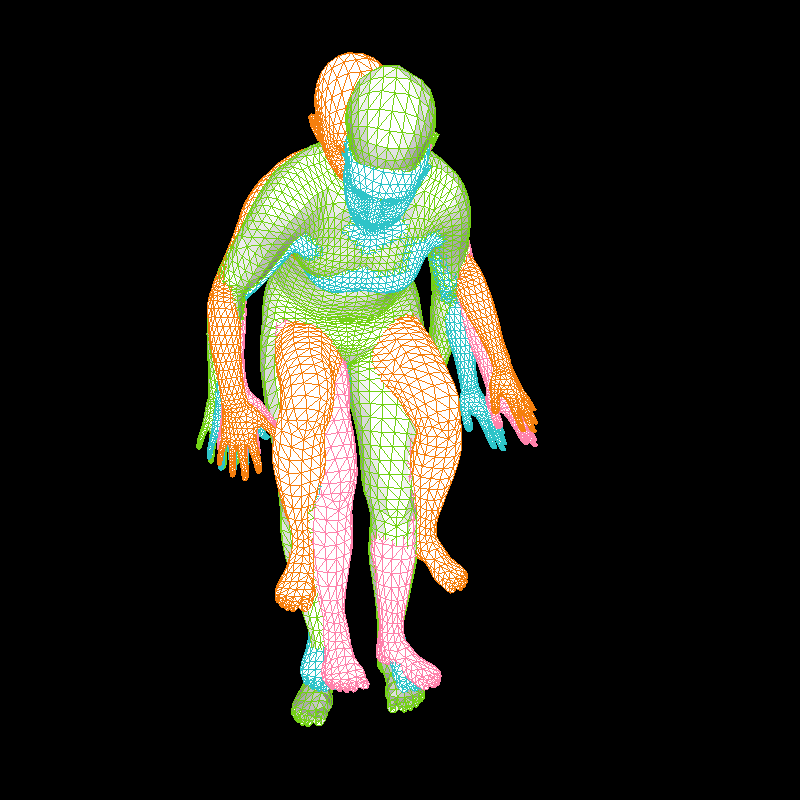}\\
        \includegraphics[width=0.187\linewidth]{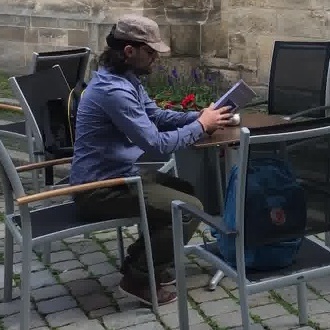}
        \includegraphics[width=0.187\linewidth]{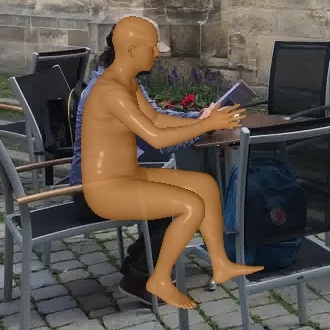}
        \includegraphics[width=0.187\linewidth]{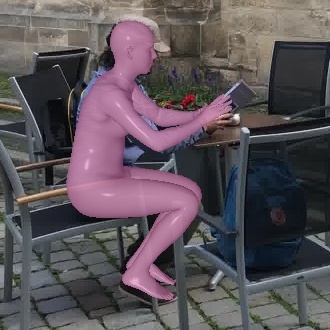}
        \includegraphics[width=0.187\linewidth]{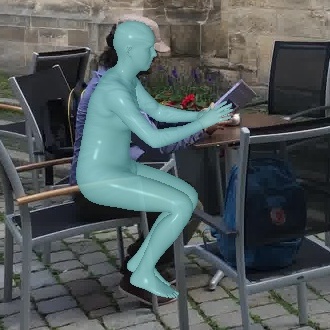}
        \includegraphics[width=0.187\linewidth]{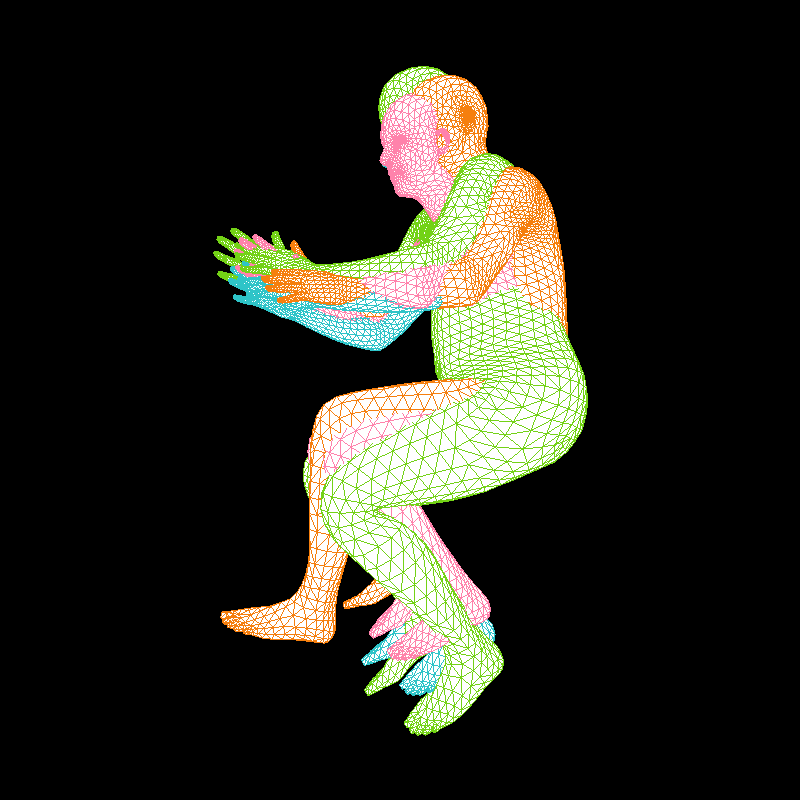}\\
        \includegraphics[width=0.187\linewidth]{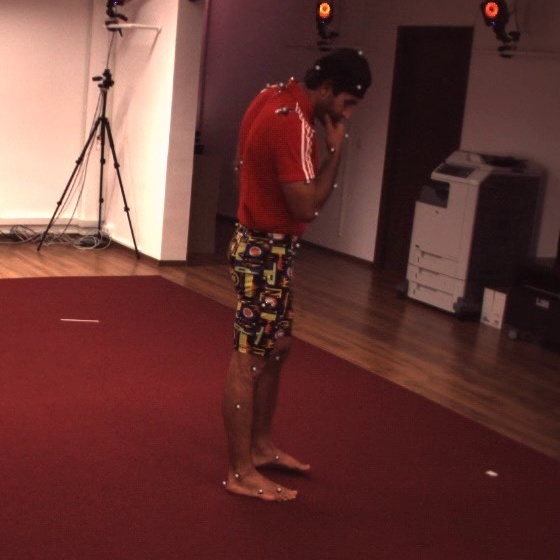}
        \includegraphics[width=0.187\linewidth]{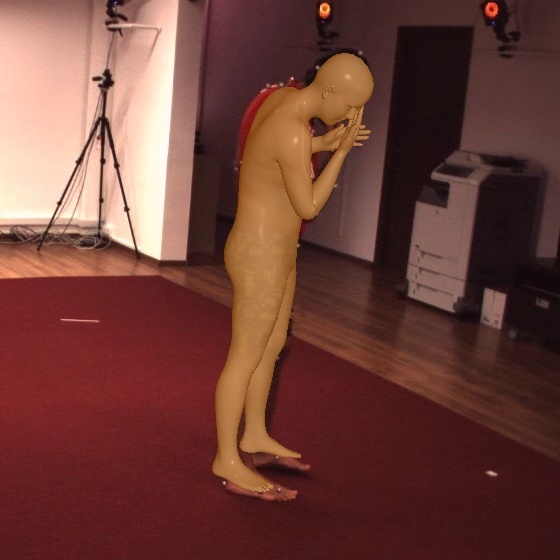}
        \includegraphics[width=0.187\linewidth]{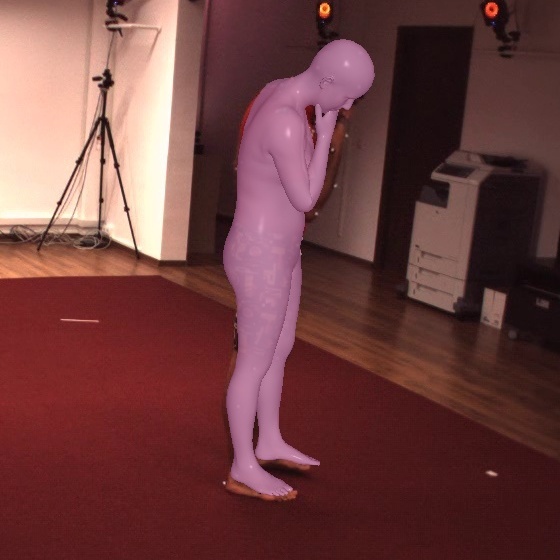}
        \includegraphics[width=0.187\linewidth]{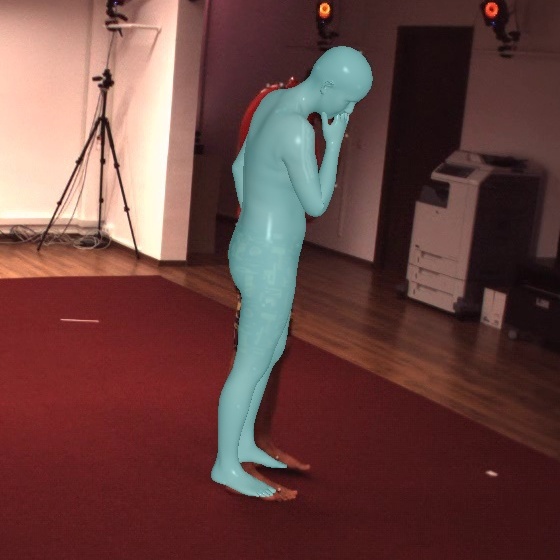}
        \includegraphics[width=0.187\linewidth]{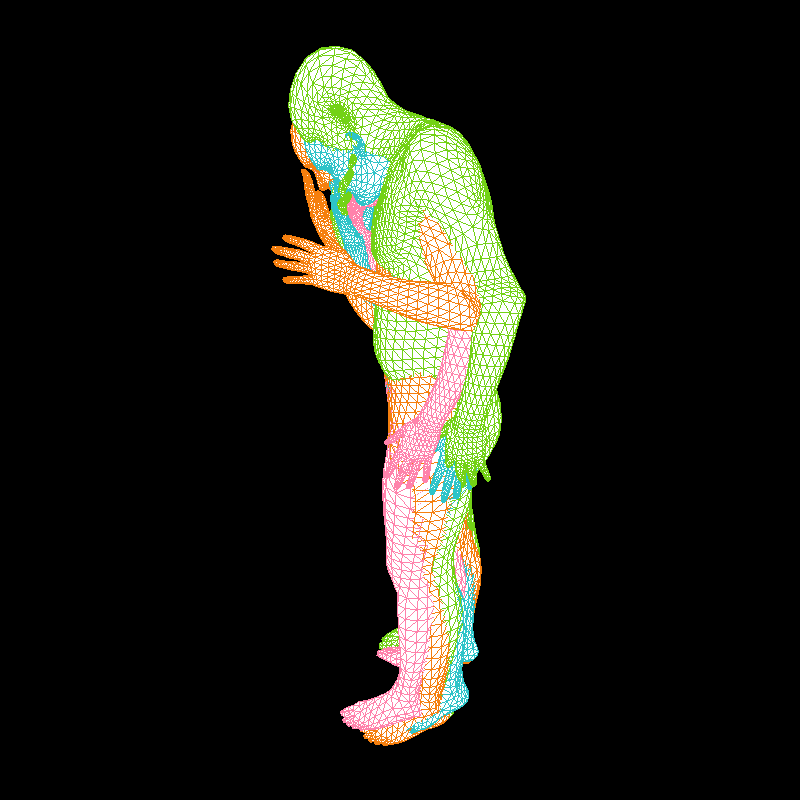}\\
    \end{minipage}
    {\vrule width 0.5pt}
    \begin{minipage}{0.49\textwidth}
        \centering
        \includegraphics[width=0.187\linewidth]{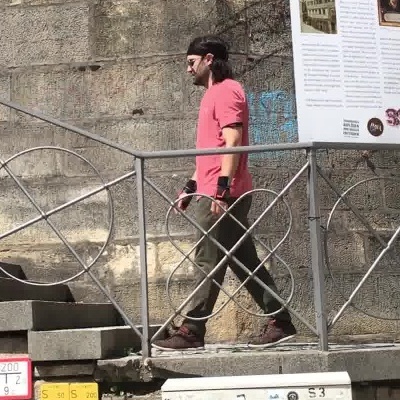}
        \includegraphics[width=0.187\linewidth]{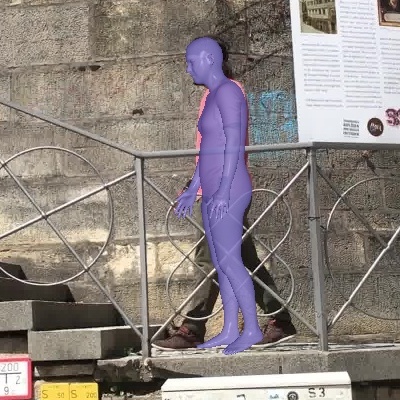}
        \includegraphics[width=0.187\linewidth]{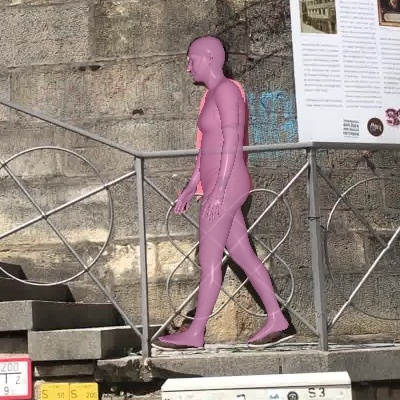}
        \includegraphics[width=0.187\linewidth]{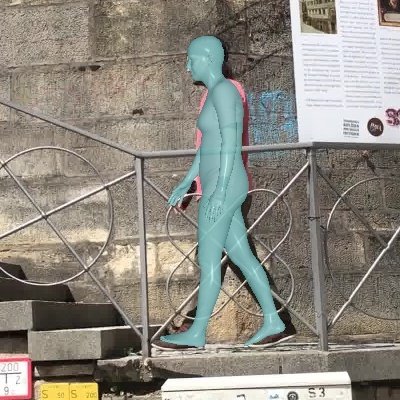}
        \includegraphics[width=0.187\linewidth]{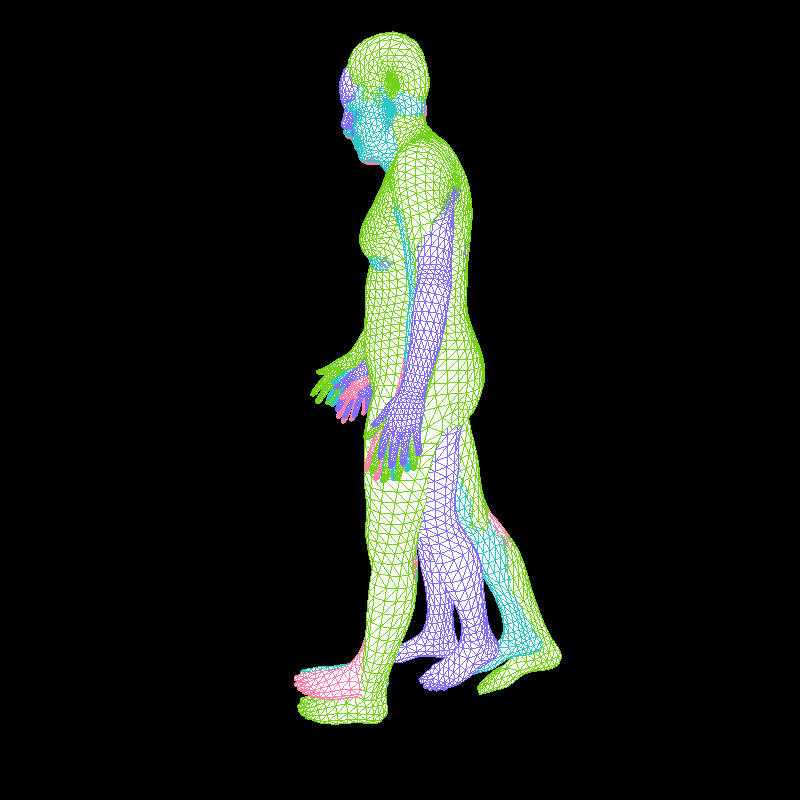}\\
        \includegraphics[width=0.187\linewidth]{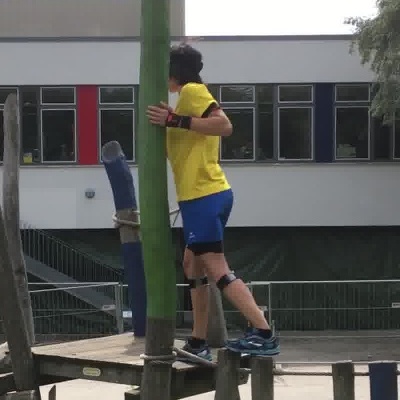}
        \includegraphics[width=0.187\linewidth]{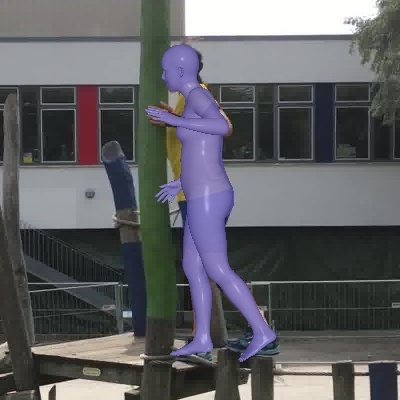}
        \includegraphics[width=0.187\linewidth]{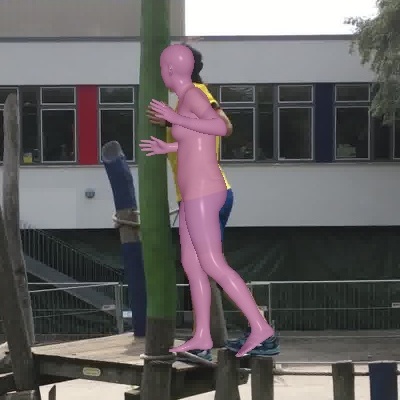}
        \includegraphics[width=0.187\linewidth]{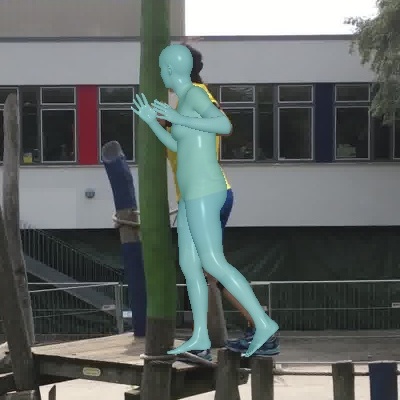}
        \includegraphics[width=0.187\linewidth]{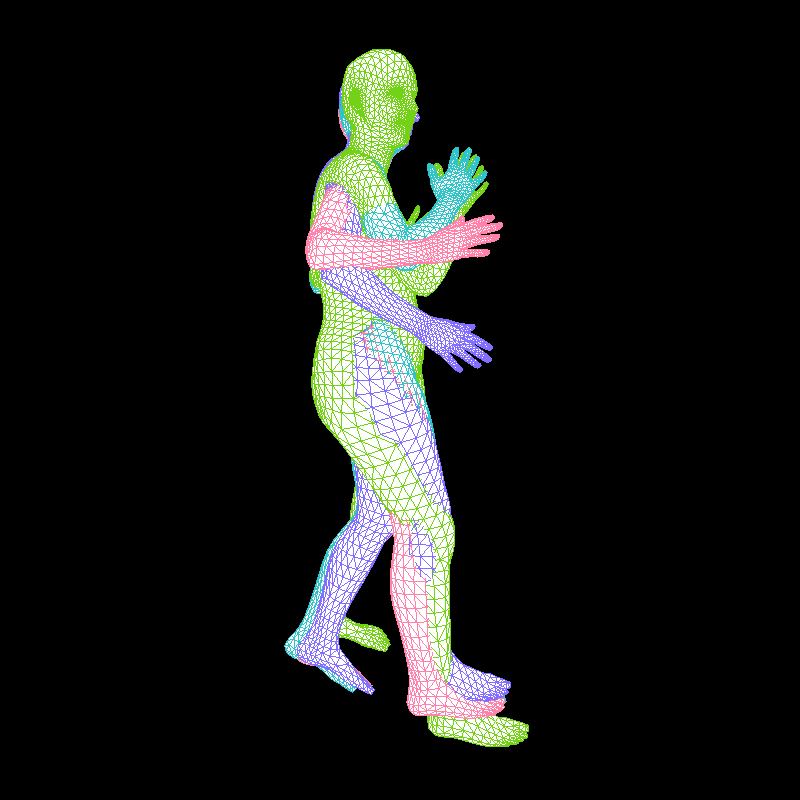}\\
        \includegraphics[width=0.187\linewidth]{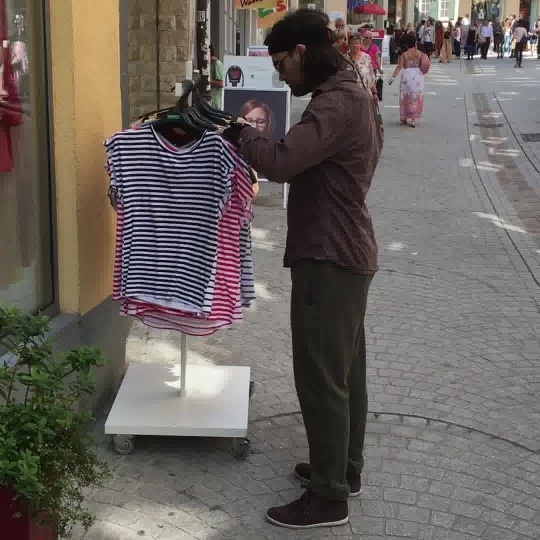}
        \includegraphics[width=0.187\linewidth]{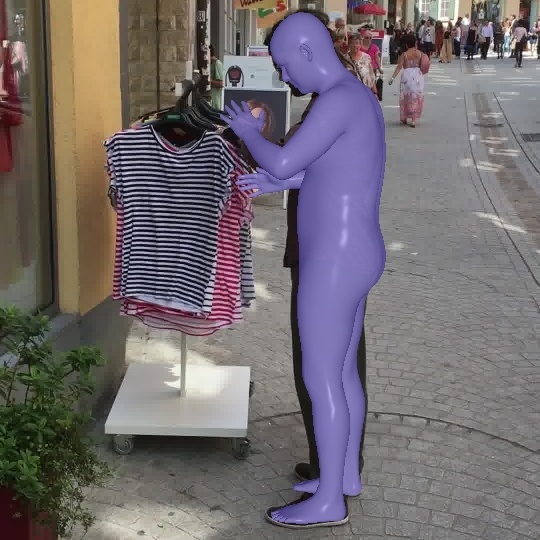}
        \includegraphics[width=0.187\linewidth]{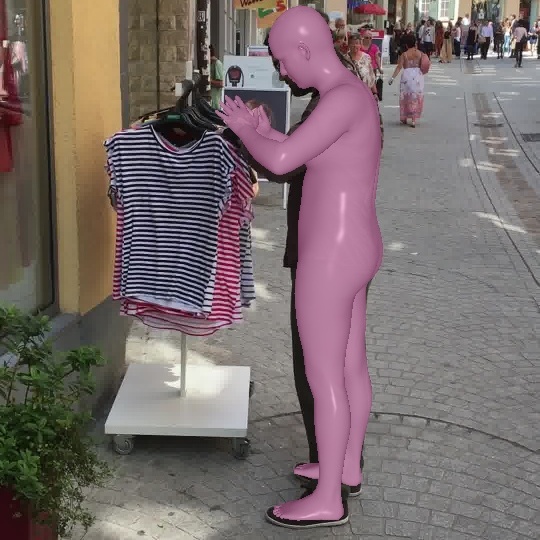}
        \includegraphics[width=0.187\linewidth]{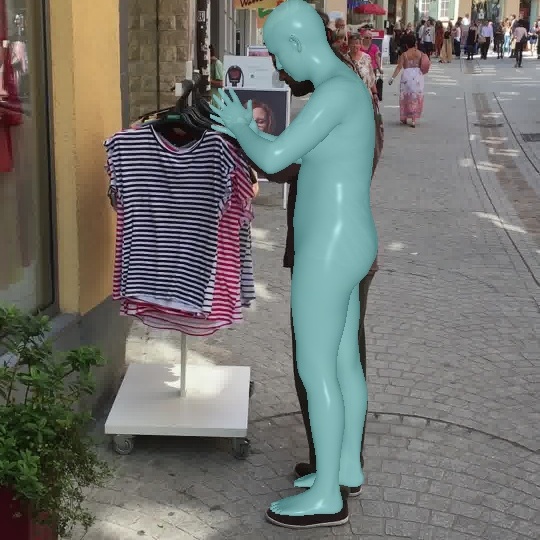}
        \includegraphics[width=0.187\linewidth]{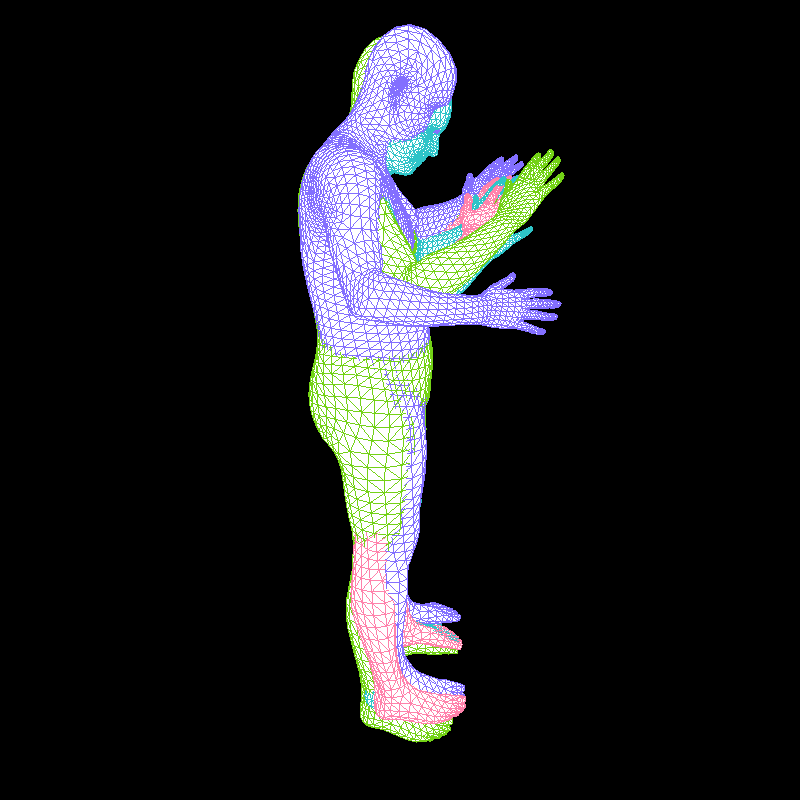}\\
        \includegraphics[width=0.187\linewidth]{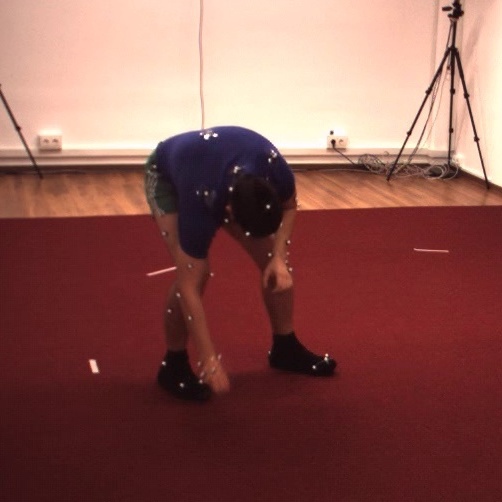}
        \includegraphics[width=0.187\linewidth]{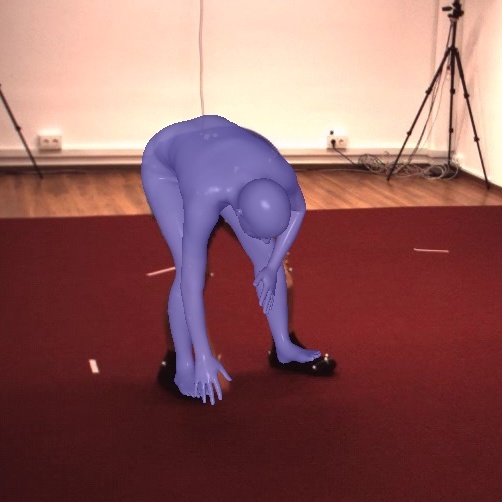}
        \includegraphics[width=0.187\linewidth]{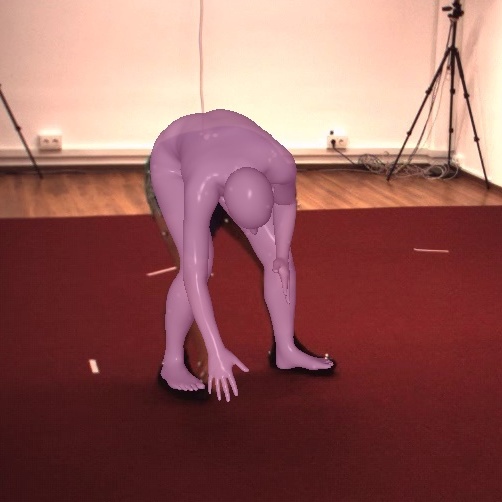}
        \includegraphics[width=0.187\linewidth]{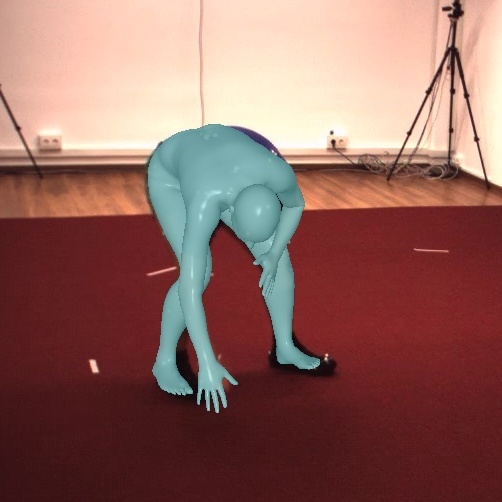}
        \includegraphics[width=0.187\linewidth]{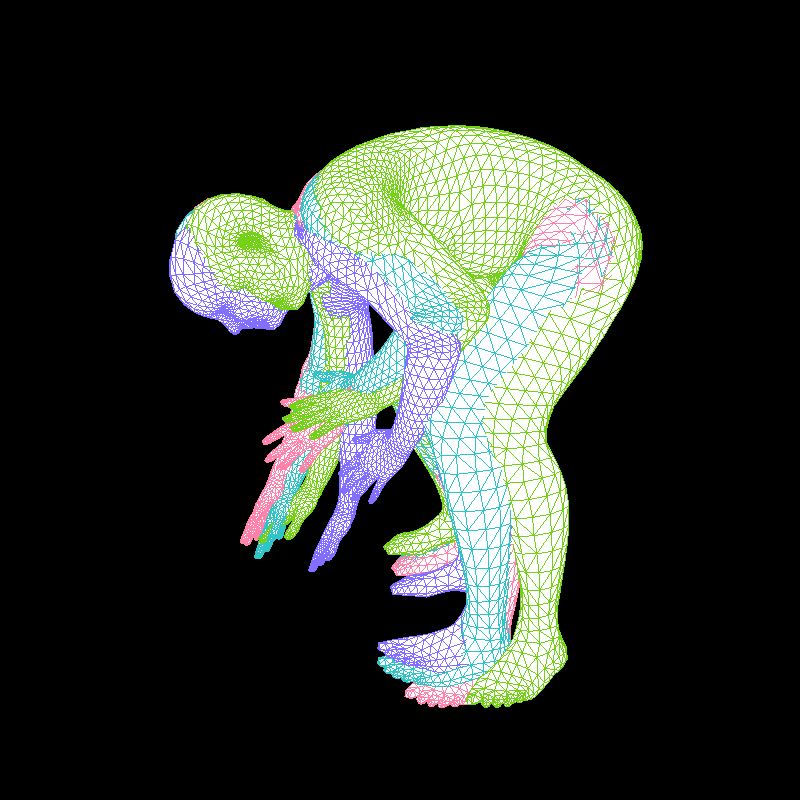}\\
    \end{minipage}
    
    \begin{minipage}{0.49\textwidth} 
        \centering
        \begin{minipage}[t]{0.187\linewidth}
        \caption*{\scalebox{0.75}{Input}}
        \end{minipage}
        \begin{minipage}[t]{0.187\linewidth}
        \caption*{\scalebox{0.75}{CLIFF}}
        \end{minipage}
        \begin{minipage}[t]{0.187\linewidth}
        \caption*{\scalebox{0.75}{Ours$\dagger$}}
        \end{minipage}
        \begin{minipage}[t]{0.187\linewidth}
        \caption*{\scalebox{0.75}{Ours$\ast$}}
        \end{minipage}
        \begin{minipage}[t]{0.187\linewidth}
        \caption*{\scalebox{0.75}{Novel view}}
        \end{minipage}
    \end{minipage}
    \textcolor{white}{\vrule width 0.5pt}
    \begin{minipage}{0.49\textwidth}
        \centering
         \begin{minipage}[t]{0.187\linewidth}
        \caption*{\scalebox{0.75}{Input}}
        \end{minipage}
        \begin{minipage}[t]{0.187\linewidth}
        \caption*{\scalebox{0.75}{Refit}}
        \end{minipage}
        \begin{minipage}[t]{0.187\linewidth}
        \caption*{\scalebox{0.75}{Ours$\dagger$}}
        \end{minipage}
        \begin{minipage}[t]{0.187\linewidth}
        \caption*{\scalebox{0.75}{Ours$\ast$}}
        \end{minipage}
        \begin{minipage}[t]{0.187\linewidth}
        \caption*{\scalebox{0.75}{Novel view}}
        \end{minipage}
    \end{minipage}
    \vspace{-0.5cm}
    \caption{\textbf{Qualitative comparison with SOTA methods.} We show results produced by CLIFF \cite{li2022cliff}, ReFit \cite{wang2023refit}, and our method ({$\dagger$}: OpenPose, $\ast$: RSN). All the three methods use HRNet-W48 as the backbone. In the novel views, green represents the ground truth, orange represents CLIFF, purple represents ReFit, pink and blue represent the two variants of our method, respectively.}
    \label{fig:quantitive}
\end{figure}

\textbf{Qualitative results.} We show qualitative comparison with CLIFF \cite{li2022cliff} and Refit \cite{wang2023refit} in Figure~\ref{fig:quantitive}. Our method estimates faithful human poses and meshes which are better than those of the compared approaches. Comparisons with HybrIK \cite{li2021hybrik}, NIKI \cite{li2023niki}, ProPose \cite{fang2023learning} and EFT$_{\rm CLIFF}$ can be found in the supplementary material.

\subsection{Ablation study}
\label{sec:ablation}

\textbf{Influence of Regression Model.} The adopted regression model $f$ influences the effectiveness of our method. In Table~\ref{tab:2djoints-type}, we show the results of using HMR \cite{kanazawa2018end} or CLIFF \cite{li2022cliff} as the regression model. Since CLIFF is a stronger baseline than HMR, our method based on CLIFF performs better than that based on HMR. 

\textbf{Influence of Accuracy of 2D Joints.} Since our method needs 2D joints as the supervision at test time, it is interesting to see how the quality of the 2D joints affects the effectiveness of the method. We have already repported results on 2D joints detected by OpenPose \cite{cao2017realtime} and RSN \cite{cai2020learning}. In Table~\ref{tab:2djoints-type}, we further test GT 2D joints. As seen, our method with GT joints outperforms our method using detected joints by OpenPose and RSN. This indicates our method will become more effective as 2D pose detectors continue to develop.

\begin{table}[!ht] \scriptsize
\captionsetup{font=normalsize}
\caption{\textbf{Ablation study on regression model and 2D joints} on 3DPW \cite{von2018recovering} and Human3.6M \cite{ionescu2013human3}. \enquote{$\dagger$}: using 2D joints detected by OpenPose \cite{cao2017realtime}, \enquote{$\diamond$}: using GT 2D joints.} 
\centering
\tabcolsep=0.1cm
\begin{tabular*}{\columnwidth}{p{2.8cm}<{\raggedright}p{1.8cm}<{\centering}p{1.5cm}<{\centering}p{2.0cm}<{\centering}p{0.95cm}<{\centering}p{1.5cm}<{\centering}p{2.0cm}<{\centering}}
\toprule
  \multirow{2.5}{*}{\ \ Method} & \multirow{2.5}{*}{Backbone} & \multicolumn{3}{c}{3DPW} & \multicolumn{2}{c}{Human3.6M} \\ \cmidrule(lr){3-5} \cmidrule(lr){6-7}
      &  & MPJPE$\downarrow$    & PA-MPJPE$\downarrow$    & PVE$\downarrow$   & MPJPE$\downarrow$    & PA-MPJPE$\downarrow$    \\ 
\midrule
\ \ Ours$_{\rm HMR}$$\dagger$ & Res-50 & 73.3  &   44.3   &  90.3   &   55.8  &   36.4    \\
\rowcolor{gray!20}\ \ Ours$_{\rm HMR}$$\diamond$ & Res-50   &  68.9    &  39.6   &  85.5   &  53.7 &  33.8 \\
\ \ Ours$_{\rm CLIFF}$$\dagger $ & HR-W48   & 62.9 &  39.7  &  80.1  & 43.9  & 30.3 \\
\rowcolor{gray!20}\ \ Ours$_{\rm CLIFF}$$\diamond $  & HR-W48   &57.8 & 35.3 &  74.4 &  39.4  &  27.5   \\ 
\bottomrule
\end{tabular*}
\label{tab:2djoints-type}
\end{table}

\begin{figure}[!ht]
    \centering
    \begin{minipage}{1.0\columnwidth}
    \centering
    \includegraphics[height=0.45cm]{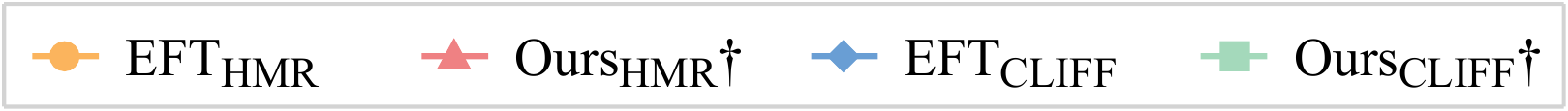}
    \end{minipage}
    \begin{minipage}{1.0\columnwidth}
    \centering
    \includegraphics[width=0.30\linewidth]{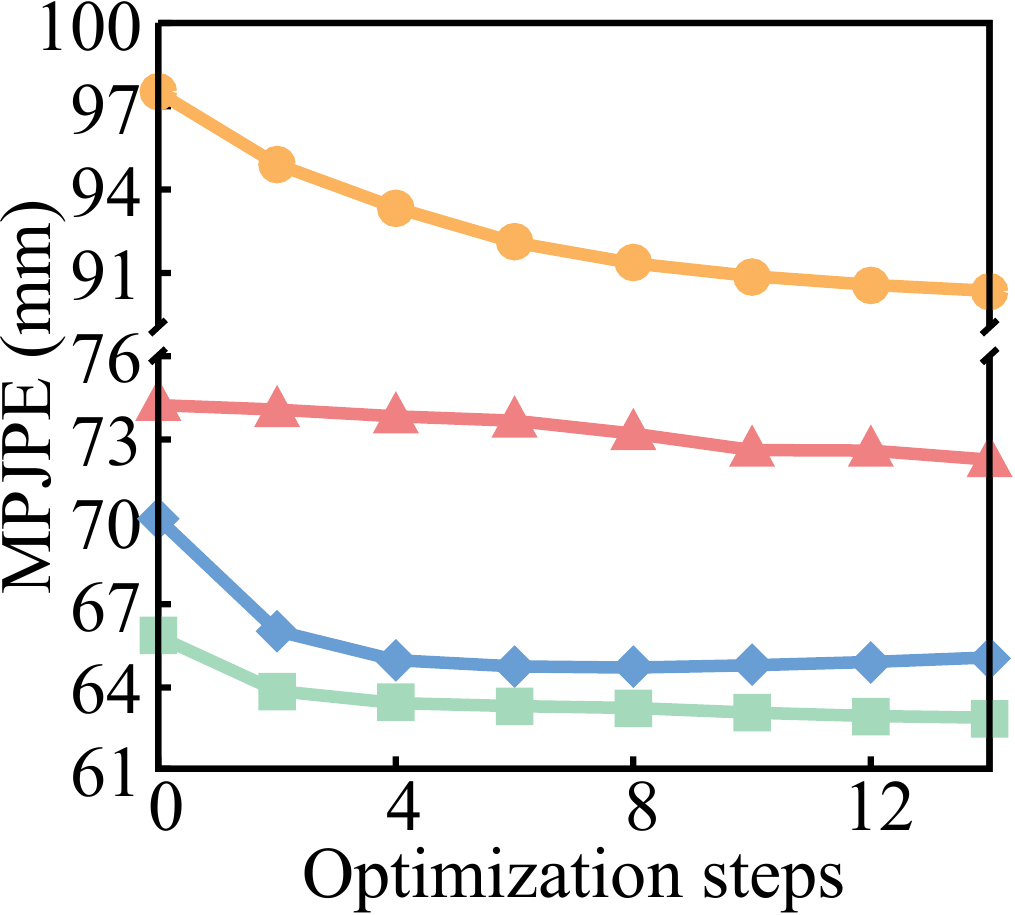}
    \includegraphics[width=0.30\linewidth]{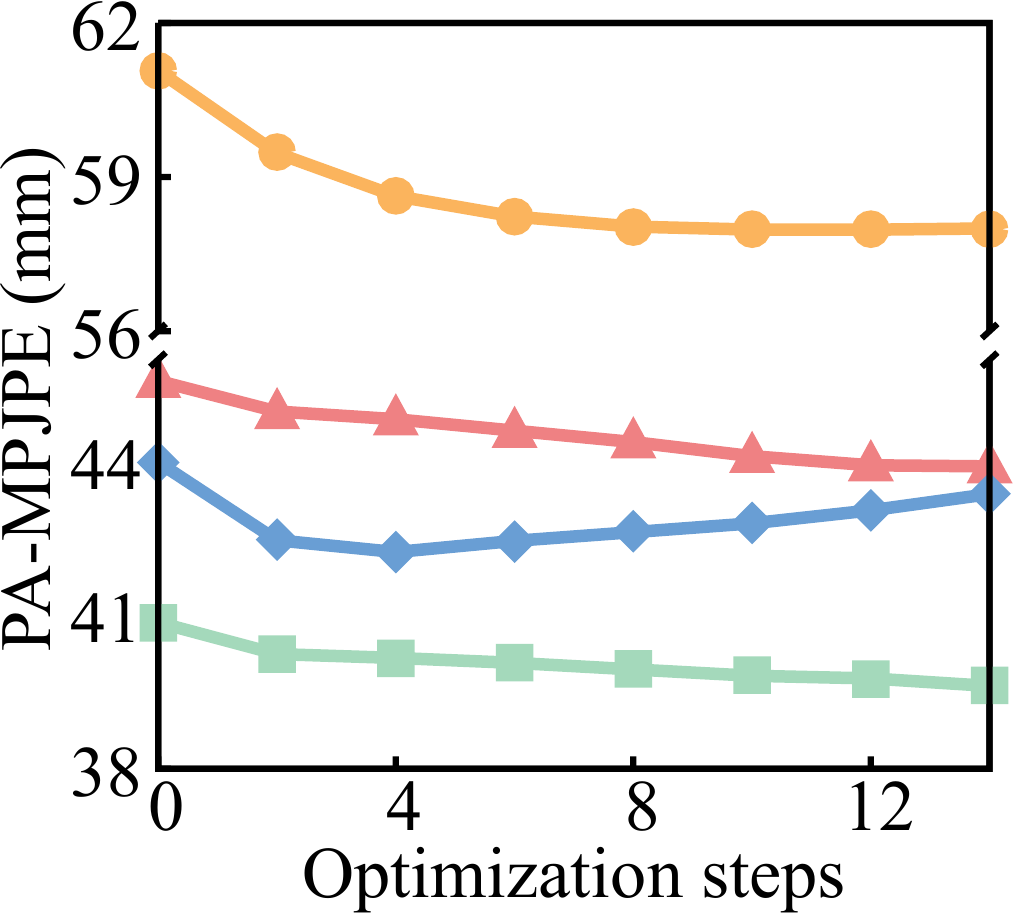}
    \includegraphics[width=0.34\linewidth]{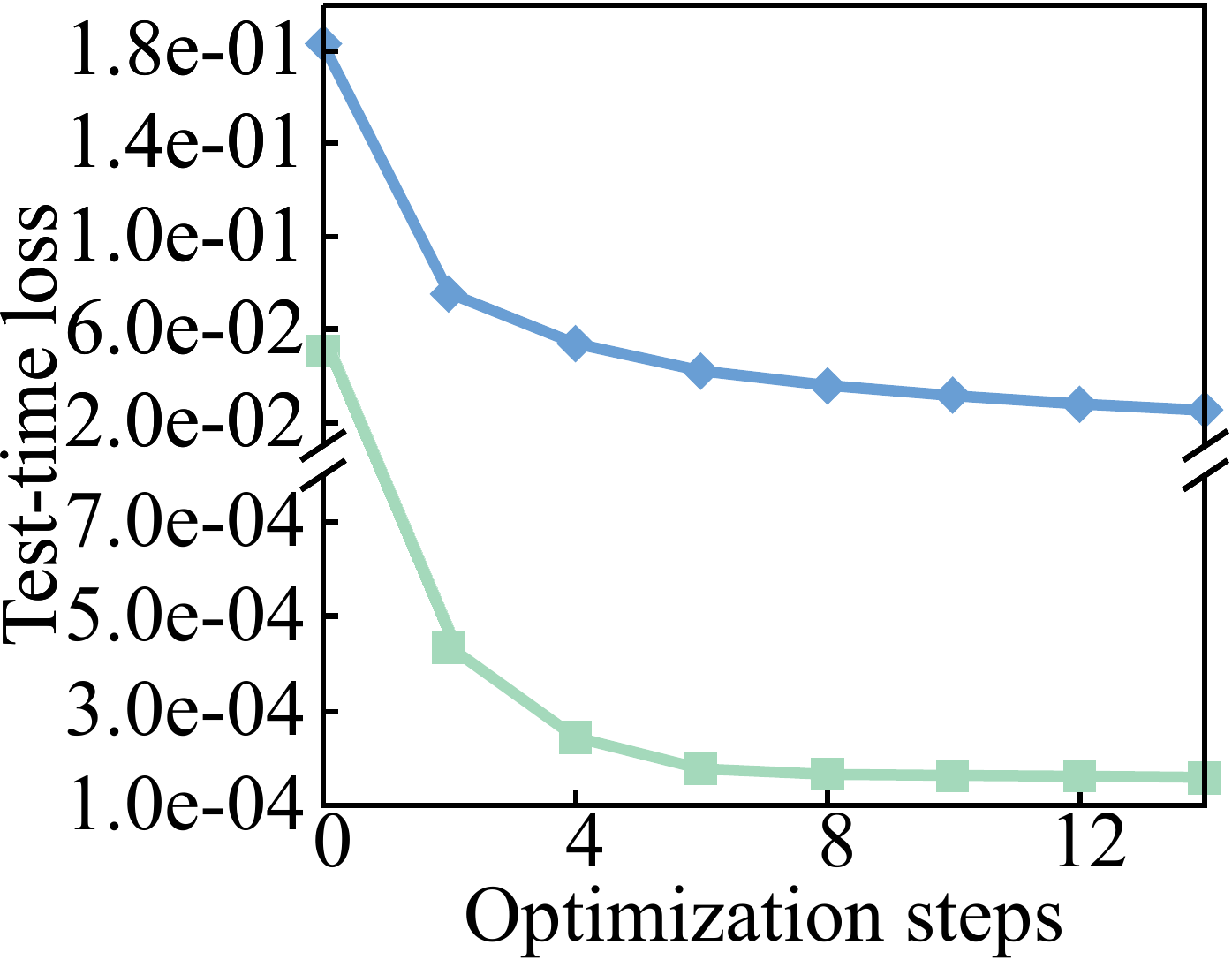}
    \end{minipage}
    \vspace{-0.1cm}
    \begin{minipage}{1.0\columnwidth}
    \raggedright
    \hspace{2.4cm}(a) \hspace{3.7cm}(b) \hspace{4.4cm}(c)
    \end{minipage}
    \caption{\textbf{Influence of optimization steps during inference.} Our method outperforms EFT when using the same regression model. As optimization proceeds, our results continuously become better, while those of EFT become better at first and then become worse (see (a) and (b)). (c) shows that our method achieves faster convergence compared to EFT.}
    \label{fig:op-step}
\end{figure}

\textbf{Influence of Optimization Steps at Inference Time.} At inference time, we perform at most $m$ test-time optimization steps. In Figure~\ref{fig:op-step} (a) and (b), we show how the evaluation metrics become as the number of optimization steps increases. As seen, our results consistently become better in terms of both MPJPE and PA-MPJPE. We also show the results of EFT$_{\rm HMR}$ and EFT$_{\rm CLIFF}$. With the same regression model, our method is better than EFT \cite{joo2021exemplar}. The results of EFT become better at the first few optimization steps, but become worse as more optimization steps execute. This is probably because EFT is only finetuned with 2D reprojection loss and is more sensitive to the errors in the 2D joints. At the first several optimization steps, the estimated 3D SMPL approaches the 2D joints from a relatively distant initialization, therefore the result gets better gradually. With more optimization steps, the SMPL may overfit the 2D joints whose annotation-errors then distort the SMPLs, thus yielding worse evaluation metrics. In contrast, our method is guided by both 3D and 2D supervisions. The 3D pseudo SMPLs plays the role of regularization that mitigates the influences of errors in 2D joints (see more explanations in the supplemental material). In Figure~\ref{fig:op-step} (c), we present the loss curve of the test-time optimization of EFT and our method. Our method converges in about 6 steps, demonstrating a faster convergence compared with EFT.

\begin{figure}[!t]
    \centering
    \begin{minipage}{1.0\columnwidth}
        \centering
        \begin{minipage}[t]{0.135\columnwidth}
        \caption*{\scalebox{0.85}{Input}}
        \end{minipage}
        \begin{minipage}[t]{0.135\columnwidth}
        \caption*{\scalebox{0.85}{Step 0}}
        \end{minipage}
        \begin{minipage}[t]{0.135\columnwidth}
        \caption*{\scalebox{0.85}{Step 1}}
        \end{minipage}
        \begin{minipage}[t]{0.135\columnwidth}
        \caption*{\scalebox{0.85}{Step 2}}
        \end{minipage}
        \begin{minipage}[t]{0.135\columnwidth}
        \caption*{\scalebox{0.85}{Step 3}}
        \end{minipage}
        \begin{minipage}[t]{0.135\columnwidth}
        \caption*{\scalebox{0.85}{Step 4}}
        \end{minipage}
        \begin{minipage}[t]{0.135\columnwidth}
        \caption*{\scalebox{0.85}{Final}}
        \end{minipage}
    \end{minipage}
    \vspace{-0.45cm}
    \\
    \centering
    \begin{minipage}{1.0\columnwidth}
        \includegraphics[width=0.135\linewidth]{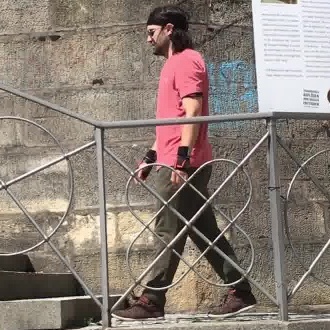}
        \includegraphics[width=0.135\linewidth]{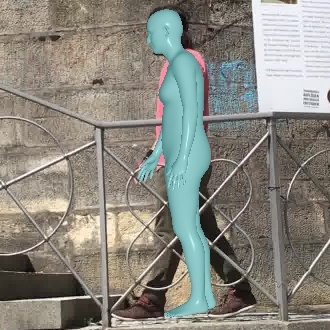}
        \includegraphics[width=0.135\linewidth]{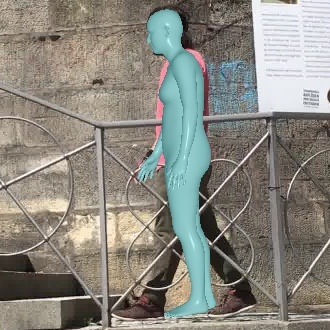}
        \includegraphics[width=0.135\linewidth]{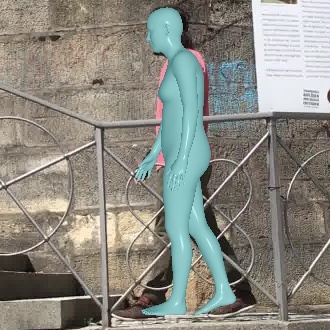}
        \includegraphics[width=0.135\linewidth]{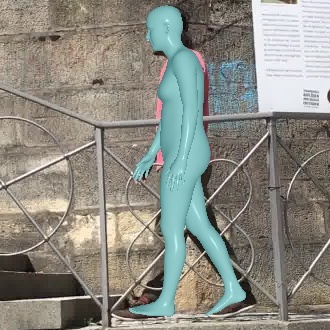}
        \includegraphics[width=0.135\linewidth]{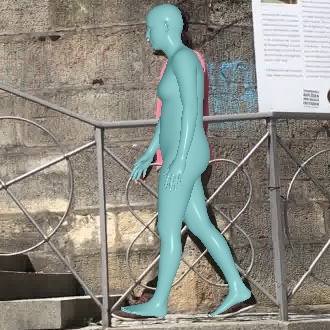}
        \includegraphics[width=0.135\linewidth]{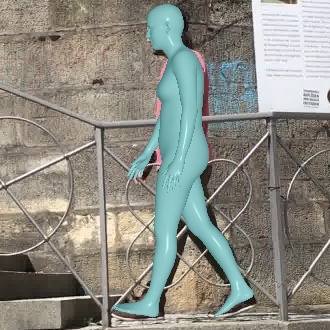}\\
        \includegraphics[width=0.135\linewidth]{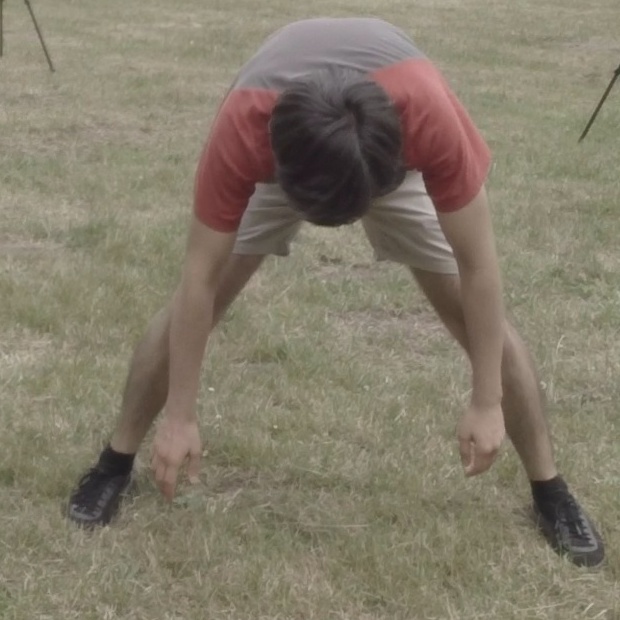}
        \includegraphics[width=0.135\linewidth]{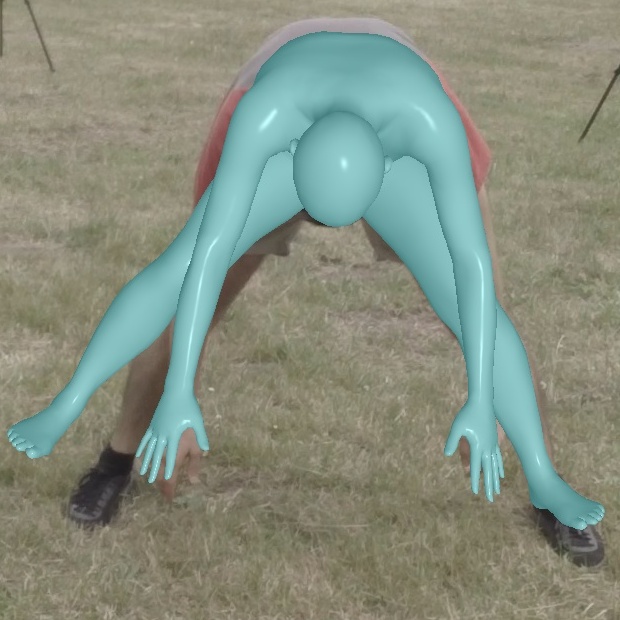}
        \includegraphics[width=0.135\linewidth]{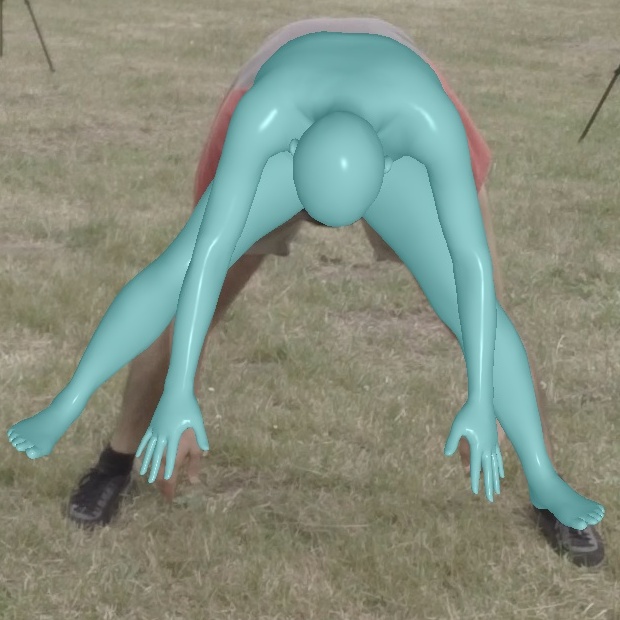}
        \includegraphics[width=0.135\linewidth]{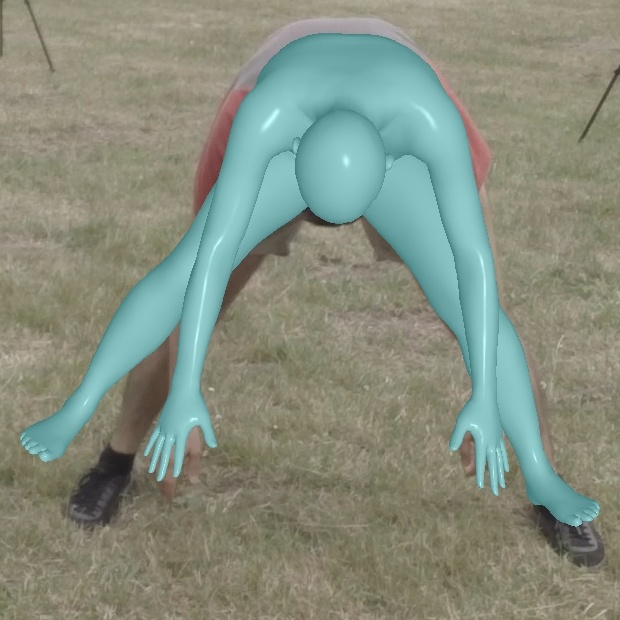}
        \includegraphics[width=0.135\linewidth]{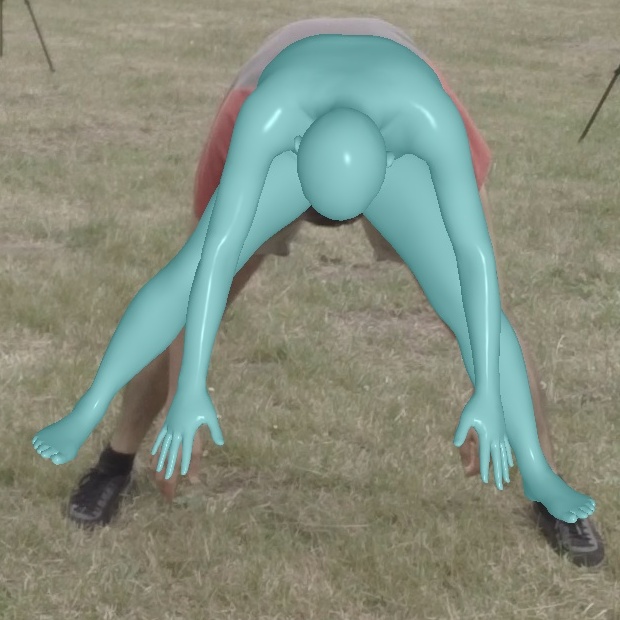}
        \includegraphics[width=0.135\linewidth]{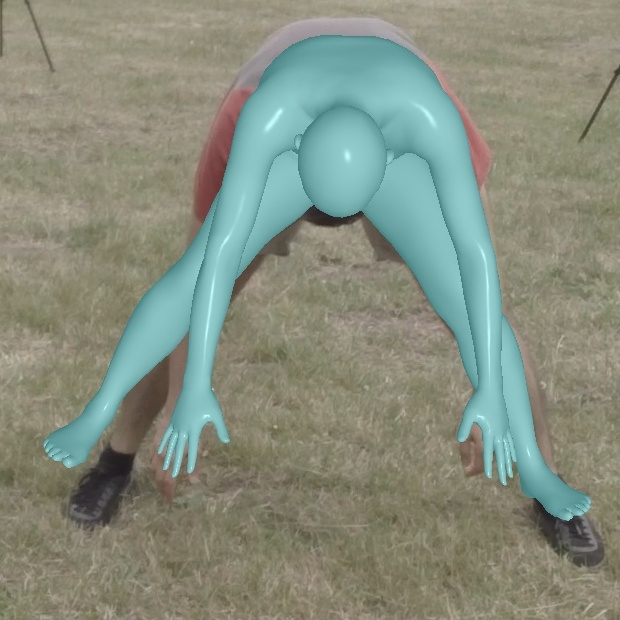}
        \includegraphics[width=0.135\linewidth]{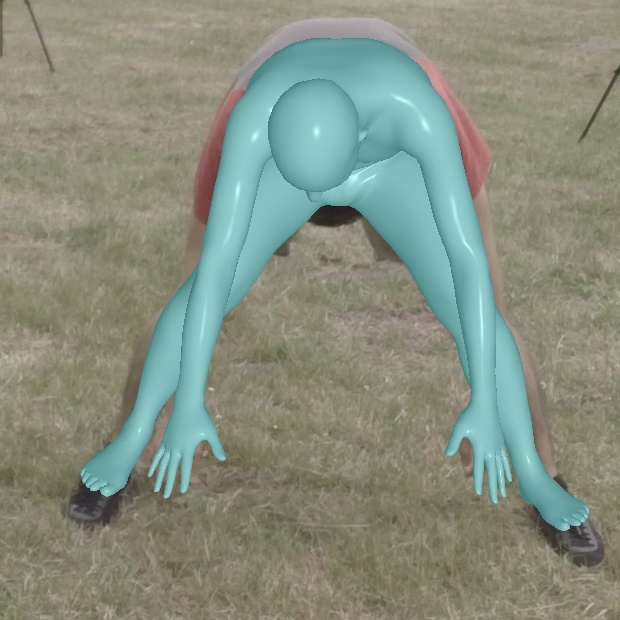}   
    \end{minipage}
    \caption{\textbf{Stepwise visualization.} From left to right, we showcase results after different steps of test-time optimization during testing.}
    \label{fig:deform}
\end{figure}

Figure~\ref{fig:deform} shows our meshes after different optimization steps. Initially, the mesh does not fit with the target human. After more steps, the mesh progressively deforms itself to achieve perfect fitting. 


\begin{table}[!t]\scriptsize
\captionsetup{font=normalsize}
\caption{\textbf{Ablation study on meta-learning and auxiliary network.} Models are trained on COCO \cite{lin2014microsoft} and tested on 3DPW \cite{von2018recovering}. ``Test. Opt.'' is ``test-time optimization''.}
\renewcommand\arraystretch{1.25}
\centering
\tabcolsep=0.06cm
\begin{tabular*}{\columnwidth}{m{2.9cm}<{\centering}m{3.7cm}<{\centering}m{2.6cm}<{\centering}m{1.85cm}<{\centering}m{2.3cm}<{\centering}}
\toprule  
 Model  & Integrating Test. Opt. (meta-learning) & Auxiliary Net  & MPJPE $\downarrow$      & PA-MPJPE $\downarrow$   \\
\midrule
EFT$_{\rm CLIFF}$ &\usym{2715}  & \usym{2715} & 84.6 & 54.2 \\
Ours$_{\rm CLIFF}$ & \usym{1F5F8} & \usym{2715} & 78.5 & 49.9 \\
\rowcolor{gray!20} Ours$_{\rm CLIFF}$ & \usym{1F5F8} &\usym{1F5F8} & \textbf{76.7} & \textbf{49.5} \\
\bottomrule
\end{tabular*}
\label{tab:ablation3}
\end{table}
\textbf{Influence of Meta-Learning and Dual Networks.} We propose meta-learning to improve the performance of test-time optimization. We also introduce an auxiliary network to unify the formulation of test-time and training optimizations, and hope this can reduce the conflict in training and improve the estimation accuracy. We conduct an ablation study to validate the two components as shown in Table \ref{tab:ablation3}. We train all the models in the ablation study on COCO \cite{lin2014microsoft} and test them on 3DPW \cite{von2018recovering}. The first row in Table \ref{tab:ablation3} shows results of training with no test-time (Test.) optimization (Opt.) and no auxiliary network, \ie, CLIFF \cite{li2022cliff}. The second row shows our method without auxiliary network. The third row shows our full method. As shown, the introduction of the meta-learning and auxiliary network strategies both improve the evaluation results.

\begin{table}[!t]\scriptsize
\captionsetup{font=normalsize}
\begin{minipage}{0.45\textwidth}
\caption{\textbf{Quantitative comparison} with EFT$_{\rm CLIFF}$ on the LSP-Extended dataset~\cite{johnson2011learning}.}
\centering
\begin{tabular*}{\columnwidth}{p{3cm}<{\centering}p{2cm}<{\centering}}
\toprule
Method  & 2D Loss \\
\midrule
EFT$_{\rm CLIFF}$ & 8.3e-3 \\
Ours$_{\rm CLIFF}$ &  6.1e-3 \\
\bottomrule
\end{tabular*}
\label{tab:ood-lsp}
\end{minipage}
\hspace{0.05\textwidth}
\begin{minipage}{0.45\textwidth}
\caption{\textbf{Quantitative comparison} with EFT$_{\rm CLIFF}$ on the Human3.6M dataset~\cite{ionescu2013human3}.}
\centering
\begin{tabular*}{\columnwidth}{p{2cm}<{\centering}p{1cm}<{\centering} p{2cm}<{\centering}}
\toprule
Method  & MPJPE$\downarrow$ & PA-MPJPE$\downarrow$ \\
\midrule
EFT$_{\rm CLIFF}$ & 85.5 & 51.0 \\
Ours$_{\rm CLIFF}$ & 83.8 & 48.6 \\
\bottomrule
\end{tabular*}
\label{tab:ood-h36m}
\end{minipage}
\end{table}

\begin{figure}[!t]
    \centering
    \begin{subfigure}{1.0\columnwidth}
    \centering
        \includegraphics[width=0.24\linewidth]{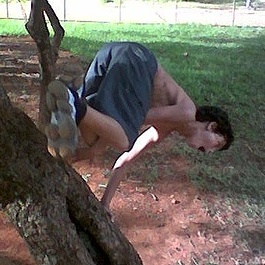}
        \includegraphics[width=0.24\linewidth]{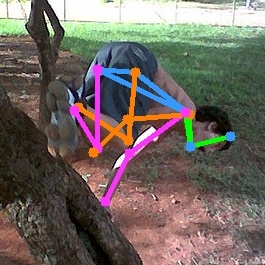}
        \includegraphics[width=0.24\linewidth]{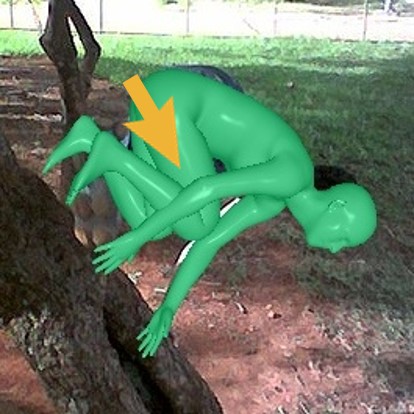}
        \includegraphics[width=0.24\linewidth]{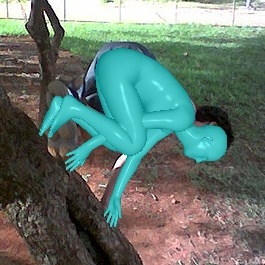}\\
        \includegraphics[width=0.24\linewidth]{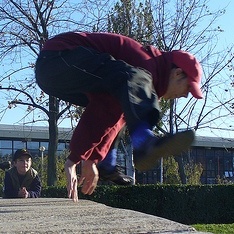}
        \includegraphics[width=0.24\linewidth]{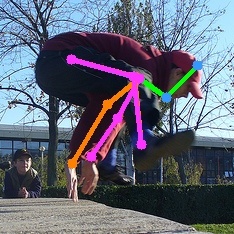} 
        \includegraphics[width=0.24\linewidth]{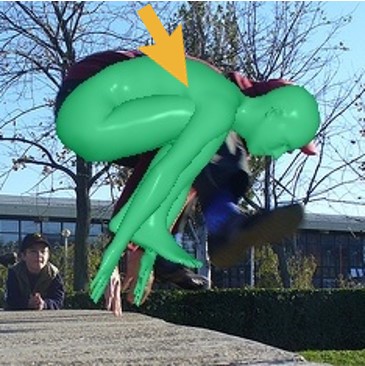}
        \includegraphics[width=0.24\linewidth]{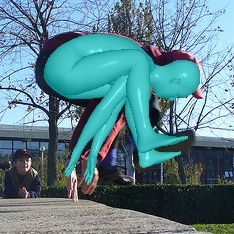}
    \end{subfigure}\\
    \vspace{-0.25cm}
    \centering
    \begin{minipage}{1.0\columnwidth}
        \begin{minipage}[t]{0.24\linewidth}
        \caption*{\scalebox{1.0}{Input}}
        \end{minipage}
        \begin{minipage}[t]{0.24\linewidth}
        \caption*{\scalebox{1.0}{GT Joints}}
        \end{minipage}
        \begin{minipage}[t]{0.24\linewidth}
        \caption*{\scalebox{1.0}{EFT$_{\rm CLIFF}$}}
        \end{minipage}
        \begin{minipage}[t]{0.24\linewidth}
        \caption*{\scalebox{1.0}{Ours}}
        \end{minipage}
    \end{minipage}
    \vspace{-0.35cm}
    \caption{\textbf{Out-of-domain adaptation.} Please see the depth order of the arms where EFT$_{\rm CLIFF}$ fails to infer. Our method correctly identify the correct configuration.}
    \label{fig:Out of domain}
\end{figure}

\textbf{Out-of-Domain Adaptation.} Test-time optimization performs post-processing on each test image and has the ability of adaptation to out-of-domain data. To comprehensively validate our method's effectiveness in out-of-distribution (OOD) scenarios, we first conduct a quantitative comparison with EFT$_{\rm CLIFF}$ on the LSP-Extended dataset, as shown in Table~\ref{tab:ood-lsp} (Training dataset: COCO, MPII, MPI-INF-3DHP, Human3.6M, 3DPW, Backbone: HR-W48). Since LSP provides ground truth 2D joints but not GT SMPLs, the comparison is based on 2D loss relative to the GT joints. Our method achieves a lower 2D loss than EFT$_{\rm CLIFF}$, with results of 6.1e-3 (our) versus 8.3e-3 (EFT$_{\rm CLIFF}$), indicating that our method is more accurate in approaching the GT joints.

To further evaluate the OOD performance, we train both our method and EFT$_{\rm CLIFF}$ on the COCO dataset (an outdoor dataset) and test them on the Human3.6M dataset (an indoor dataset). With ground truth SMPLs available, we report results using MPJPE and PA-MPJPE metrics. As shown in Table~\ref{tab:ood-h36m}, our method demonstrates superior performance over EFT$_{\rm CLIFF}$, further validating its effectiveness in OOD scenarios.

Figure~\ref{fig:Out of domain} shows two examples from the LSP-Extended dataset. The persons in the two images take complex actions. The shadows in the images and the similar color of the black pants and shoes make it difficult even for humans to identify the configuration of the 3D meshes. Our method successfully estimates correct meshes, while the arms in the results of EFT$_{\rm CLIFF}$ exhibit wrong depth orders.


\section{Conclusion}
To conclude, this paper presents a new training paradigm towards better test-time optimization performance at test time. We mainly propose two strategies. First, we integrate the test-time optimization into the training procedure, which performs test-time optimization before running the typical training in each training iteration. Second, we propose a dual-network architecture to implement the proposed novel training paradigm, aiming at unifying the space of the test-time and training optimization problems. Experiments and comparisons prove that the proposed training scheme improves the effectiveness of the test-time optimization during testing, demonstrating that it successfully learns meta-parameters that benefit the test-time optimization for specific samples. 
Our method can perform even better with stronger regressor baseline or better 2D joints, and can adapt to out-of-domain challenging test cases.

\section*{Acknowledgments}
The work was supported by Guangdong Provincial Natural Science Foundation for Outstanding Youth Team Project (No. 2024B1515040010), the Fundamental Research Funds for the Central Universities (No. 2024ZYGXZR021), China National Key R\&D Program (Grant No. 2023YFE0202700), Key-Area Research and Development Program of Guangzhou City (No.2023B01J0022), National Natural Science Foundation of China for Key Program (No. 62237001), Natural Science Foundation of China (No. 62072191).


\bibliographystyle{splncs04}
\bibliography{ref}

\newpage
\appendix

\section{Appendix}
\subsection{Inference Speed Comparison with Previous Methods}
In Table~\ref{tab:inference-speed}, we provide a comparison with ReFit~\cite{wang2023refit}, NIKI~\cite{li2023niki}, and PLIKS~\cite{shetty2023pliks} in terms of inference speed. Except for NIKI using ResNet-34, all others employ HRNet-W48, and all tested on a single NVIDIA RTX3090 GPU. Ours$_{\rm CLIFF}$ takes 0.072s for an iteration, and about 1.1s for 14 iterations in default. ReFit, NIKI, and PLIKS take 0.043s, 0.068s, 0.041s, respectively, which are faster. When our method performs single-step optimization, the time required is comparable to the above three methods. When we conduct additional iterations of optimization, our method consumes more time. As a reward, the additional optimizations improve human mesh recovery accuracy upon regression approaches.

\begin{table}[h]\scriptsize
\captionsetup{font=normalsize}
\caption{\textbf{Inference speed comparison} with ReFit~\cite{wang2023refit}, NIKI~\cite{li2023niki}, PLIKS~\cite{shetty2023pliks}.}
\centering
\begin{tabular*}{\columnwidth}{p{5cm}<{\centering}p{8cm}<{\centering}}
\toprule
Method  & Inference Speed (per sample)  \\
\midrule
ReFit & 0.043s \\
NIKI &  0.068s\\
PLIKS    &  0.041s  \\
Ours$_{\rm CLIFF}$ & 0.072s (1 iteration) \\
\bottomrule
\end{tabular*}
\label{tab:inference-speed}
\end{table}

\subsection{Ablation on Learning Rate}
In Table~\ref{tab:learning-rate}, we report the experimental results when using different learning rates in our method. We adjusted the learning rates for both the test-time optimization and the ordinary training optimization. Among the feasible learning rates, we observe that utilizing $1 \times 10^{-5}$ for the test-time optimization and $1 \times 10^{-4}$ for the training optimization is the most suitable configuration. We find that our training process is unstable under some combinations of the two learning rates, e.g., when the two learning rates are very different from each other (1e-6 for the test-time optimization and 1e-4 for the training optimization), or using too large learning rates (e.g., 1e-3).

\begin{table}[h]\scriptsize
\captionsetup{font=normalsize}
\caption{\textbf{Ablation study of different learning rate settings}, with COCO \cite{lin2014microsoft} as the training dataset and 3DPW \cite{von2018recovering} as the testing dataset. ``-'' means the training is not stable, and no result is obtained. Gray row is the default setting.}
\centering
\begin{tabular*}{\columnwidth}{p{2.7cm}<{\centering}p{2.7cm}<{\centering}p{2.2cm}<{\centering}p{2.5cm}<{\centering}p{1.8cm}<{\centering}}

\toprule
Test-time\_lr   & Training\_lr    & MPJPE $\downarrow$      & PA-MPJPE $\downarrow$    & PVE $\downarrow$                 \\
\midrule
1e-6 & 1e-4 & - & - & - \\
\rowcolor{gray!20}1e-5 & 1e-4 & \textbf{76.8} & \textbf{49.5} & 90.1 \\
1e-4   &    1e-4  &   77.1& 49.6  &    \textbf{89.1} \\
1e-3 & 1e-4 & - & - & - \\
1e-4 & 1e-3 & - & - & - \\
1e-4    &    1e-5   &  78.8    &   50.8  &  91.0  \\
1e-4 & 1e-6 & 121.6 & 74.0 & 134.5 \\
\bottomrule
\end{tabular*}

\label{tab:learning-rate}
\end{table}

\subsection{Ablation on the Number of Test-time Optimization Steps at Training}

In the main paper, we perform just one step of test-time optimization in each training iteration. Here, we explore using more test-time optimization steps during training, and show its impact on the model performance. The results are shown in Table~\ref{tab:more-test-time-opt-steps}. It can be observed that with the increment in the number of steps, there is a slight improvement in model performance. However, this comes at the cost of increased memory usage and longer training time. To save training time, we opted for a single test-time optimization step at the training stage.

\begin{table}[h]\scriptsize
\captionsetup{font=normalsize}
\caption{\textbf{Ablation study on using different test-time optimization (Opt.) steps at training stage}, with COCO \cite{lin2014microsoft} as the training dataset and 3DPW \cite{von2018recovering} as the testing dataset. Gray row is the default setting. Bold values are the best.}
\tabcolsep=0.05cm
\centering
\begin{tabular*}{\columnwidth}{p{4.5cm}<{\centering}p{1.5cm}<{\centering}p{2.2cm}<{\centering}p{0.9cm}<{\centering}p{4.3cm}<{\centering}}
\toprule
Test-time Opt. Steps at Training  & MPJPE $\downarrow$      & PA-MPJPE $\downarrow$    & PVE $\downarrow$ & Training Time (mins/epoch)    \\
\midrule
\rowcolor{gray!20}1 & 76.6 & 49.5 & 90.1 & \textbf{5.8mins} \\
2 & 76.7 & 49.0 & 90.0 & 6.8mins\\
3 & \textbf{76.3} & \textbf{48.5} & \textbf{89.5} & 7.4mins\\
\bottomrule
\end{tabular*}
\label{tab:more-test-time-opt-steps}
\end{table}

\subsection{Per-Joint Error Analysis}
\begin{figure}[!t]\footnotesize
\captionsetup{font=normalsize}
    \centering  
    \includegraphics[width=0.49\linewidth]{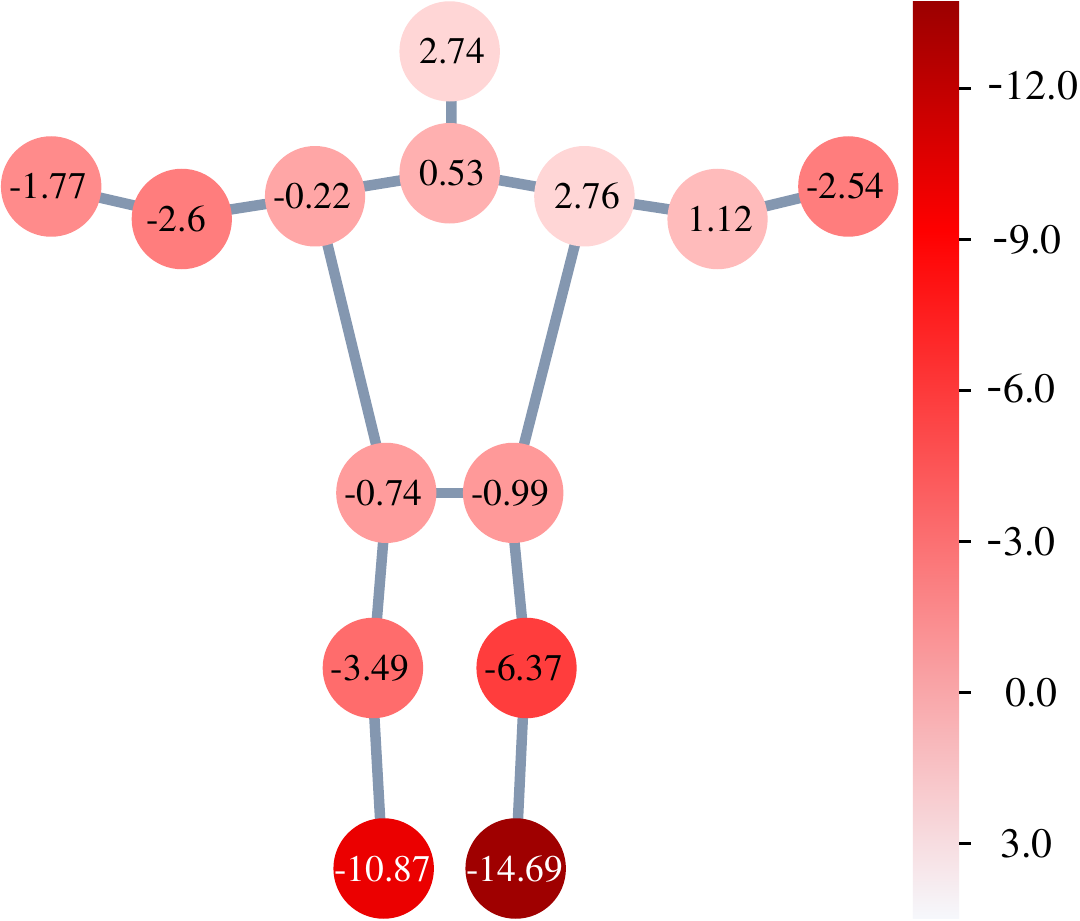}
    \includegraphics[width=0.49\linewidth]{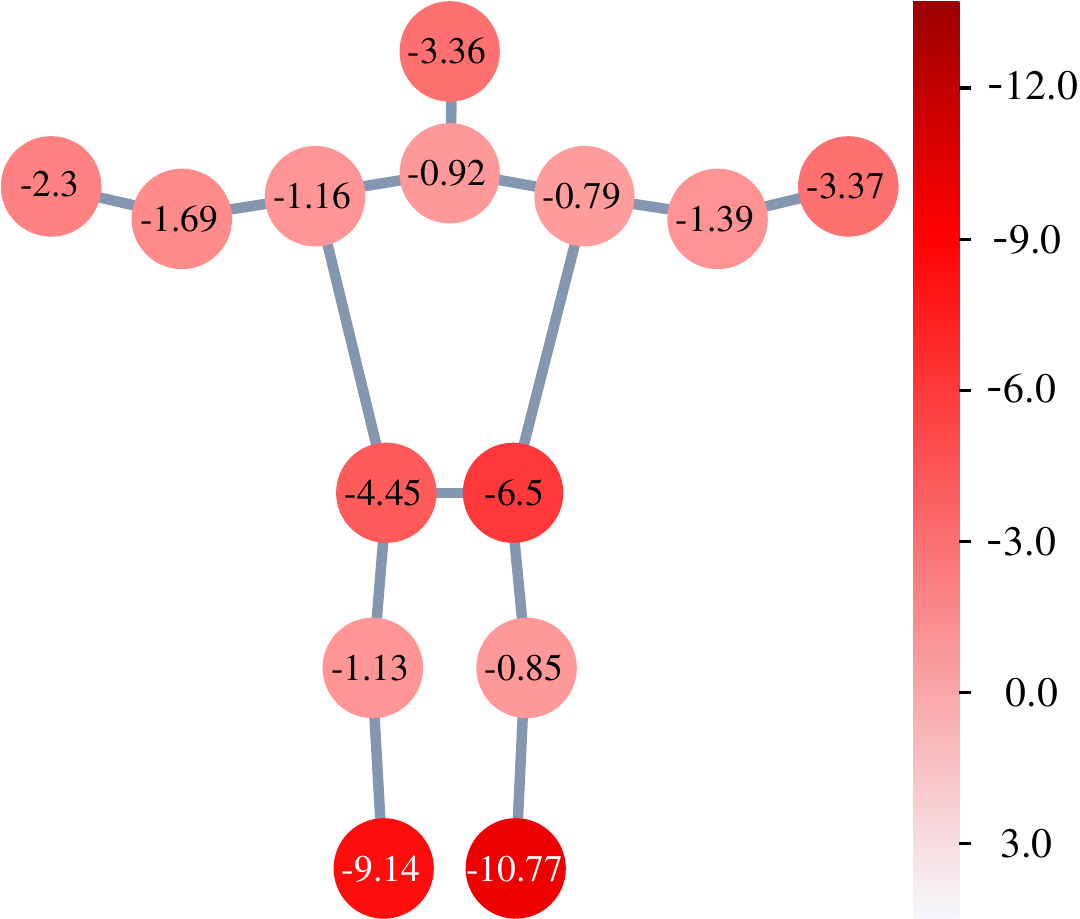} \\
    \hspace{-1.0cm} MPJPE \hspace{5.6cm} PA-MPJPE \\
    \caption{\textbf{Per-joint error analysis} between Our$_{\rm CLIFF}$ and EFT$_{\rm CLIFF}$. The testing dataset is 3DPW \cite{von2018recovering}.}
    \label{fig:error}
\end{figure}

As shown in Figure \ref{fig:error}, we explore the performance gains of Ours$_{\rm CLIFF}$ over EFT$_{\rm CLIFF}$ at each joint. Specifically, we compute the per-joint error of MPJPE and PA-MPJPE, then we subtract the result of EFT$_{\rm CLIFF}$ from our result. Darker red means our method is more better. It can be observed that the joints on feet achieve larger performance gains, primarily due to the higher motion frequency in foot joints. This phenomenon was also observed in the study \cite{dang2022diverse, ma2022progressively, dang2021msr, xiao2025multi, xu2023pose, xiao2022spatial, xu2020gdface}. The advantages of our method are more obvious on PA-MPJPE, \ie, our method outperforms EFT$_{\rm CLIFF}$ in terms of PA-MPJPE for all joints, demonstrating that our method can better capture the pose and shape of the 2D human than EFT$_{\rm CLIFF}$.

\begin{table}[!t]\scriptsize
\captionsetup{font=normalsize}
\caption{\textbf{Ablation study on 3D Pseudo SMPLs.} Models are trained on full training datasets and tested on 3DPW \cite{von2018recovering}.}
\centering
\begin{tabular*}{\columnwidth}{p{4cm}<{\centering}p{3cm}<{\centering} p{3cm}<{\centering} p{2cm}<{\centering}}
\toprule
Method  & MPJPE$\downarrow$ & PA-MPJPE$\downarrow$ & PVE$\downarrow$\\
\midrule
Ours $_{\rm CLIFF}$$\dagger $ (Res-50)  w/o pseudo & 68.7 & 44.7 & 85.9\\
Ours$_{\rm CLIFF}$$\dagger $ (Res-50) w/ pseudo & 66.0 & 42.1 & 83.6 \\
\bottomrule
\end{tabular*}
\label{tab:pseudo-labels}
\end{table}

\subsection{More Explanations about Unifying the Training and Testing Objectives with Dual Networks}
There is a concern that even with the introduction of the auxiliary network, we still can not fully achieve the goal of matching training and testing objectives, i.e., there still remains a discrepancy between the training loss (using ground truth labels) and the testing-time loss (using pseudo labels). We argue that, since the auxiliary network is trained simultaneously with the main network, the auxiliary network learns the pseudo 3D meshes that are most suitable for optimizing the network during test-time optimization. In other words, we use the pseudo 3D labels generated by the auxiliary network to help mitigate the gap between training and testing losses, compared with using the 2D joints only as supervision at the test stage. 

To verify the impact of 3D pseudo SMPLs, we have already conducted an ablation study, as shown in Rows 2 and 3 of Table~\ref{tab:ablation3}. In Row 2, we discard the auxiliary network, meaning the pseudo SMPLs generated by the auxiliary network are not used in the test-time optimization function. The comparison reveals that the inclusion of pseudo SMPLs improves the model's performance, demonstrating their contribution to the optimization process.

Since the above ablation was conducted using a smaller dataset (COCO), we further validate the results by repeating the experiment on the full training datasets. As shown in Table~\ref{tab:pseudo-labels}, these additional findings reinforce the effectiveness of the generated pseudo 3D SMPLs, confirming their positive influence on model performance.


One may concern that using dual networks introduces an ensembling effect, with the second term of Eq. \ref{eq:one-step-update2} serving as an adjustment towards an intermediate estimation between the two networks.

To evaluate the ensembling effect introduced by our dual-network setup, we conducted an experiment using two CLIFF models to generate SMPLs and computed their average SMPL, on which EFT optimization was performed. The final results, including the averaged output, are shown in the Table~\ref{tab:ensembling-effects} as EFT$_{\rm 2CLIFFs}$, demonstrating the impact of using two CLIFFs within EFT.

The results indicate that incorporating two CLIFFs does indeed improve EFT performance, highlighting an ensembling effect. However, even with these improvements, EFT$_{\rm 2CLIFFs}$ underperforms compared to our method. This suggests that the ensembling effect achieved through meta-learning in our approach is more effective than simply using two CLIFF networks.

\begin{table}[!t]\scriptsize
\captionsetup{font=normalsize}
\caption{\textbf{Ablation Study about Ensembling Effects.} Models are trained on COCO \cite{lin2014microsoft} and tested on 3DPW \cite{von2018recovering}.}
\begin{tabular*}{\columnwidth}{p{4cm}<{\centering}p{4cm}<{\centering}p{4cm}<{\centering}}
\toprule
Method & MPJPE$\downarrow$ & PA-MPJPE$\downarrow$ \\
\midrule
EFT$_{\rm CLIFF}$ & 84.6 & 54.2 \\
EFT$_{\rm 2CLIFFs}$ & 82.7 & 53.6 \\
Ours$_{\rm CLIFF}$ & 76.7 & 49.5 \\
\bottomrule
\end{tabular*}
\label{tab:ensembling-effects}
\end{table}



\subsection{Comparison with Kim et al. \cite{kim2022meta} and Analysis}

To provide a quantitative comparison, we collect results from Kim et al. \cite{kim2022meta}. Kim et al. \cite{kim2022meta} use the SPIN backbone—a stronger model than our HMR backbone, and include the LSP dataset in their training data (which we exclude). Our method achieves a better PA-MPJPE score (44.3 vs. 57.88), as shown in Table~\ref{tab:kim-comparison}. This result indicates that our approach surpasses Kim et al.'s performance despite using a relatively simpler backbone and fewer training resources.

We elaborate the key difference between Kim et al. \cite{kim2022meta} and our method in detail.

The key difference is that that Kim et al. \cite{kim2022meta} use the 2D reprojection loss only (please see Eq. 1 in their paper) in both inner and outer loops of meta learning, while our method uses 2D and 3D losses. This is a small difference in formulation, but a large difference in contextualization.

EFT \cite{joo2021exemplar} is performed under the guidance of 2D reprojection loss. Kim et al. \cite{kim2022meta} extended EFT \cite{joo2021exemplar} to both the inner and outer loops of meta learning. In this sense, the method of Kim et al. \cite{kim2022meta} is a direct extension of EFT \cite{joo2021exemplar} to meta learning.

In contrast, our method considers meta learning from the perspective of the complete HMR model trained with 2D and 3D losses. In other words, we use the complete HMR model in both inner and outer loops of meta learning. This is more reasonable, as our aim is to personalize the HMR model on each test sample, rather than a model trained with only 2D reprojection loss.

The above differences enable us to design a dual-network architecture that is very different from the network used in Kim et al. \cite{kim2022meta}. Overall, we find that incorporating 3D SMPLs into the meta learning is very helpful. It is the utilization of the GT SMPLs that greatly improves the results of our method. Besides utilizing GT 3D SMPLs in the outer loop of meta learning, we additionally generate pseudo SMPLs and incorporate them into the inner loop of the metal learning.

\begin{table}[!t]\scriptsize
\captionsetup{font=normalsize}
\caption{\textbf{Quantitative Comparison with Kim et al. \cite{kim2022meta}.}}
\begin{tabular*}{\columnwidth}{p{1.5cm}<{\centering}p{1.4cm}<{\centering}p{5.8cm}<{\centering} p{1.8cm}<{\centering} p{1.5cm}<{\centering}} 
\toprule
Method & Backbone & Training Dataset & Testing Dataset & PA-MPJPE$\downarrow$ \\ \midrule
Kim et al. \cite{kim2022meta} & SPIN \cite{kolotouros2019learning}& COCO, MPII, Human3.6M, MPI-INF-3DHP, 3DPW, LSP & 3DPW & 57.88 \\ 
Ours & HMR \cite{kanazawa2018end} & COCO, MPII, Human3.6M, MPI-INF-3DHP, 3DPW & 3DPW & 44.3 \\ 
\bottomrule 
\end{tabular*} 
\label{tab:kim-comparison} 
\end{table}

\subsection{More Intermediate Results}

\begin{figure*}[h]
    \centering  
    \begin{minipage}{1.0\textwidth}
        \centering
        \includegraphics[width=0.118\linewidth]{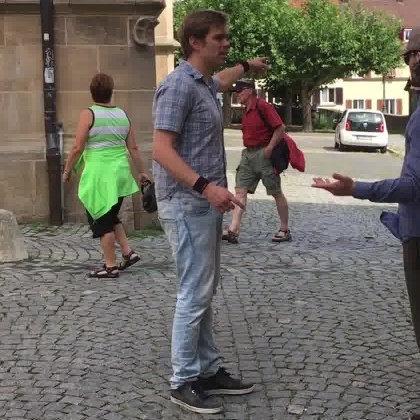}
        \includegraphics[width=0.118\linewidth]{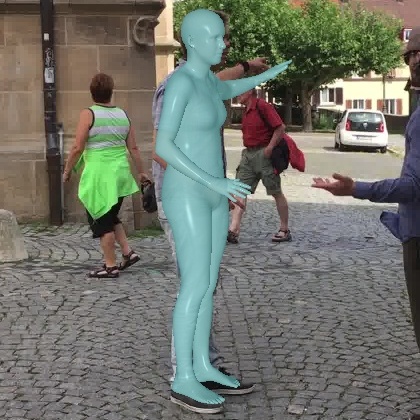}
        \includegraphics[width=0.118\linewidth]{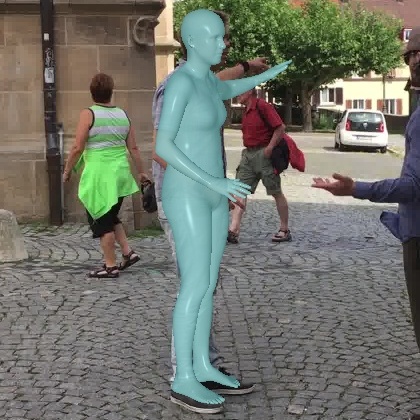}
        \includegraphics[width=0.118\linewidth]{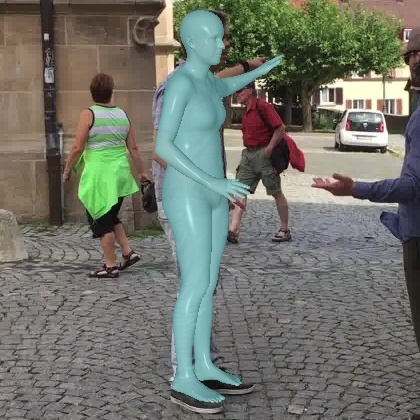}
        \includegraphics[width=0.118\linewidth]{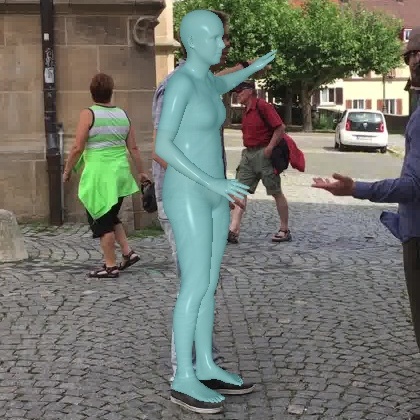}
        \includegraphics[width=0.118\linewidth]{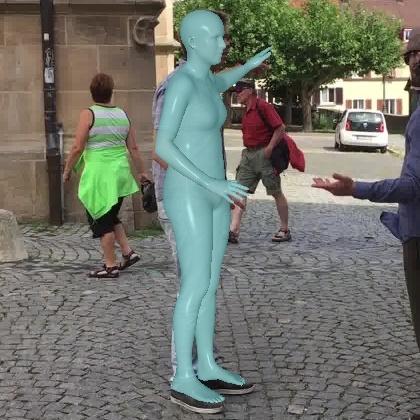}
        \includegraphics[width=0.118\linewidth]{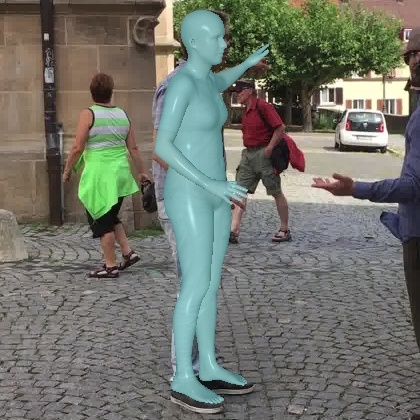}
        \includegraphics[width=0.118\linewidth]{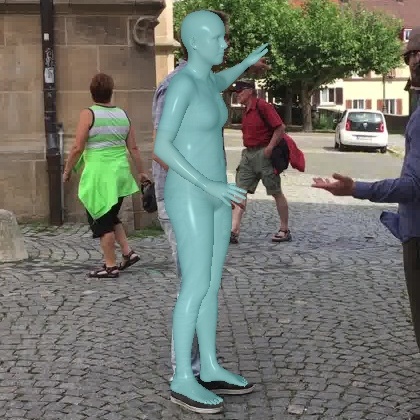}\\
        \includegraphics[width=0.118\linewidth]{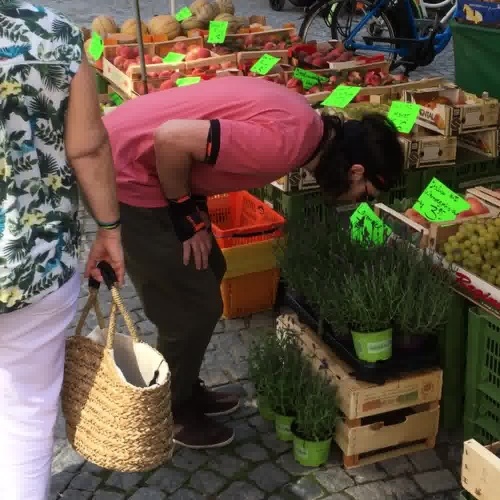}
        \includegraphics[width=0.118\linewidth]{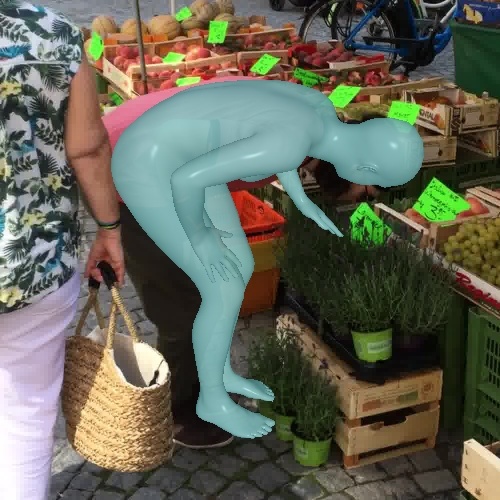}
        \includegraphics[width=0.118\linewidth]{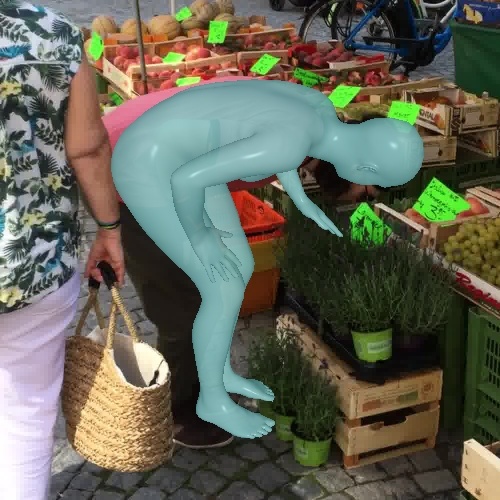}
        \includegraphics[width=0.118\linewidth]{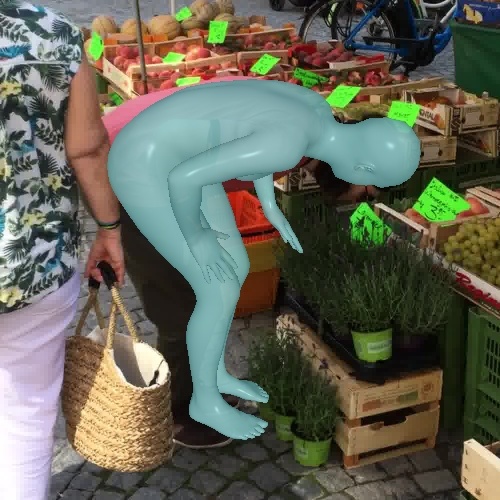}
        \includegraphics[width=0.118\linewidth]{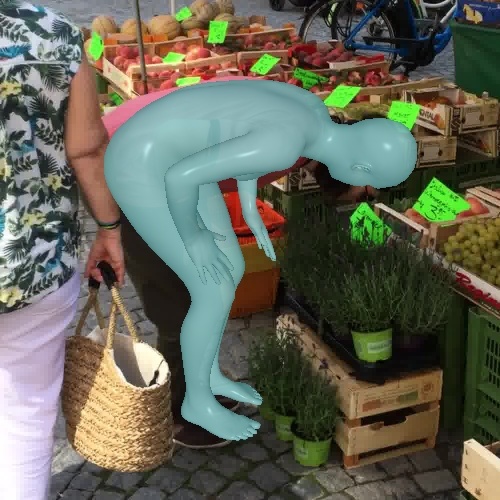}
        \includegraphics[width=0.118\linewidth]{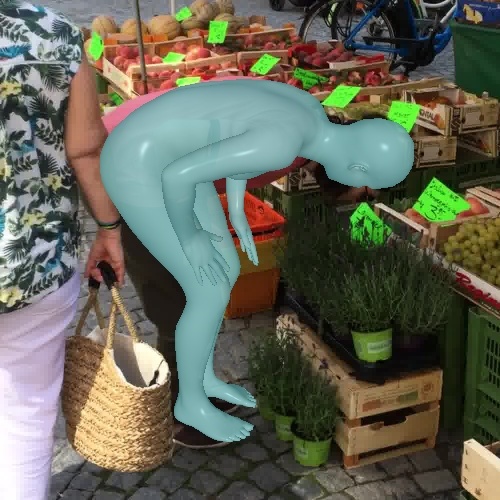}
        \includegraphics[width=0.118\linewidth]{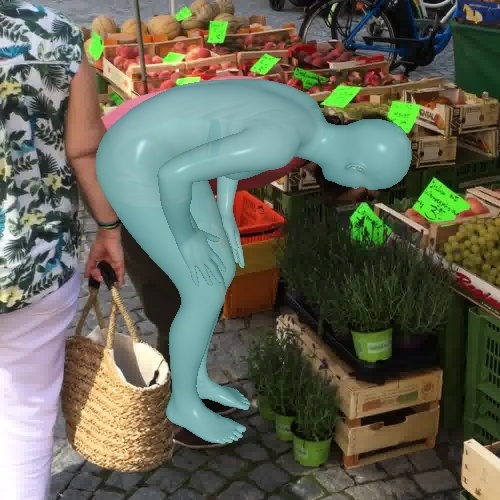}
        \includegraphics[width=0.118\linewidth]{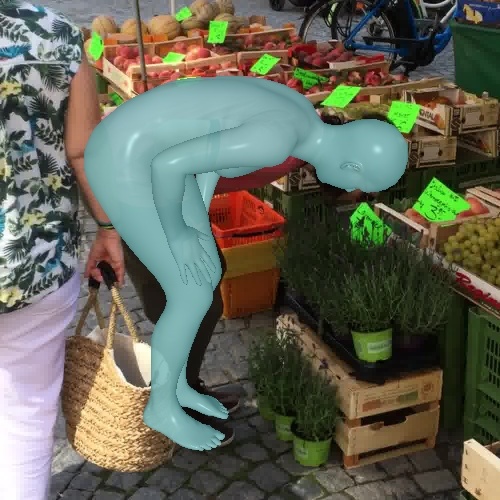}\\
        \includegraphics[width=0.118\linewidth]{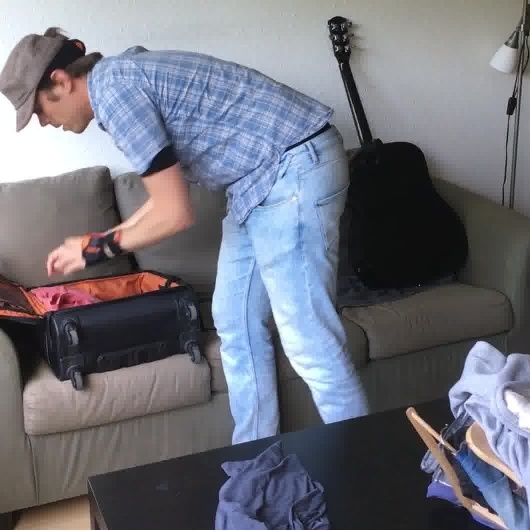}
        \includegraphics[width=0.118\linewidth]{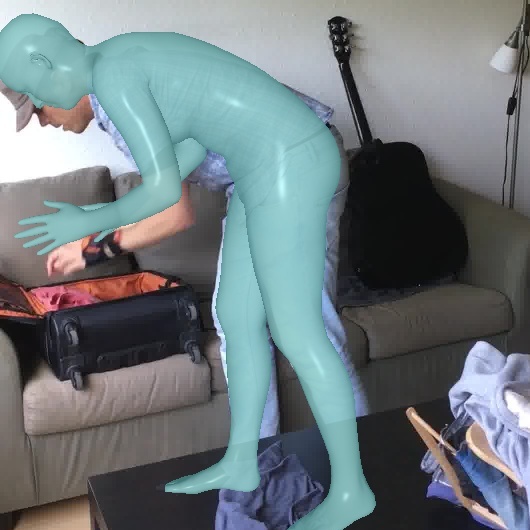}
        \includegraphics[width=0.118\linewidth]{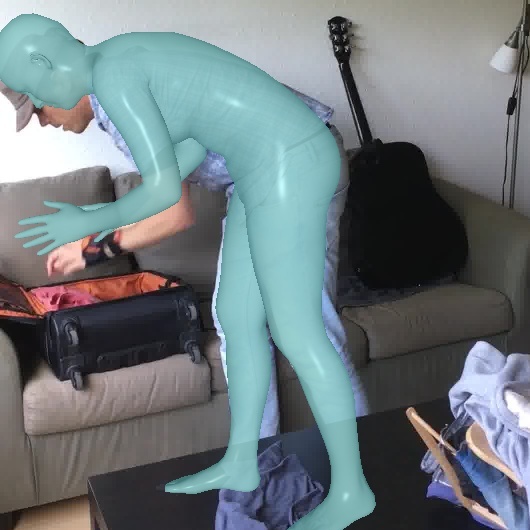}
        \includegraphics[width=0.118\linewidth]{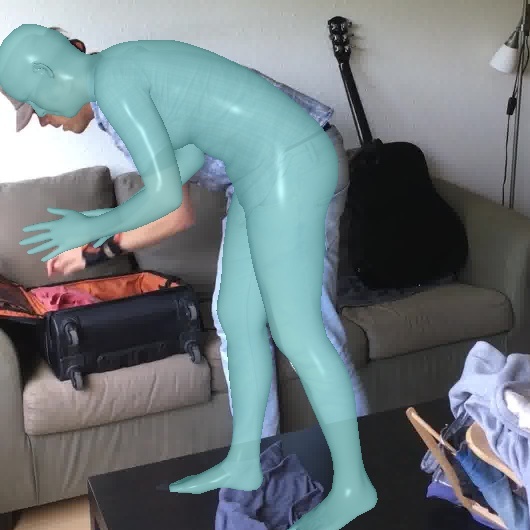}
        \includegraphics[width=0.118\linewidth]{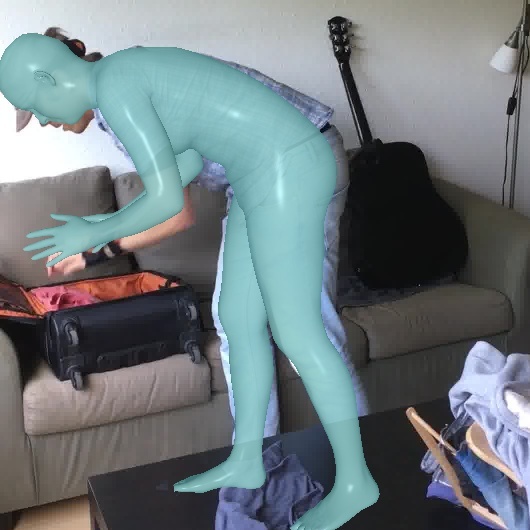}
        \includegraphics[width=0.118\linewidth]{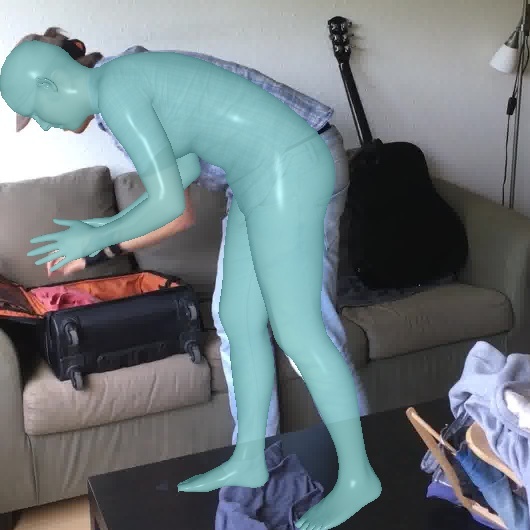}
        \includegraphics[width=0.118\linewidth]{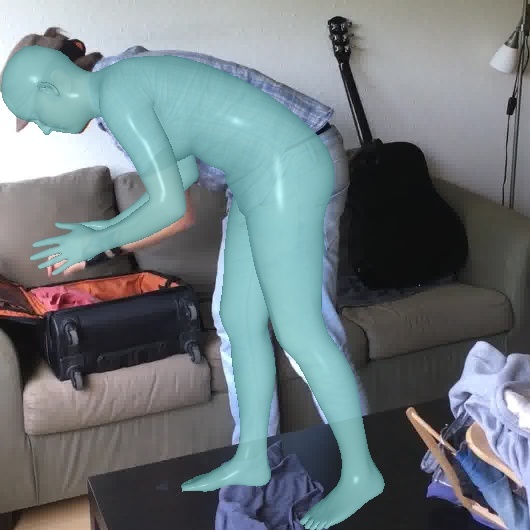}
        \includegraphics[width=0.118\linewidth]{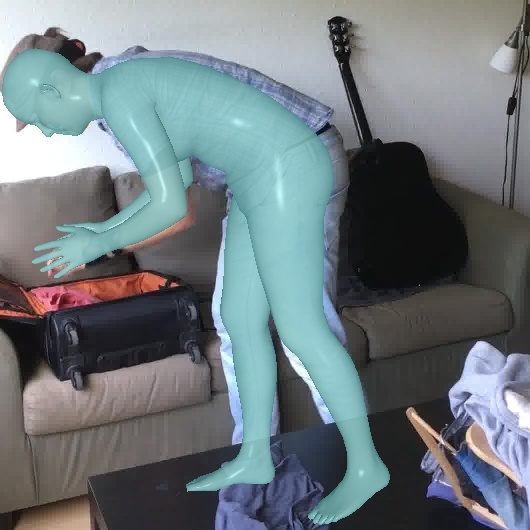}\\
         \includegraphics[width=0.118\linewidth]{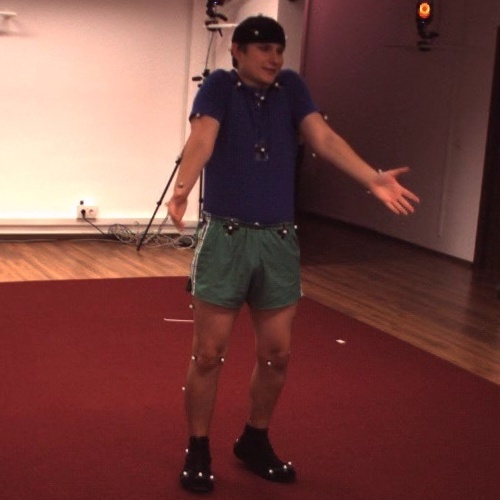}
        \includegraphics[width=0.118\linewidth]{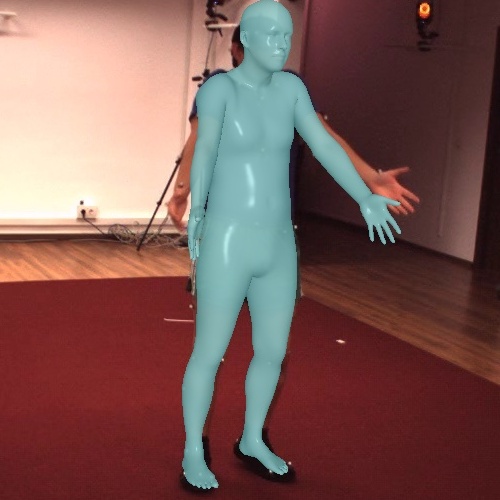}
        \includegraphics[width=0.118\linewidth]{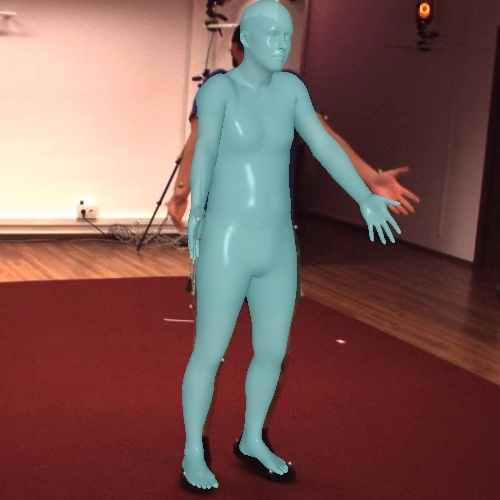}
        \includegraphics[width=0.118\linewidth]{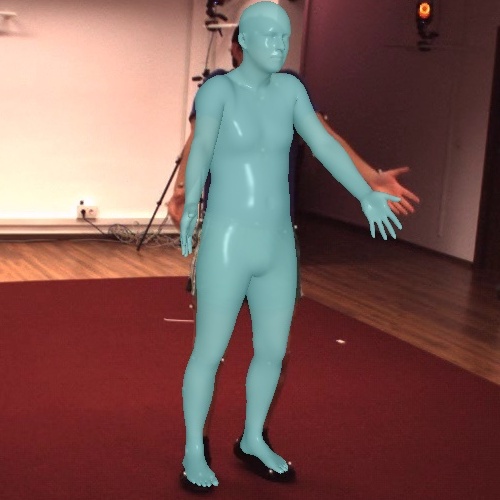}
        \includegraphics[width=0.118\linewidth]{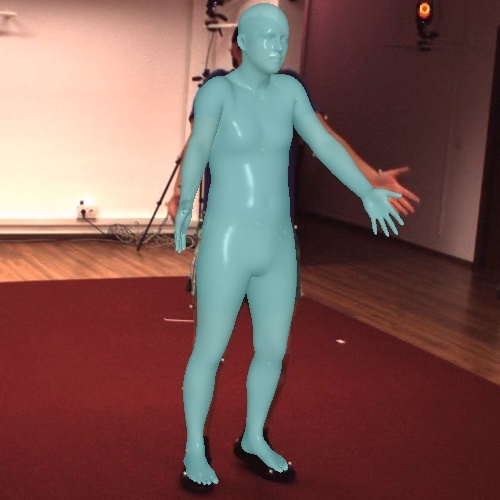}
        \includegraphics[width=0.118\linewidth]{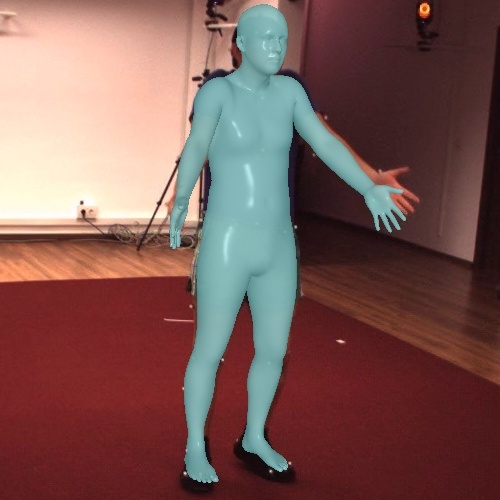}
        \includegraphics[width=0.118\linewidth]{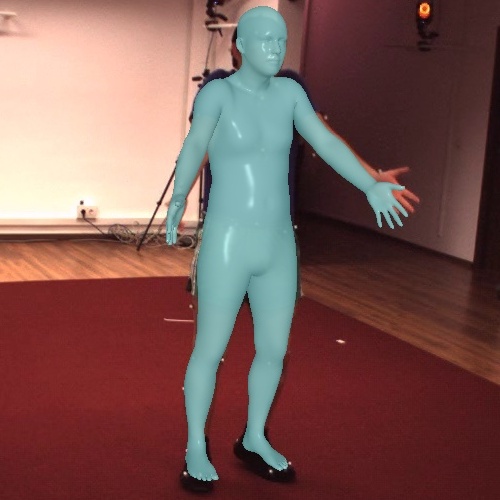}
        \includegraphics[width=0.118\linewidth]{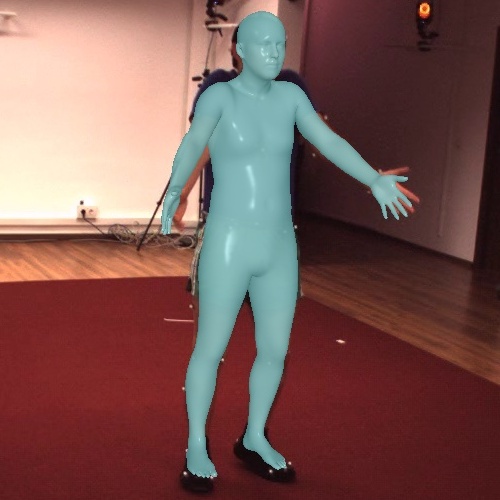}\\
         \includegraphics[width=0.118\linewidth]{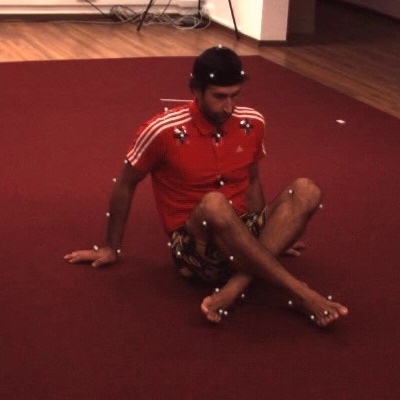}
        \includegraphics[width=0.118\linewidth]{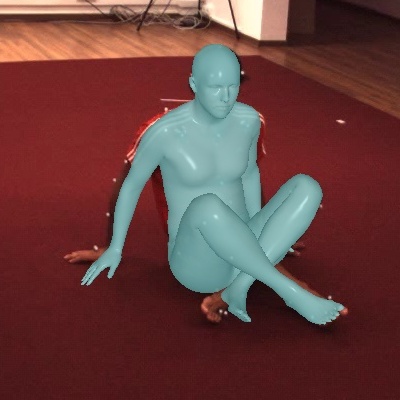}
        \includegraphics[width=0.118\linewidth]{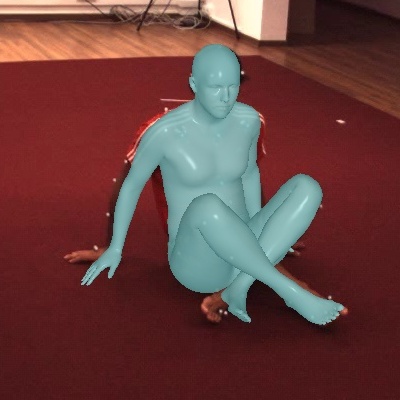}
        \includegraphics[width=0.118\linewidth]{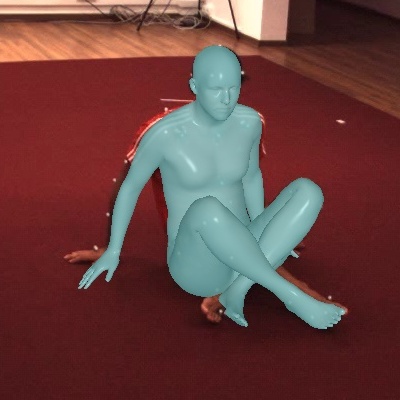}
        \includegraphics[width=0.118\linewidth]{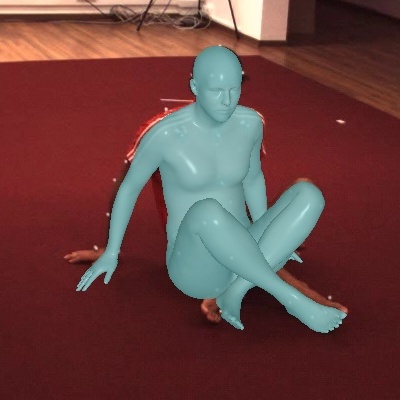}
        \includegraphics[width=0.118\linewidth]{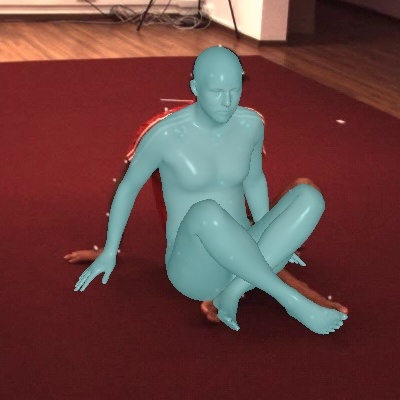}
        \includegraphics[width=0.118\linewidth]{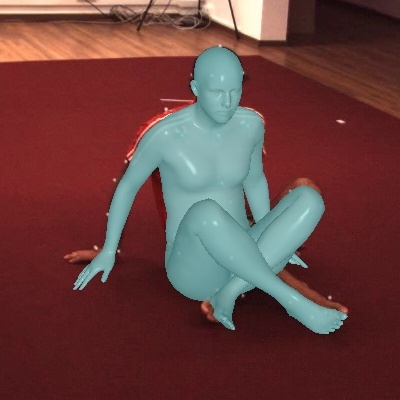}
        \includegraphics[width=0.118\linewidth]{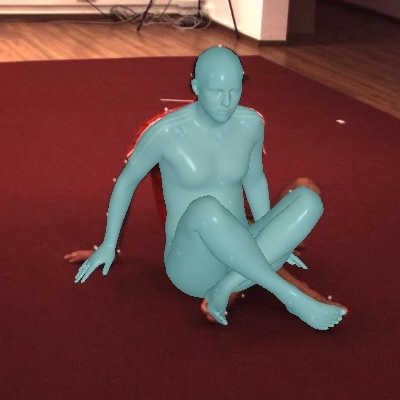}\\
        \includegraphics[width=0.118\linewidth]{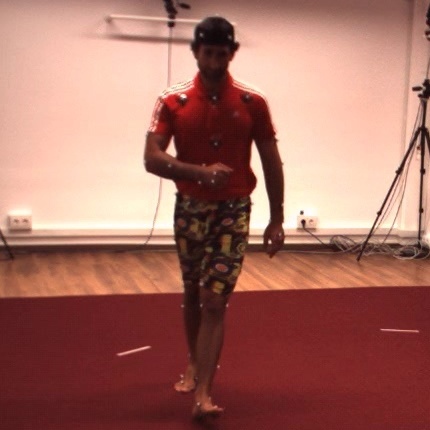}
        \includegraphics[width=0.118\linewidth]{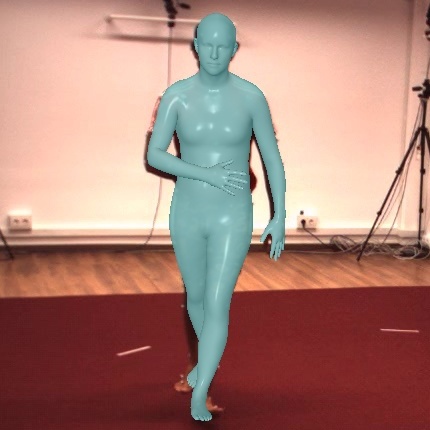}
        \includegraphics[width=0.118\linewidth]{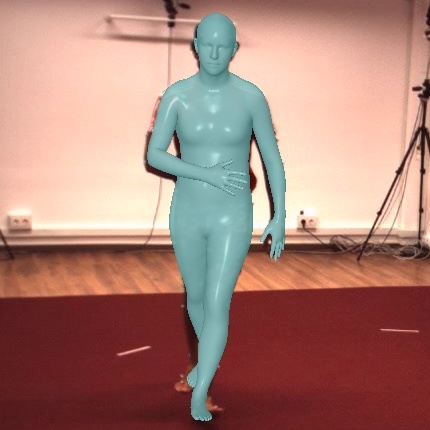}
        \includegraphics[width=0.118\linewidth]{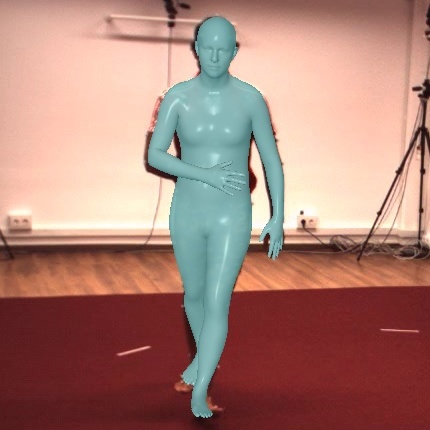}
        \includegraphics[width=0.118\linewidth]{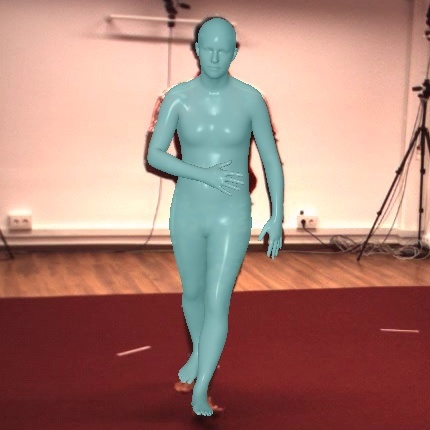}
        \includegraphics[width=0.118\linewidth]{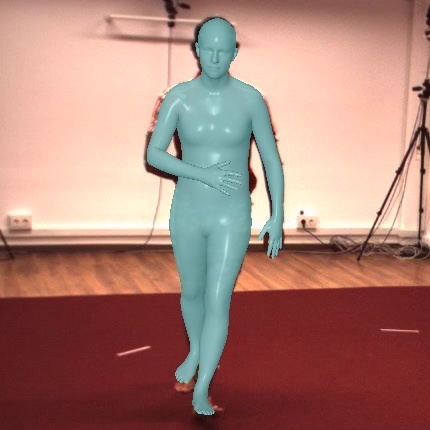}
        \includegraphics[width=0.118\linewidth]{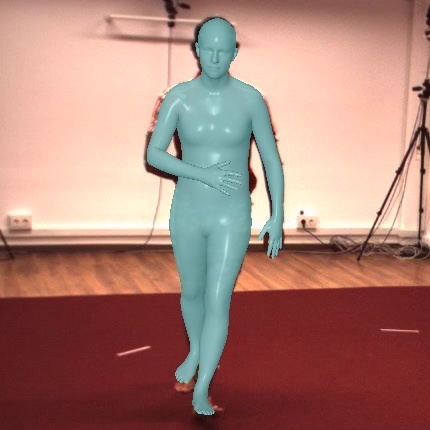}
        \includegraphics[width=0.118\linewidth]{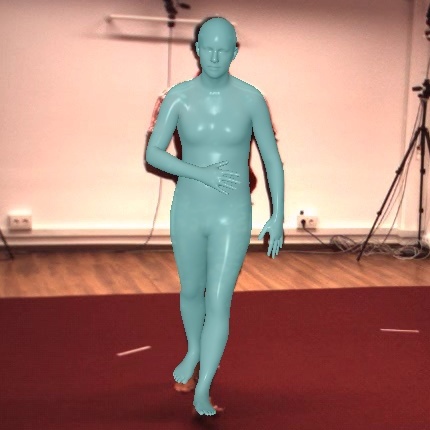}\\
        \includegraphics[width=0.118\linewidth]{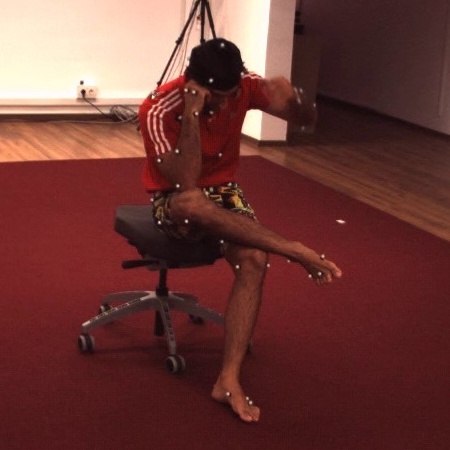}
        \includegraphics[width=0.118\linewidth]{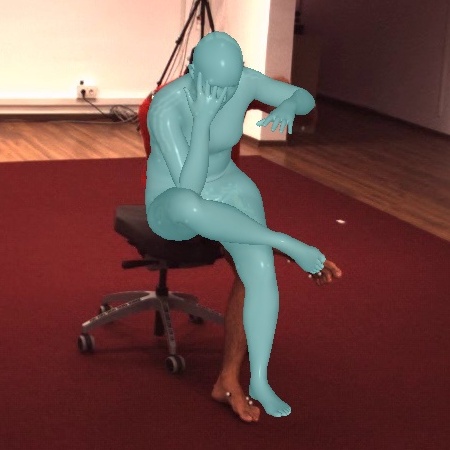}
        \includegraphics[width=0.118\linewidth]{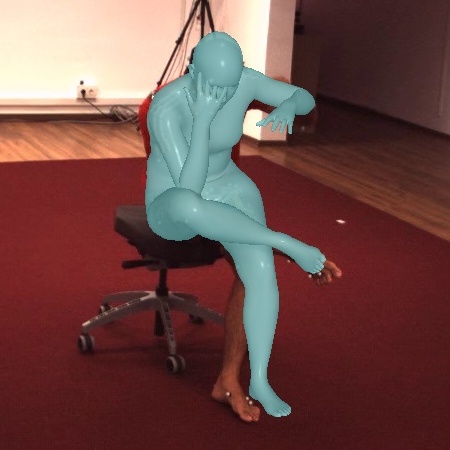}
        \includegraphics[width=0.118\linewidth]{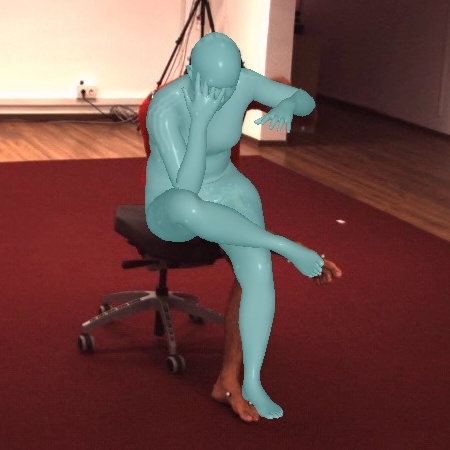}
        \includegraphics[width=0.118\linewidth]{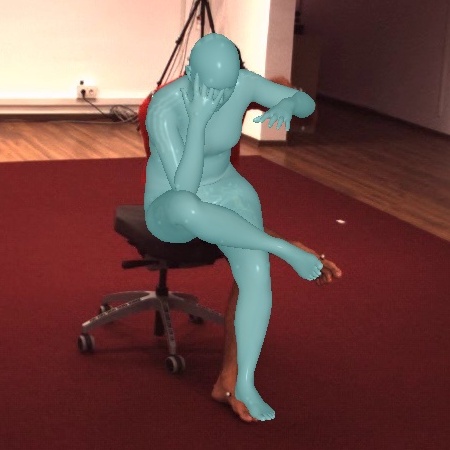}
        \includegraphics[width=0.118\linewidth]{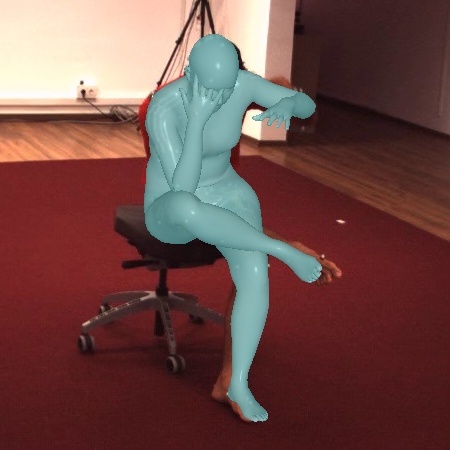}
        \includegraphics[width=0.118\linewidth]{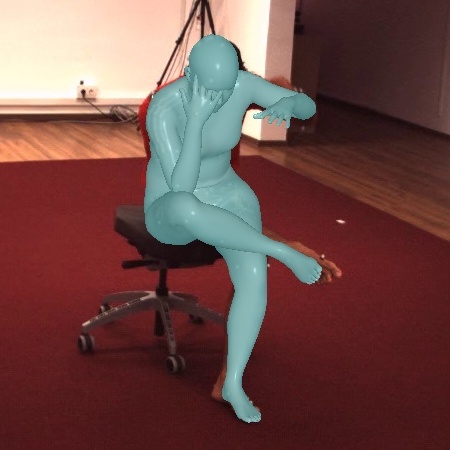}
        \includegraphics[width=0.118\linewidth]{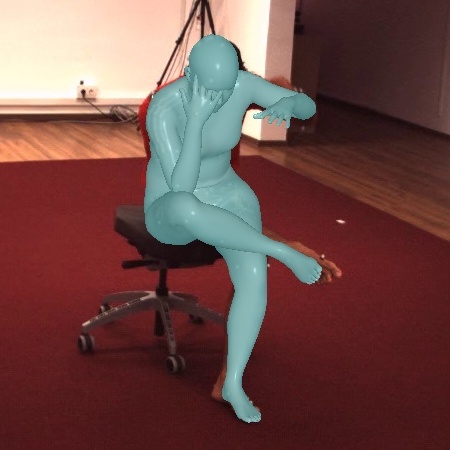}\\
        \includegraphics[width=0.118\linewidth]{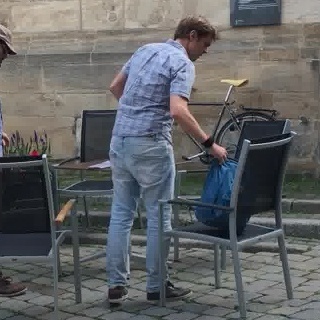}
        \includegraphics[width=0.118\linewidth]{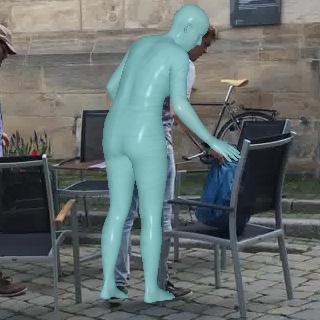}
        \includegraphics[width=0.118\linewidth]{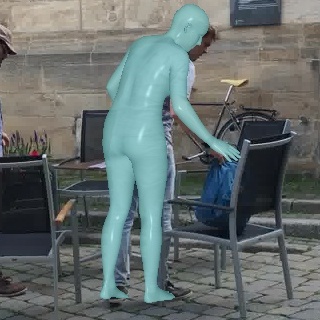}
        \includegraphics[width=0.118\linewidth]{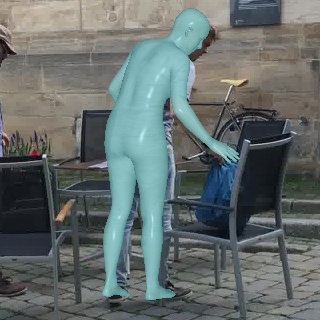}
        \includegraphics[width=0.118\linewidth]{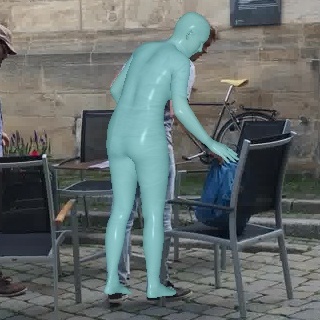}
        \includegraphics[width=0.118\linewidth]{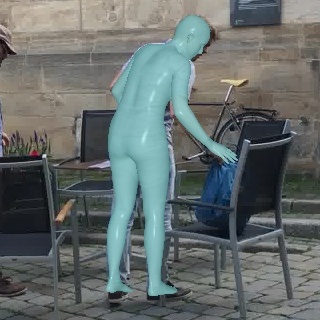}
        \includegraphics[width=0.118\linewidth]{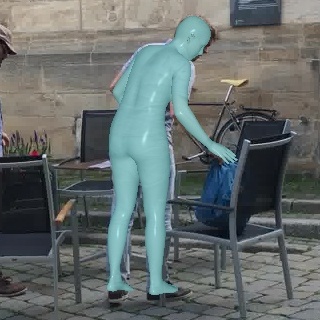}
        \includegraphics[width=0.118\linewidth]{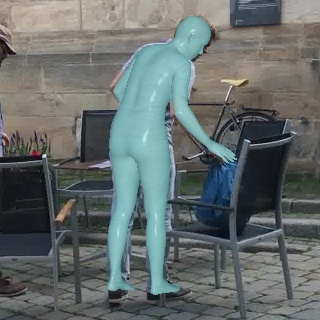}\\
        \includegraphics[width=0.118\linewidth]{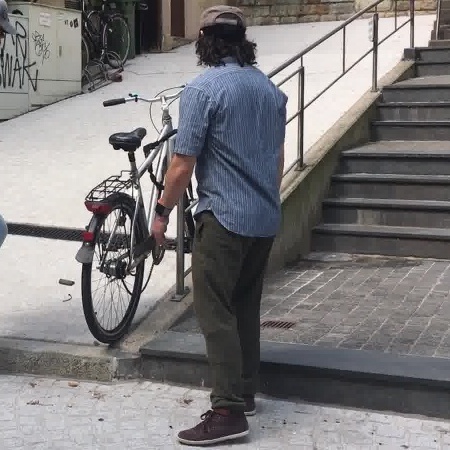}
        \includegraphics[width=0.118\linewidth]{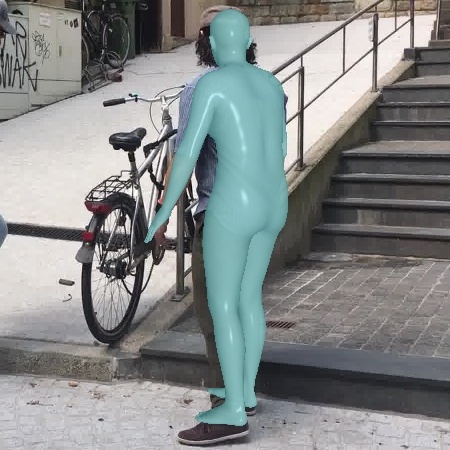}
        \includegraphics[width=0.118\linewidth]{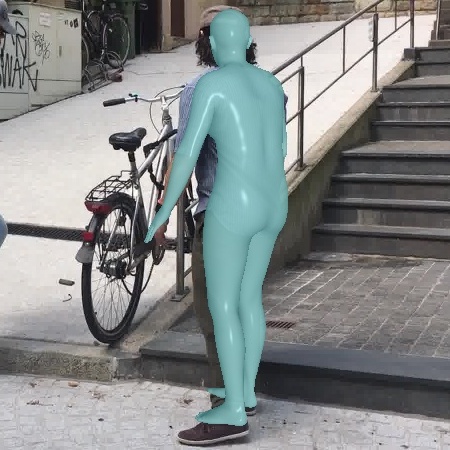}
        \includegraphics[width=0.118\linewidth]{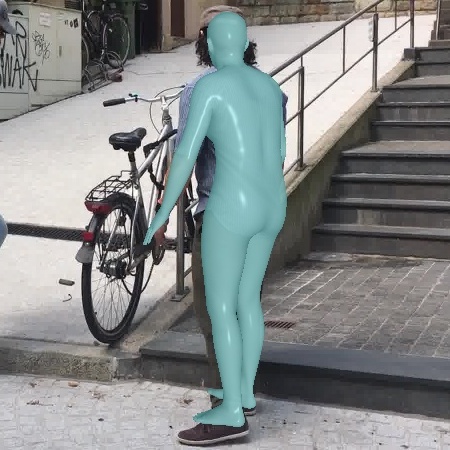}
        \includegraphics[width=0.118\linewidth]{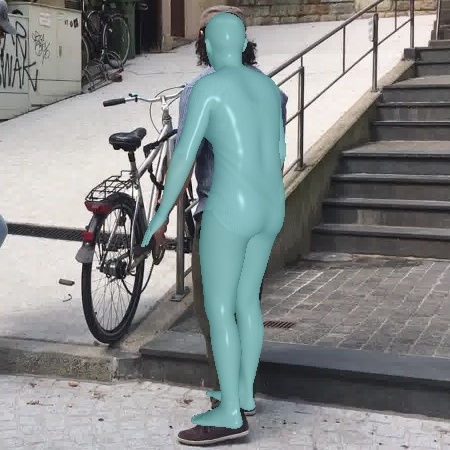}
        \includegraphics[width=0.118\linewidth]{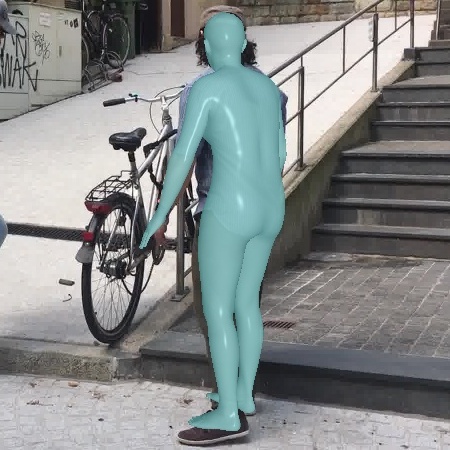}
        \includegraphics[width=0.118\linewidth]{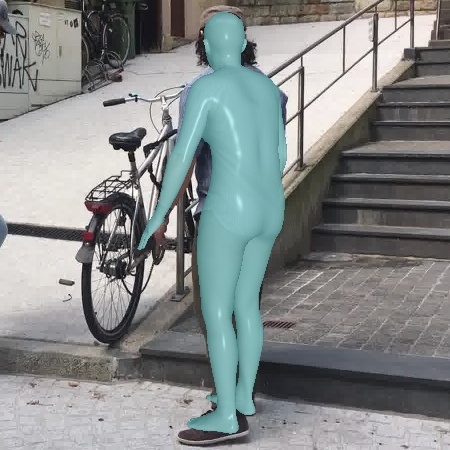}
        \includegraphics[width=0.118\linewidth]{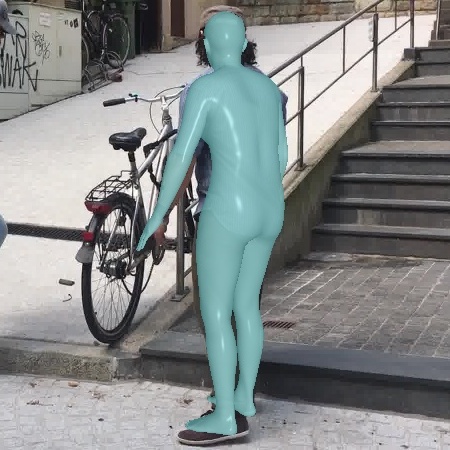}\\
        \includegraphics[width=0.118\linewidth]{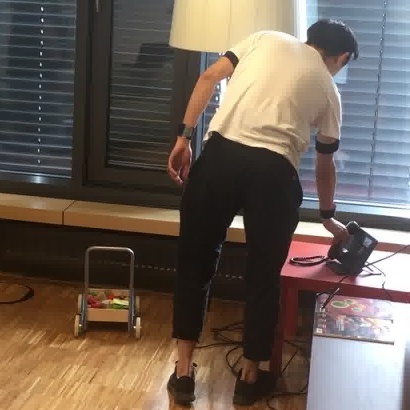}
        \includegraphics[width=0.118\linewidth]{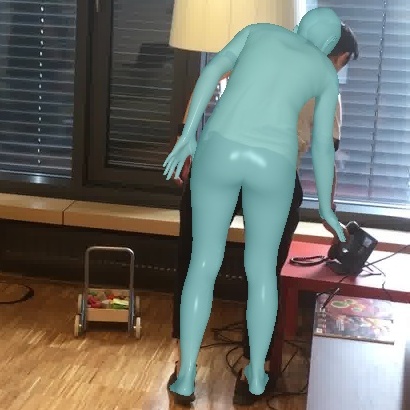}
        \includegraphics[width=0.118\linewidth]{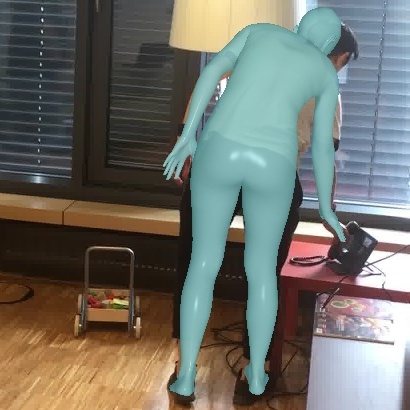}
        \includegraphics[width=0.118\linewidth]{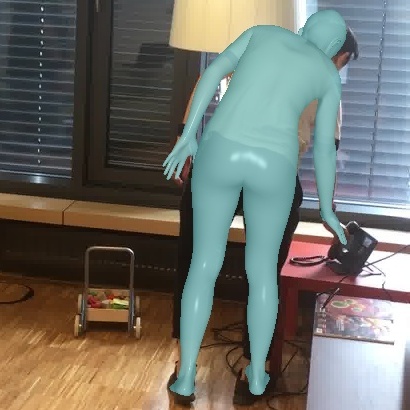}
        \includegraphics[width=0.118\linewidth]{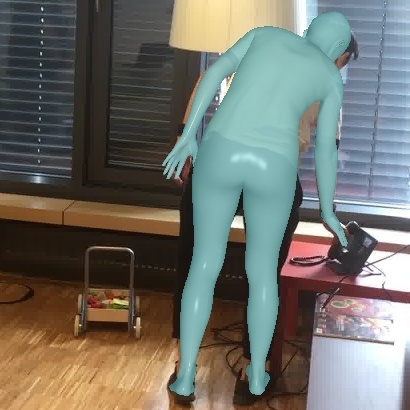}
        \includegraphics[width=0.118\linewidth]{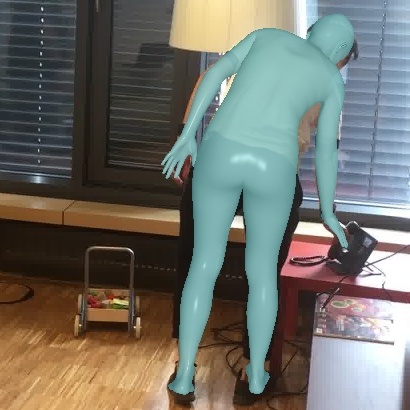}
        \includegraphics[width=0.118\linewidth]{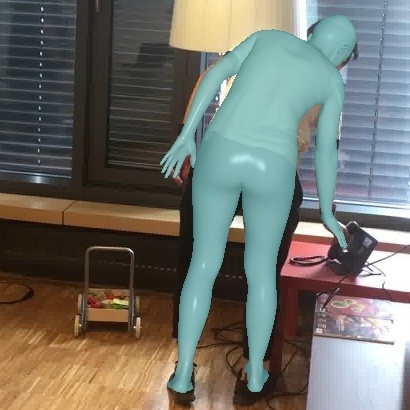}
        \includegraphics[width=0.118\linewidth]{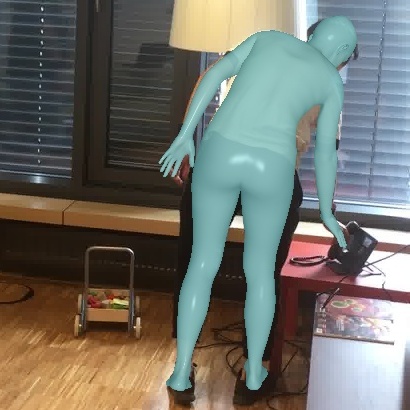}\\
    \end{minipage}
    \begin{minipage}{1.0\textwidth}
        \centering
        \begin{minipage}[t]{0.118\linewidth}
        \caption*{Input}
        \end{minipage}
        \begin{minipage}[t]{0.118\linewidth}
        \caption*{Step 0}
        \end{minipage}
        \begin{minipage}[t]{0.118\linewidth}
        \caption*{Step 1}
        \end{minipage}
        \begin{minipage}[t]{0.118\linewidth}
        \caption*{Step 2}
        \end{minipage}
        \begin{minipage}[t]{0.118\linewidth}
        \caption*{Step 3}
        \end{minipage}
        \begin{minipage}[t]{0.118\linewidth}
        \caption*{Step 4}
        \end{minipage}
        \begin{minipage}[t]{0.118\linewidth}
        \caption*{Step 5}
        \end{minipage}
        \begin{minipage}[t]{0.118\linewidth}
        \caption*{Final}
        \end{minipage}
    \end{minipage}
    \vspace{-0.5cm}
    \caption{\textbf{Stepwise visualization.} From left to right, we showcase results after different steps of test-time optimization during testing.}
    \label{fig:deform_sup}
\end{figure*}

In Figure~\ref{fig:deform_sup}, we show more stepwise optimization results during testing. It can be seen from these examples that the mesh deviates more or less from the target human after step 0, and the deviation is progressively repaired after several optimization steps. For example, please check the results in row 1. The left arm does not match with the evidence in the image at the very beginning, and this error is corrected after about 4 optimization steps. We show more examples from row 2 to 5 where the arms are wrong at first and corrected afterwards. In row 6 and 7, please pay attention to the legs. In the last three rows, please pay attention to the whole bodies.

\subsection{More Qualitative Results}

\begin{figure*}[!t]
    \centering  
    \begin{minipage}{\textwidth}
        \centering
        \newlength{\customimgwidth}
        \setlength{\customimgwidth}{0.1628\textwidth} 
        \rotatebox{90}{~~~~~~~~Input}\hspace{0.3cm}
        \includegraphics[width=\customimgwidth]{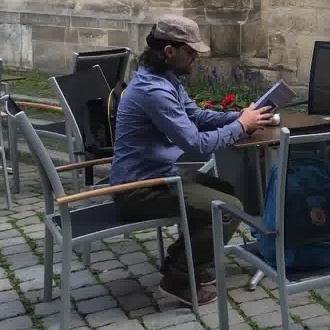}
        \includegraphics[width=\customimgwidth]
        {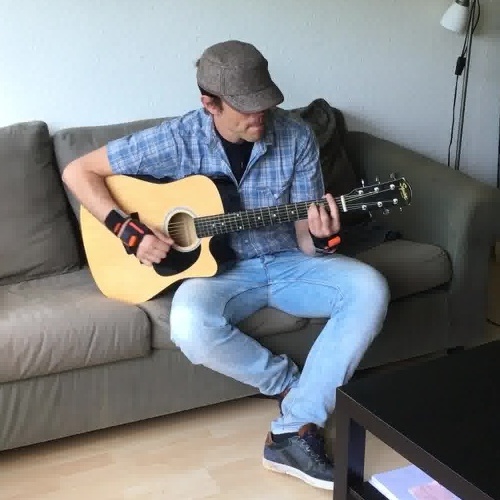}
        \includegraphics[width=\customimgwidth]{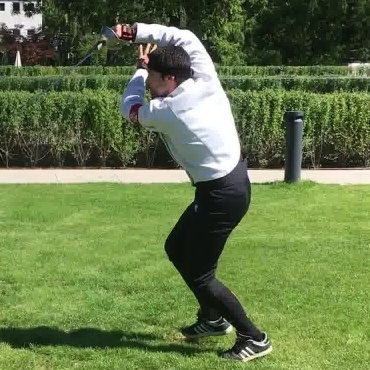}
        \includegraphics[width=\customimgwidth]{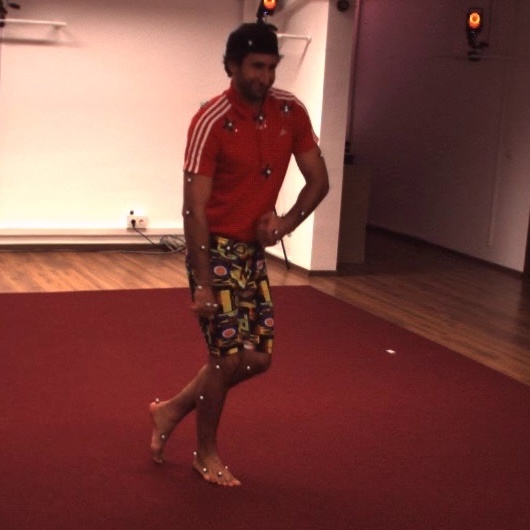}
        \includegraphics[width=\customimgwidth]{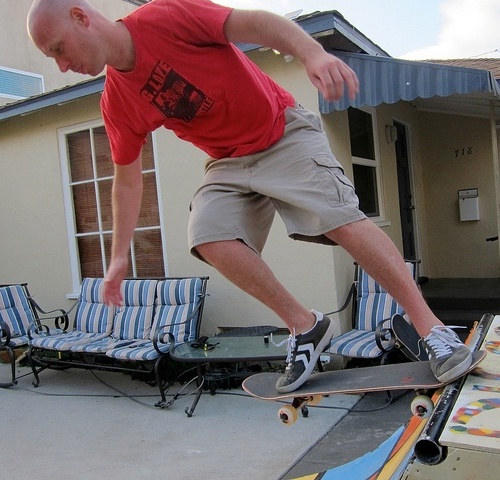}\\
        \rotatebox{90}{~~~HybrIK \cite{li2021hybrik}}\hspace{0.3cm}
        \includegraphics[width=\customimgwidth]{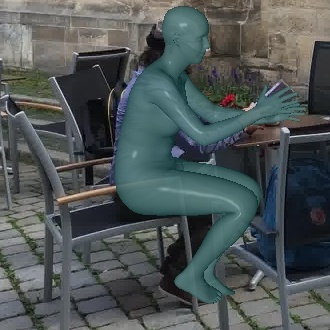}
        \includegraphics[width=\customimgwidth]{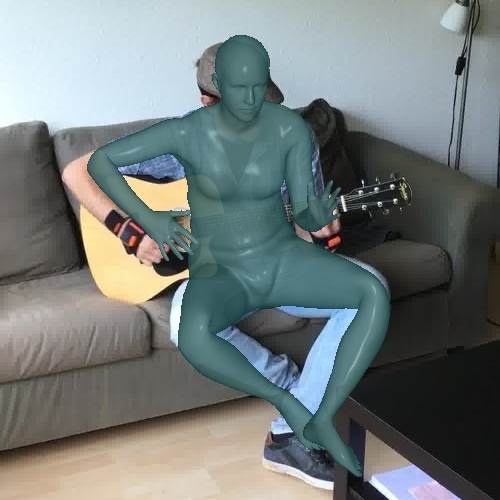}
        \includegraphics[width=\customimgwidth]{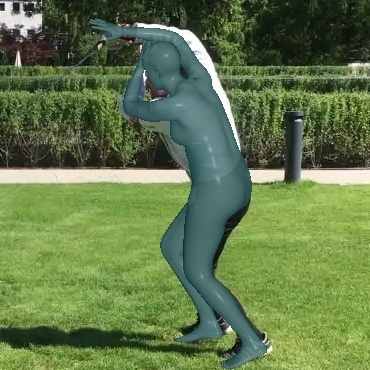}
        \includegraphics[width=\customimgwidth]{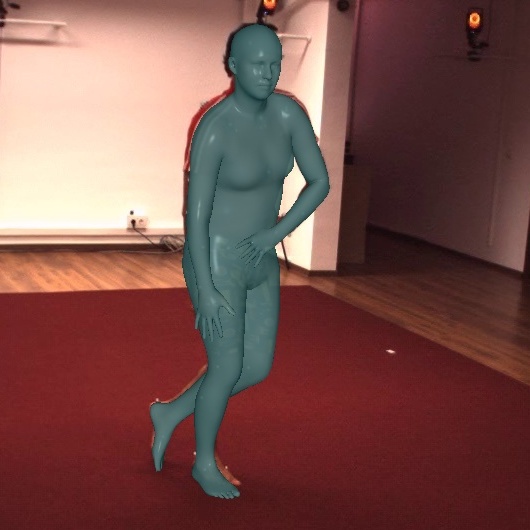}
        \includegraphics[width=\customimgwidth]{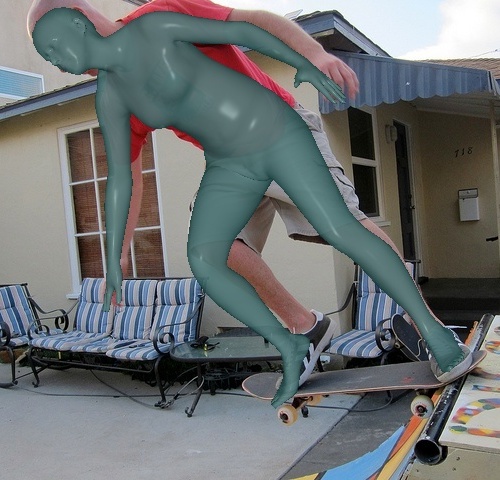}\\
        \rotatebox{90}{~~~~~NIKI \cite{li2023niki}}\hspace{0.3cm}
        \includegraphics[width=\customimgwidth]{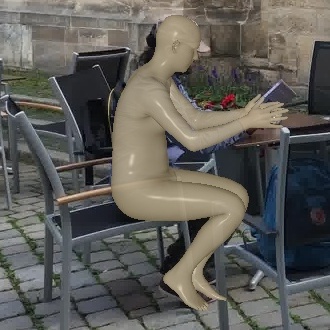}
        \includegraphics[width=\customimgwidth]{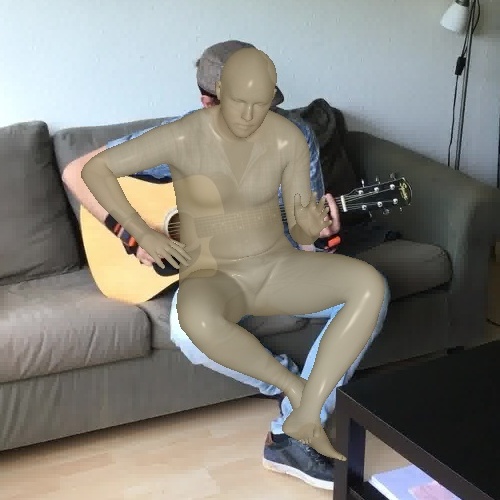}
        \includegraphics[width=\customimgwidth]{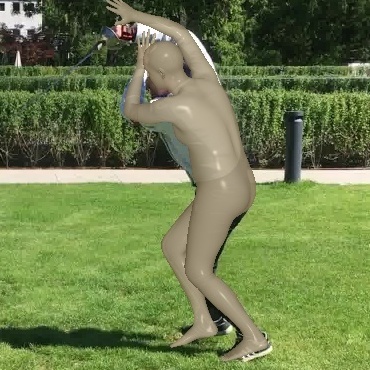}
        \includegraphics[width=\customimgwidth]{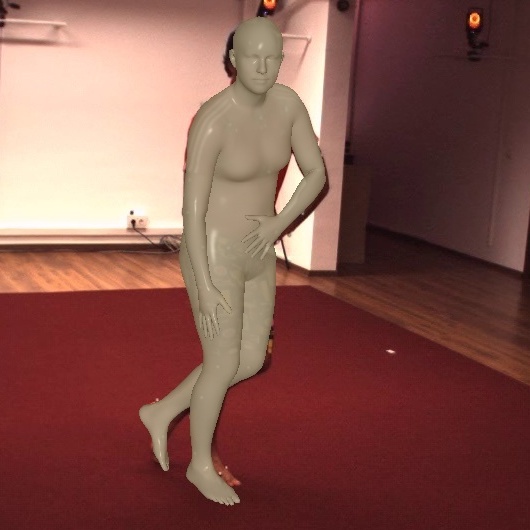}
        \includegraphics[width=\customimgwidth]{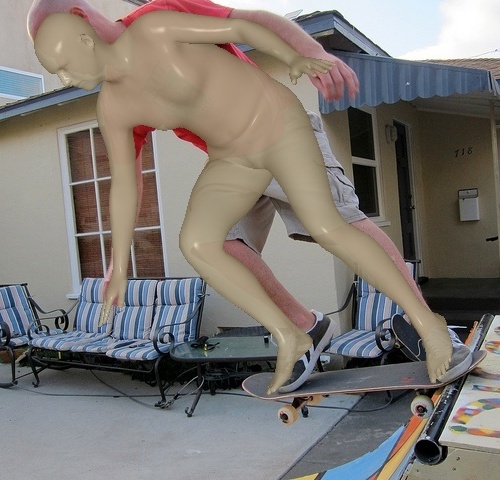}\\
        \rotatebox{90}{~~ProPose \cite{fang2023learning}}\hspace{0.3cm}
        \includegraphics[width=\customimgwidth]{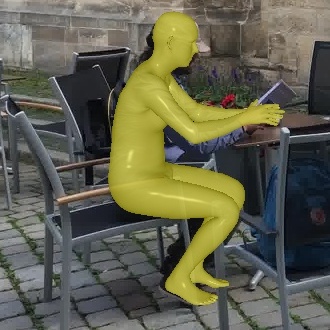}
        \includegraphics[width=\customimgwidth]
        {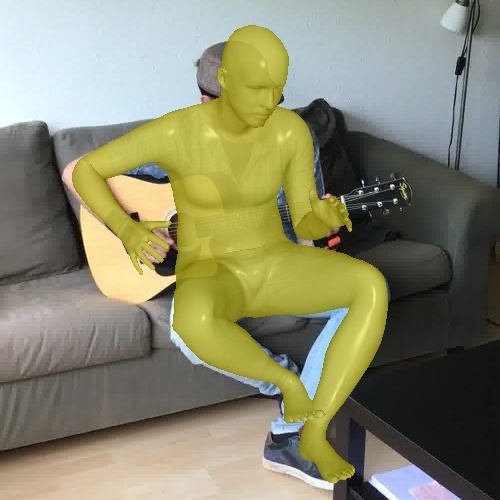}
        \includegraphics[width=\customimgwidth]{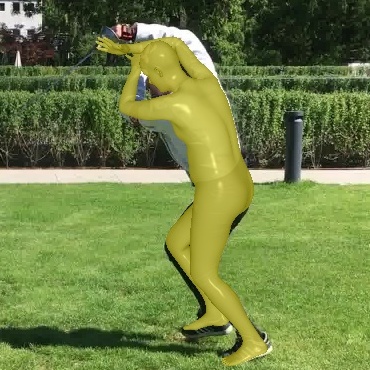}
        \includegraphics[width=\customimgwidth]{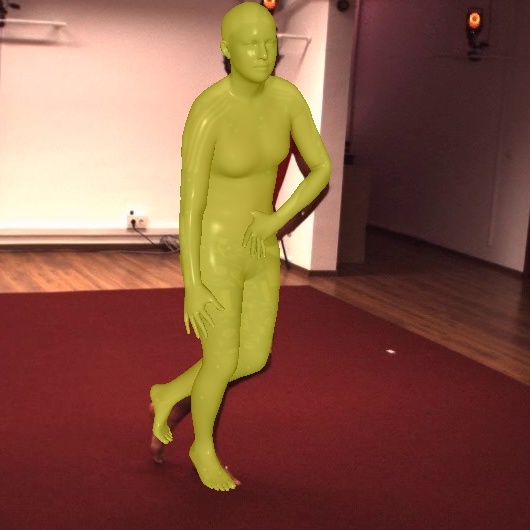}
        \includegraphics[width=\customimgwidth]{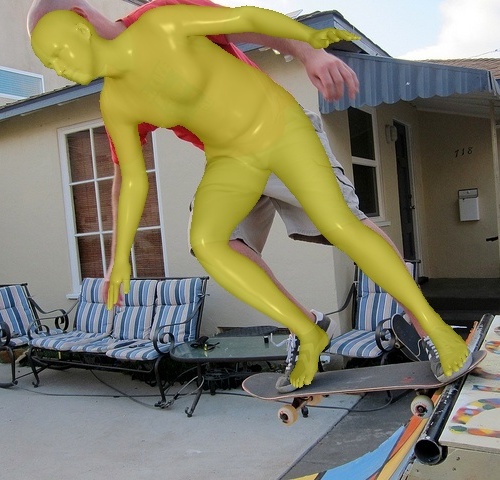}\\
        \rotatebox{90}{~~~~ReFit \cite{wang2023refit}}\hspace{0.3cm}
        \includegraphics[width=\customimgwidth]{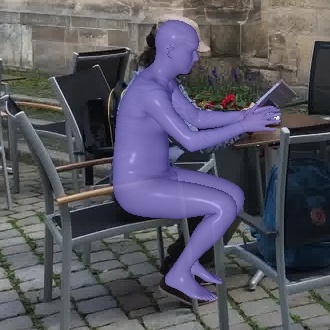}
        \includegraphics[width=\customimgwidth]{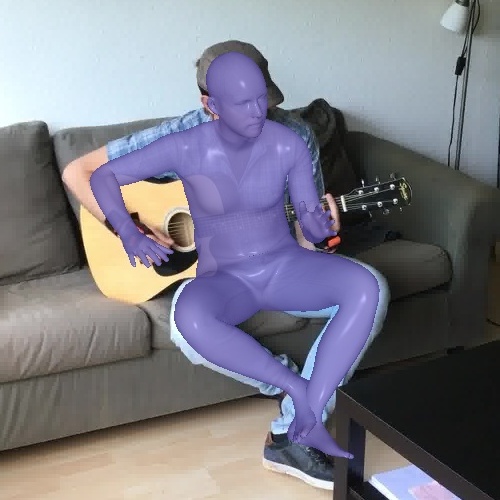}
        \includegraphics[width=\customimgwidth]{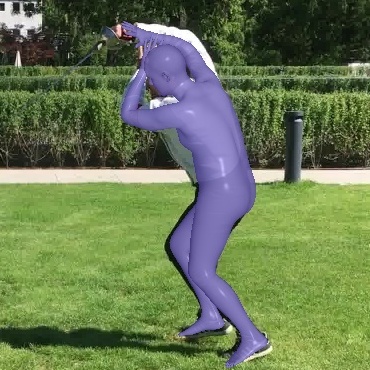}
        \includegraphics[width=\customimgwidth]{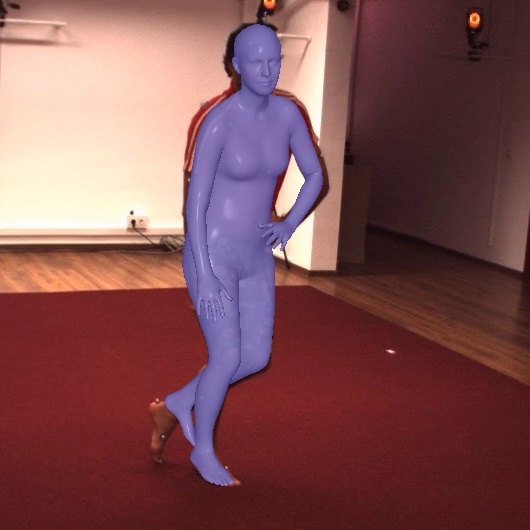}
        \includegraphics[width=\customimgwidth]{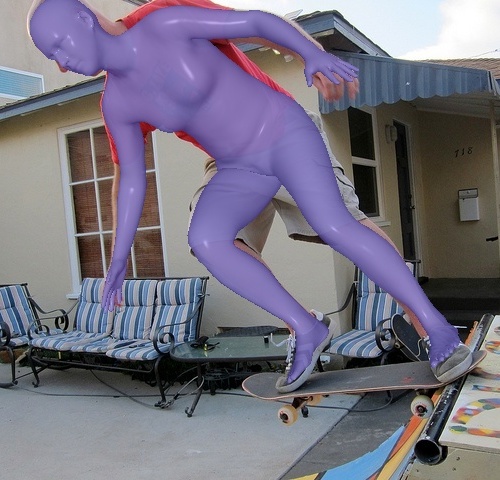}\\
        \rotatebox{90}{~~~~CLIFF \cite{li2022cliff}}\hspace{0.3cm}
        \includegraphics[width=\customimgwidth]{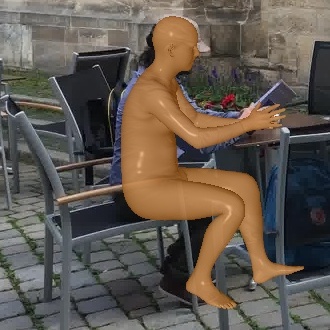}
        \includegraphics[width=\customimgwidth]{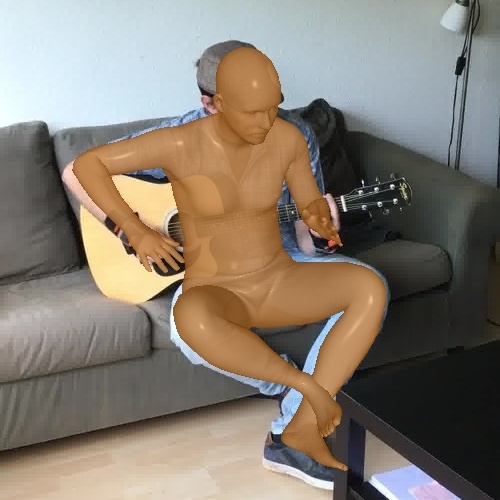}
        \includegraphics[width=\customimgwidth]{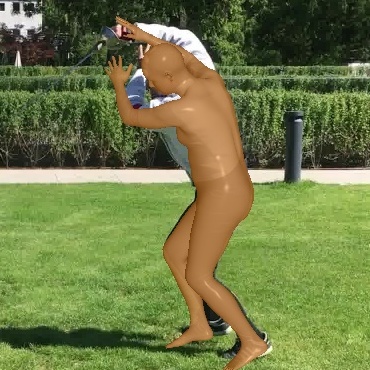}
        \includegraphics[width=\customimgwidth]{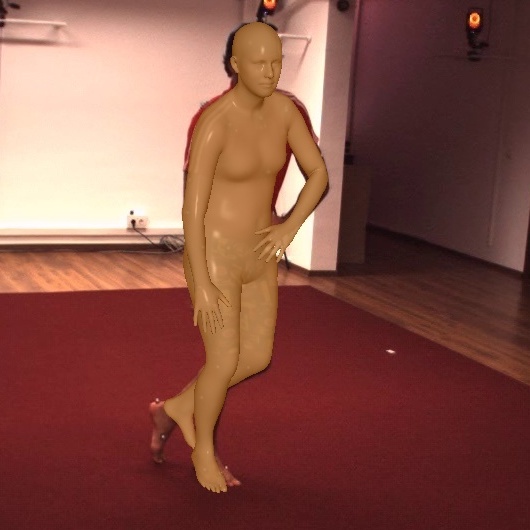}
        \includegraphics[width=\customimgwidth]{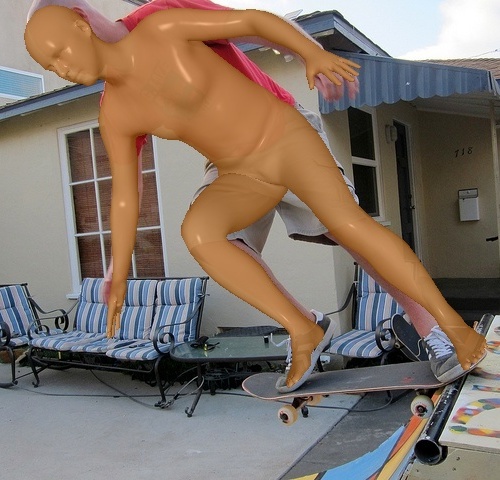}\\
        \rotatebox{90}{~~~~EFT$_{\rm CLIFF}$}\hspace{0.3cm}
        \includegraphics[width=\customimgwidth]{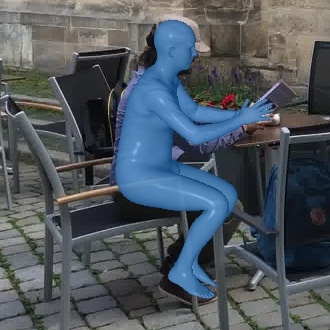}
        \includegraphics[width=\customimgwidth]{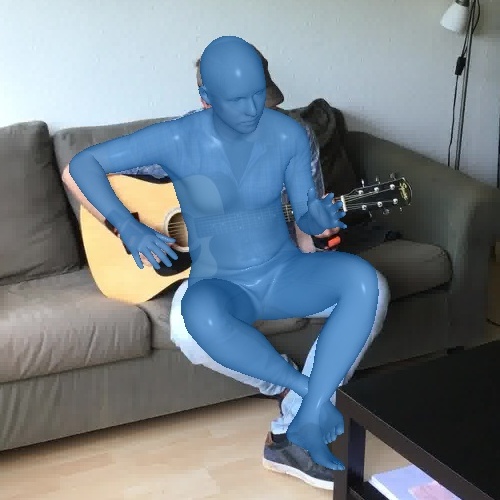}
        \includegraphics[width=\customimgwidth]{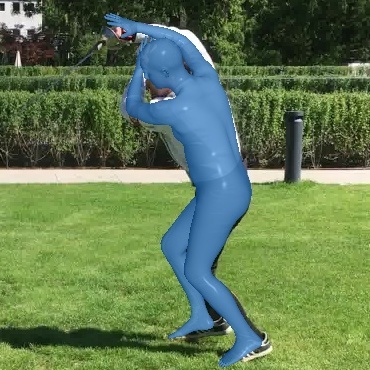}
        \includegraphics[width=\customimgwidth]{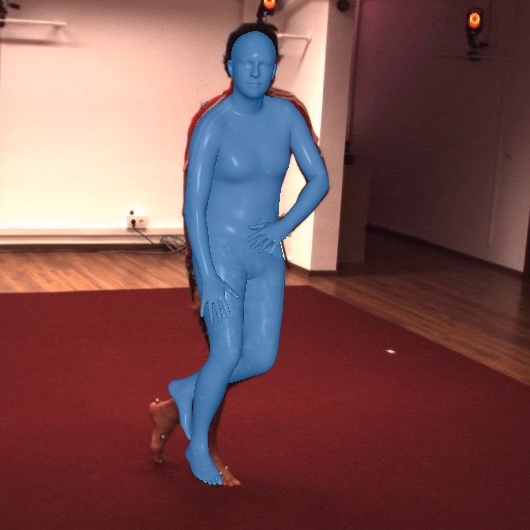}
        \includegraphics[width=\customimgwidth]{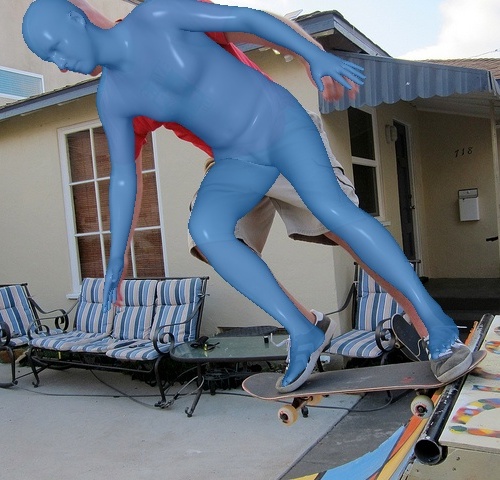}\\
        \rotatebox{90}{~~~~~~~~Ours$\dagger$}\hspace{0.3cm}
        \includegraphics[width=\customimgwidth]{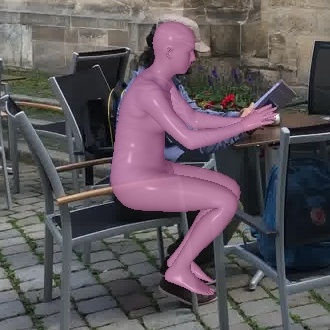}
        \includegraphics[width=\customimgwidth]{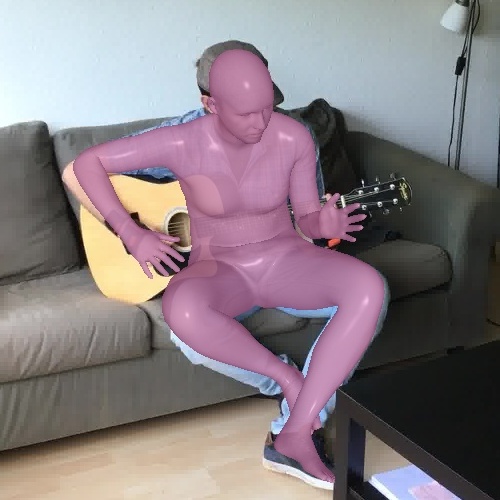}
        \includegraphics[width=\customimgwidth]{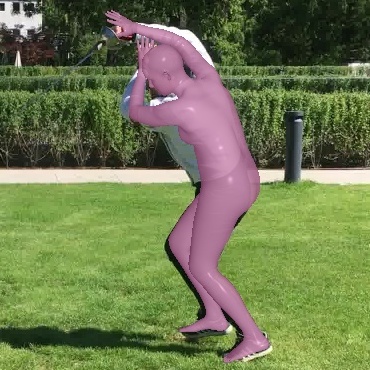}
        \includegraphics[width=\customimgwidth]{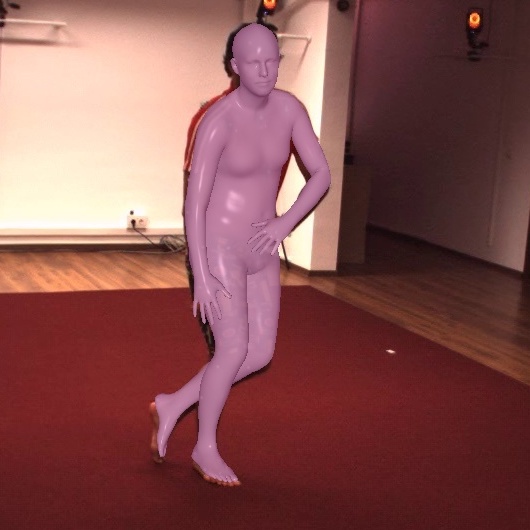}
        \includegraphics[width=\customimgwidth]{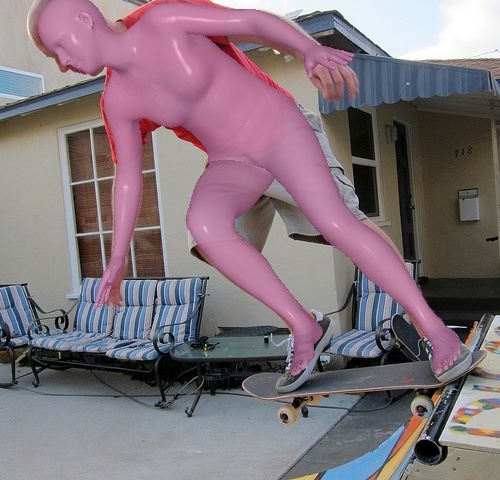}\\
        \rotatebox{90}{~~~~~~~~Ours$\ast$}\hspace{0.3cm}
        \includegraphics[width=\customimgwidth]{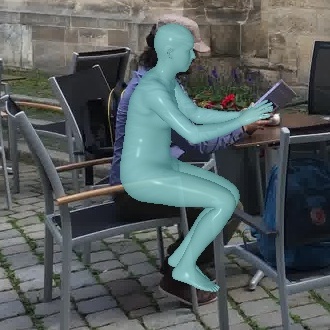}
        \includegraphics[width=\customimgwidth]{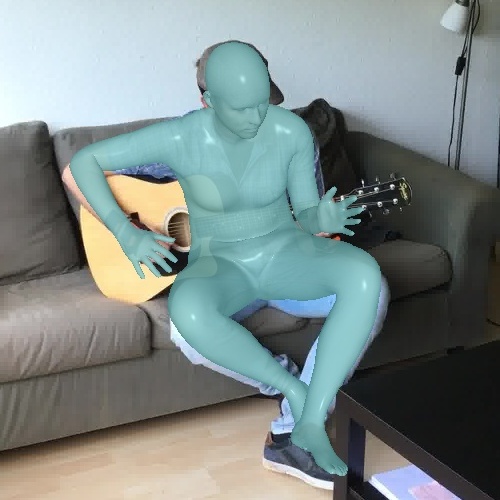}
        \includegraphics[width=\customimgwidth]{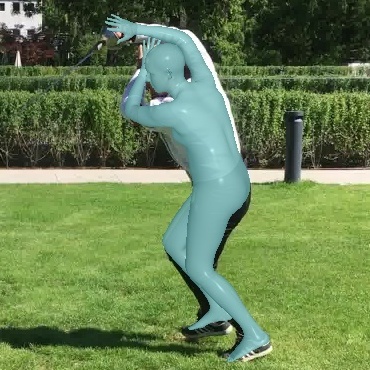}
        \includegraphics[width=\customimgwidth]{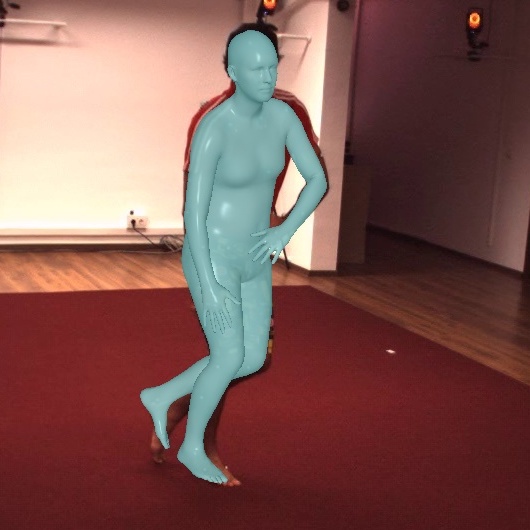}
        \includegraphics[width=\customimgwidth]{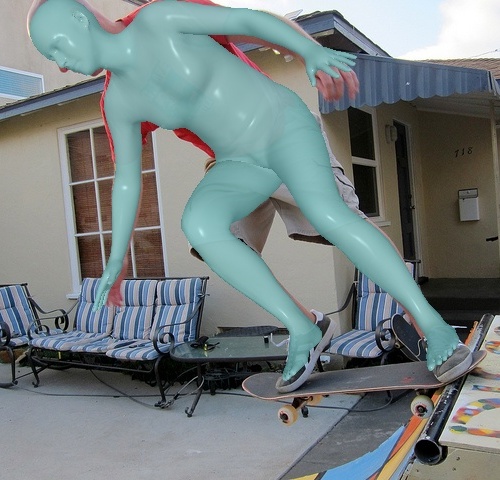}\\
    \end{minipage}
    \caption{\textbf{More qualitative comparisons with SOTA methods.} We show results produced by HybrIK \cite{li2021hybrik}, NIKI \cite{li2023niki}, ProPose \cite{fang2023learning}, ReFit \cite{wang2023refit}, CLIFF \cite{li2022cliff}, EFT$_{\rm CLIFF}$, and our method ({$\dagger$}: OpenPose, $\ast$: RSN ).}
    \label{fig:quantitive1}
\end{figure*}

\begin{figure*}[!t]
    \centering  
    \begin{minipage}{1.0\textwidth}
        \centering
        \rotatebox{90}{~~~~~~~~Input}\hspace{0.3cm}
        \includegraphics[height=2.25cm]{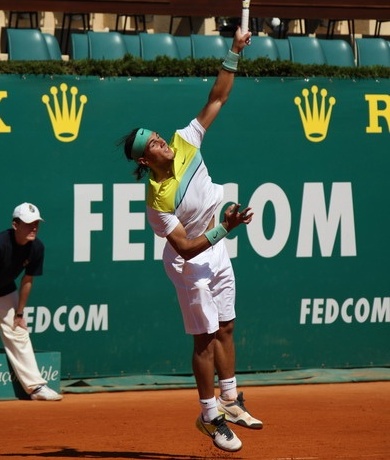}
        \includegraphics[height=2.25cm]{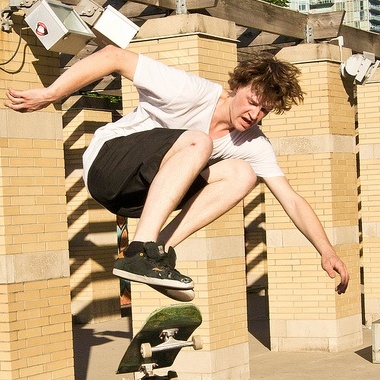}
         \includegraphics[height=2.25cm]{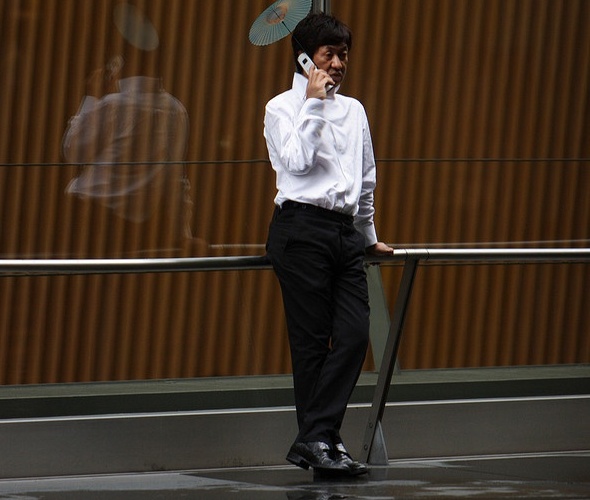}
         \includegraphics[height=2.25cm]{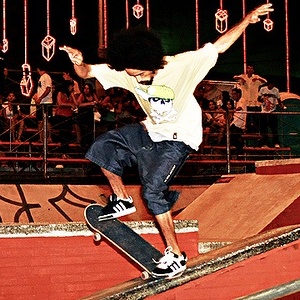}
         \includegraphics[height=2.25cm]{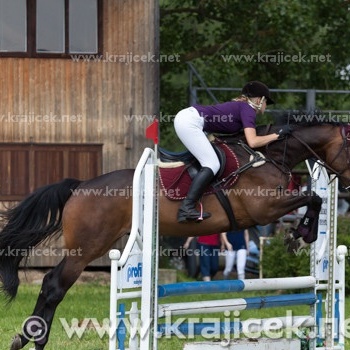}\\
         \rotatebox{90}{~~~HybrIK \cite{li2021hybrik}}\hspace{0.3cm}
        \includegraphics[height=2.25cm]{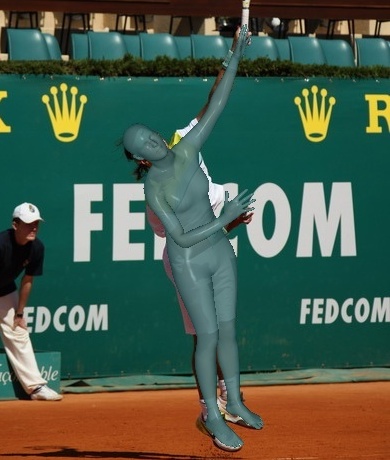}
        \includegraphics[height=2.25cm]{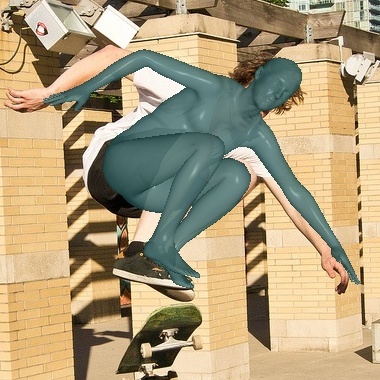}
        \includegraphics[height=2.25cm]{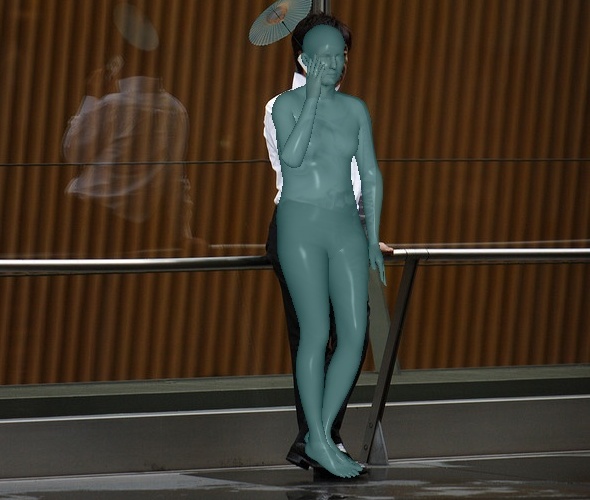}
        \includegraphics[height=2.25cm]{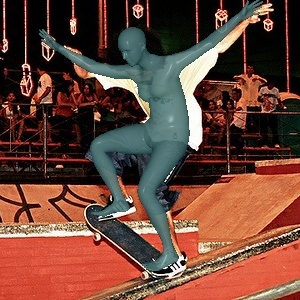}
        \includegraphics[height=2.25cm]{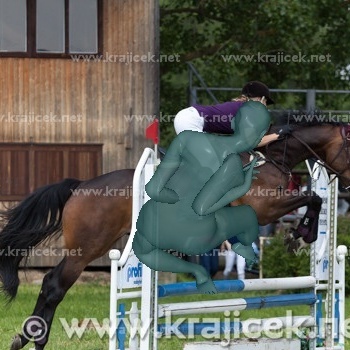}\\
        \rotatebox{90}{~~~~~NIKI \cite{li2023niki}}\hspace{0.3cm}
        \includegraphics[height=2.25cm]{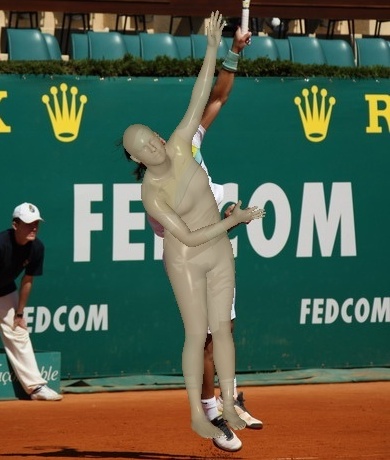}
        \includegraphics[height=2.25cm]{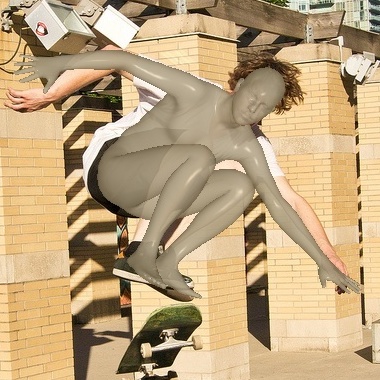}
        \includegraphics[height=2.25cm]{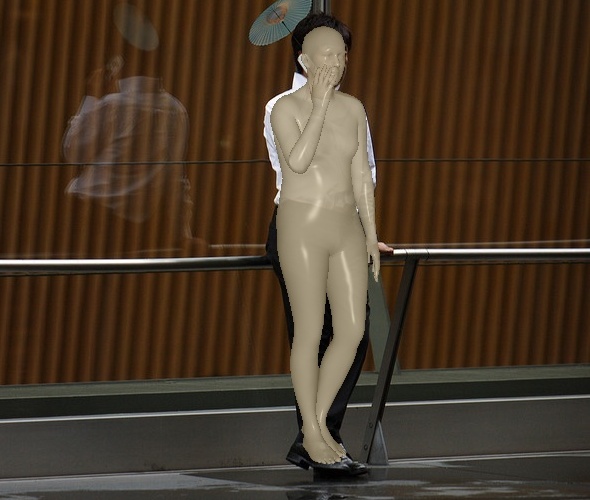}
        \includegraphics[height=2.25cm]{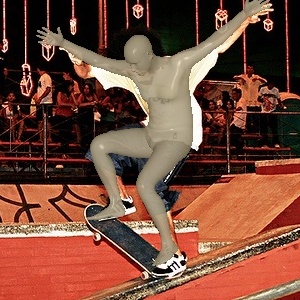}
        \includegraphics[height=2.25cm]{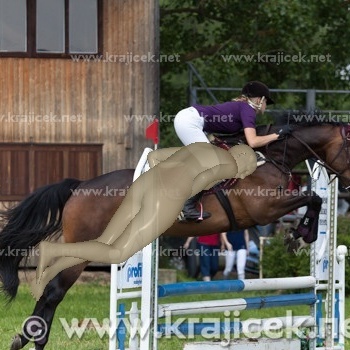}\\
        \rotatebox{90}{~~ProPose \cite{fang2023learning}}\hspace{0.3cm}
        \includegraphics[height=2.25cm]{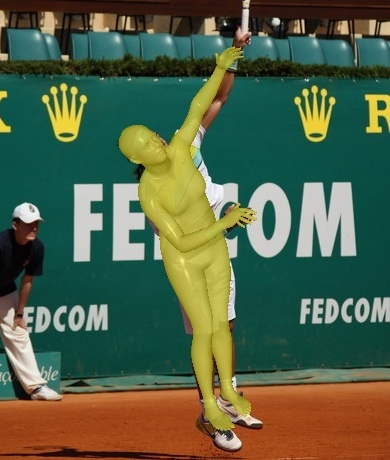}
        \includegraphics[height=2.25cm]{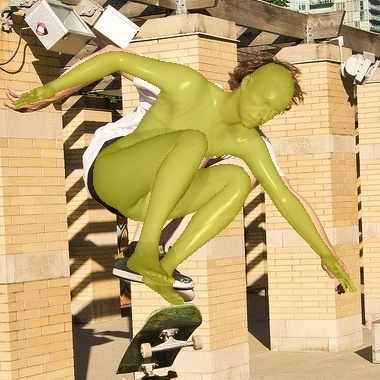}
        \includegraphics[height=2.25cm]{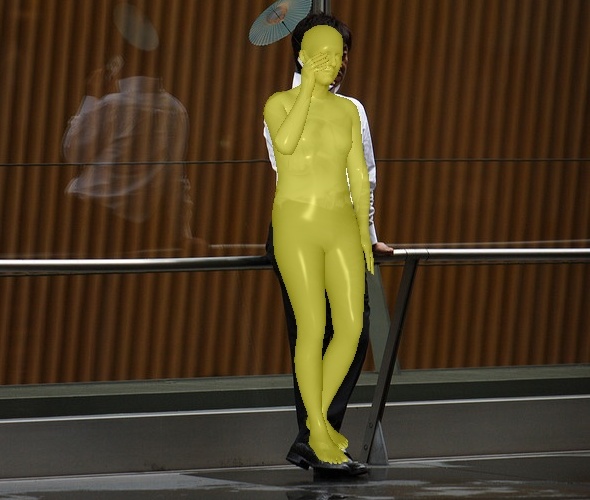}
        \includegraphics[height=2.25cm]{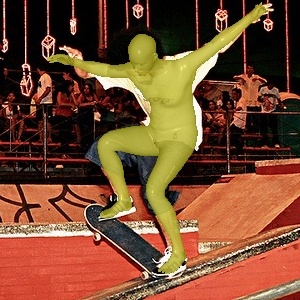}
        \includegraphics[height=2.25cm]{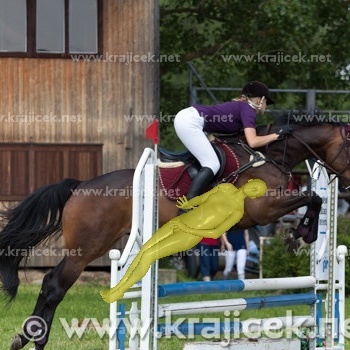}\\
        \rotatebox{90}{~~~~ReFit \cite{wang2023refit}}\hspace{0.3cm}
        \includegraphics[height=2.25cm]{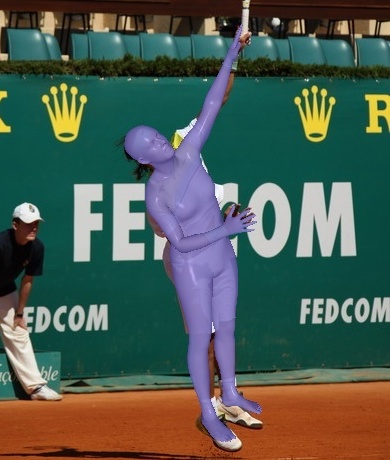}
        \includegraphics[height=2.25cm]{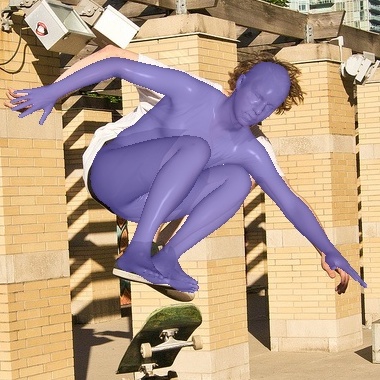}
        \includegraphics[height=2.25cm]{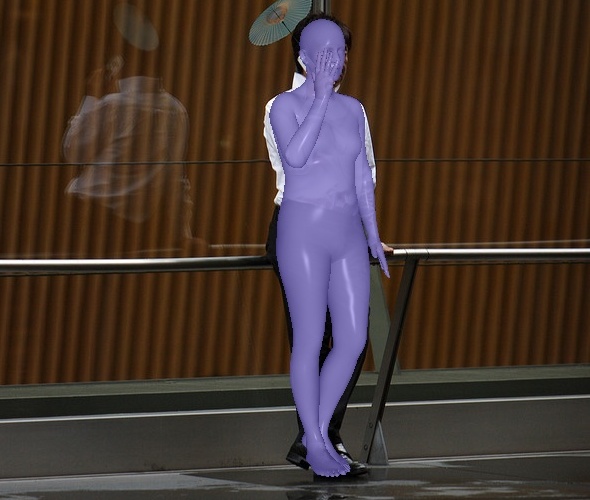}
        \includegraphics[height=2.25cm]{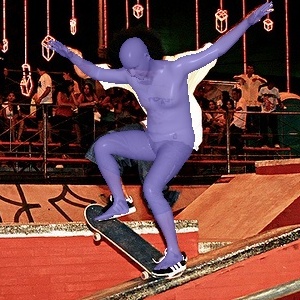}
        \includegraphics[height=2.25cm]{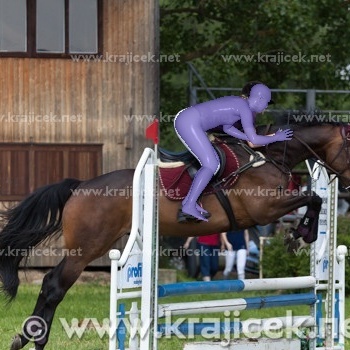}\\
        \rotatebox{90}{~~~~CLIFF \cite{li2022cliff}}\hspace{0.3cm}
        \includegraphics[height=2.25cm]{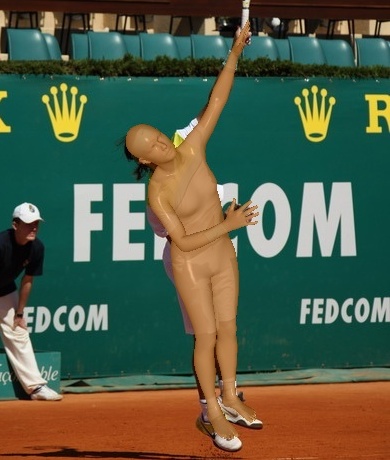}
        \includegraphics[height=2.25cm]{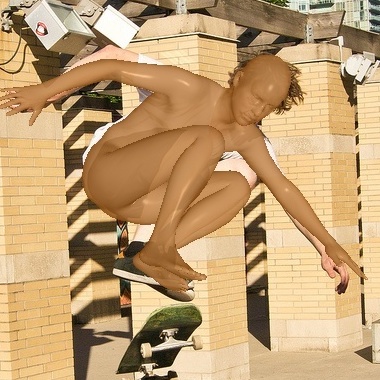}
        \includegraphics[height=2.25cm]{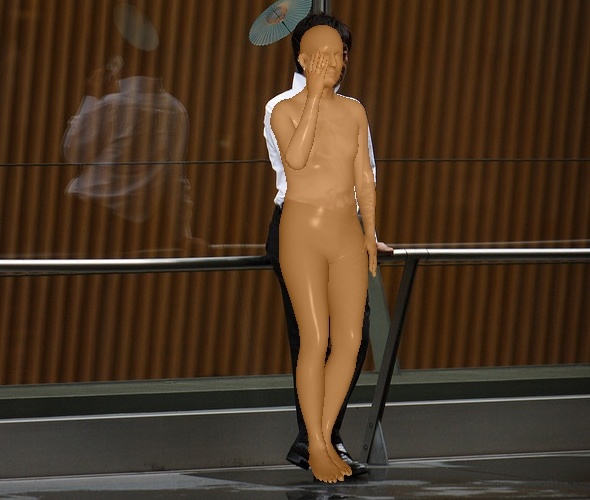}
        \includegraphics[height=2.25cm]{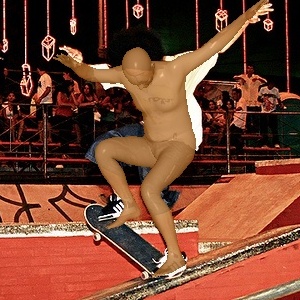}
        \includegraphics[height=2.25cm]{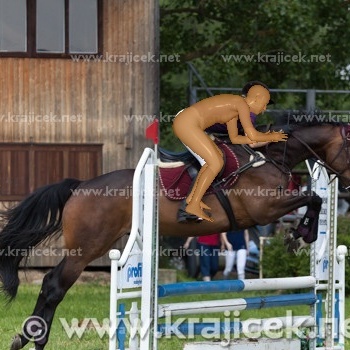}\\
        \rotatebox{90}{~~~~EFT$_{\rm CLIFF}$}\hspace{0.3cm}
        \includegraphics[height=2.25cm]{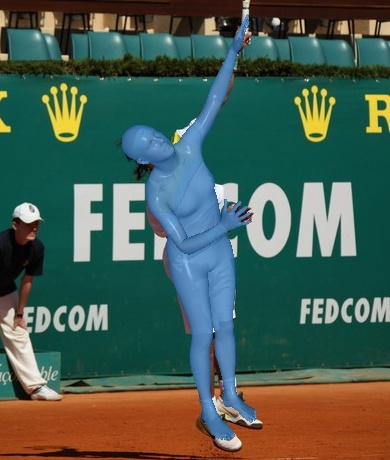}
        \includegraphics[height=2.25cm]{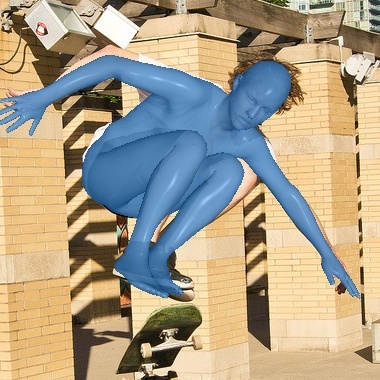}
        \includegraphics[height=2.25cm]{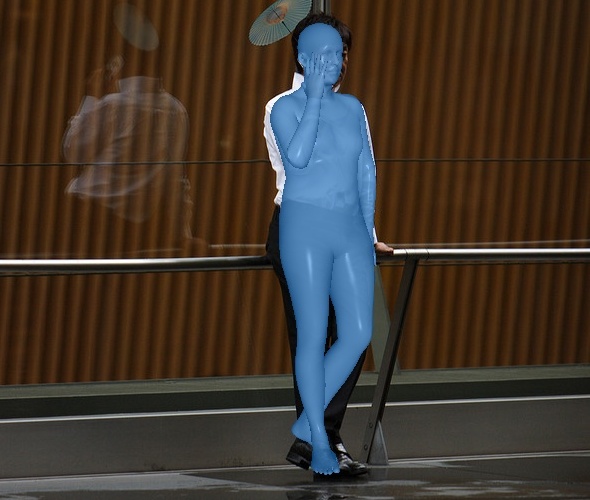}
        \includegraphics[height=2.25cm]{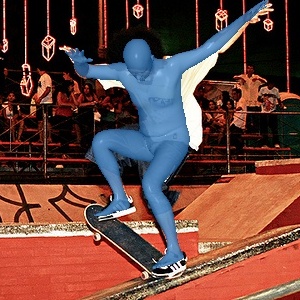}
        \includegraphics[height=2.25cm]{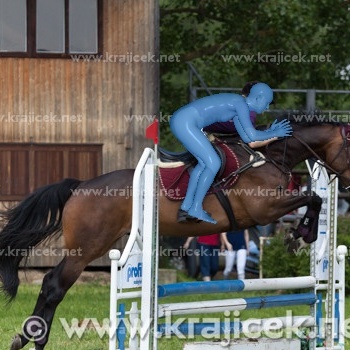}\\
        \rotatebox{90}{~~~~~~~~Ours$\dagger$}\hspace{0.3cm}
        \includegraphics[height=2.25cm]{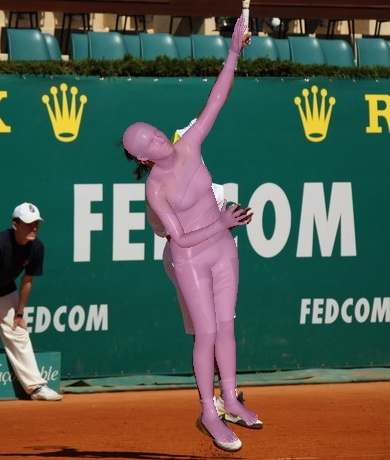}
        \includegraphics[height=2.25cm]{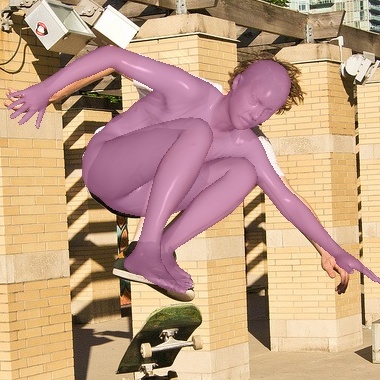}
        \includegraphics[height=2.25cm]{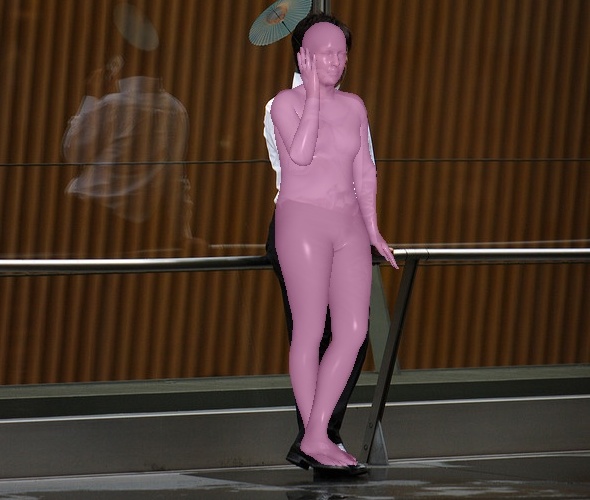}
        \includegraphics[height=2.25cm]{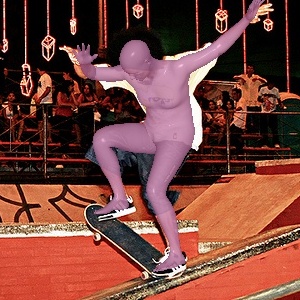}
        \includegraphics[height=2.25cm]{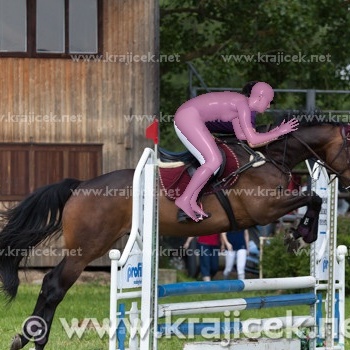}\\
        \rotatebox{90}{~~~~~~~~Ours$\ast$}\hspace{0.3cm}
        \includegraphics[height=2.25cm]{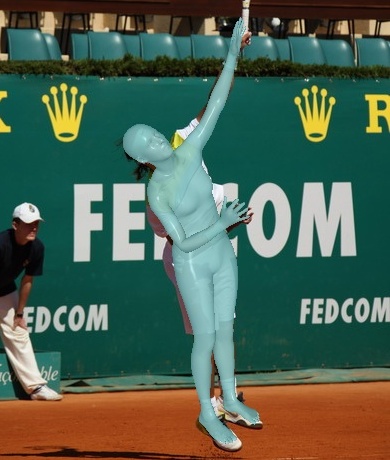}
        \includegraphics[height=2.25cm]{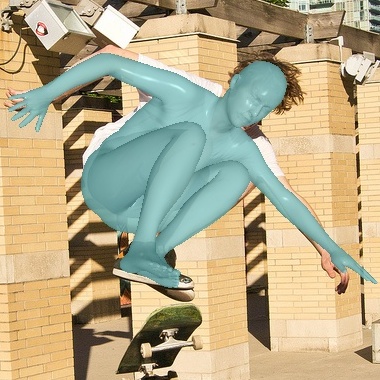}
        \includegraphics[height=2.25cm]{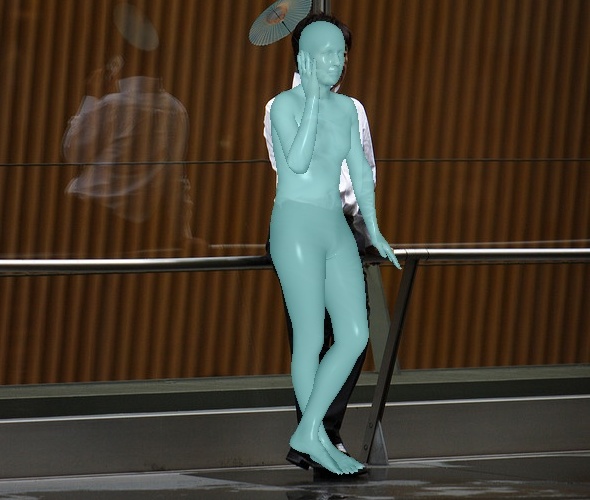}
        \includegraphics[height=2.25cm]{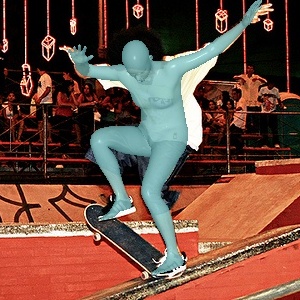}
        \includegraphics[height=2.25cm]{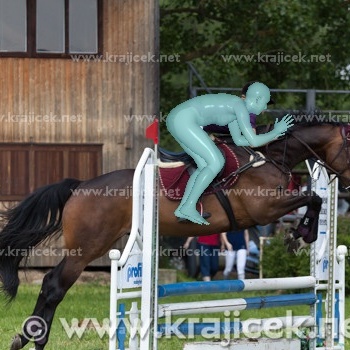}\\ 
    \end{minipage}
    \caption{\textbf{More qualitative comparisons with SOTA methods.} We show results produced by HybrIK \cite{li2021hybrik}, NIKI \cite{li2023niki}, ProPose \cite{fang2023learning}, ReFit \cite{wang2023refit}, CLIFF \cite{li2022cliff}, EFT$_{\rm CLIFF}$, and our method ({$\dagger$}: OpenPose, $\ast$: RSN ).}
    \label{fig:quantitive2}
\end{figure*}

In Figure~\ref{fig:quantitive1} and Figure~\ref{fig:quantitive2}, we show more qualitative comparisons with the latest approaches on test/validation datasets of COCO \cite{lin2014microsoft}, 3DPW \cite{von2018recovering} and Human3.6M \cite{ionescu2013human3}. Besides CLIFF \cite{li2022cliff} and ReFit \cite{wang2023refit} that have been compared with in the main paper, we additionally bring HybrIK \cite{li2021hybrik}, NIKI \cite{li2023niki}, ProPose \cite{fang2023learning} and EFT$_{\rm CLIFF}$ into the comparison. In the shown examples, HybrIK produces misaligned right arm in Figure~\ref{fig:quantitive1} column 2. NIKI produces wrong head orientations (Figure~\ref{fig:quantitive1} column 2 and column 4). For ProPose, ReFit and CLIFF, please pay attention to the feet in Figure~\ref{fig:quantitive1} column 2. EFT$_{\rm CLIFF}$ produces misaligned feet in Figure~\ref{fig:quantitive1} column 4 and Figure~\ref{fig:quantitive2} column 3. Besides, CLIFF produces wrong left leg for the example in Figure~\ref{fig:quantitive1} column 1. For these examples, our method produces visually better results. 

\begin{figure}[h]
    \centering 
    \begin{minipage}{0.8\columnwidth}
        \includegraphics[width=0.32\linewidth]{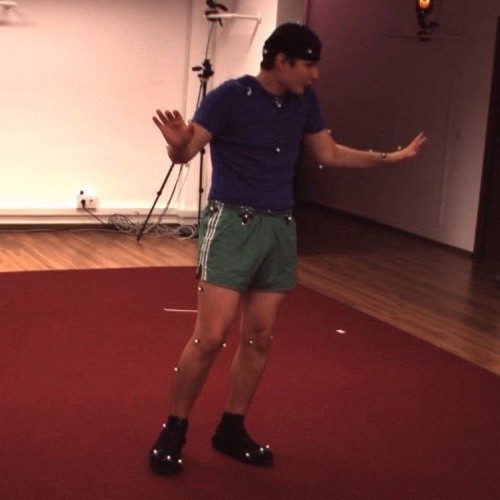}
        \includegraphics[width=0.32\linewidth]{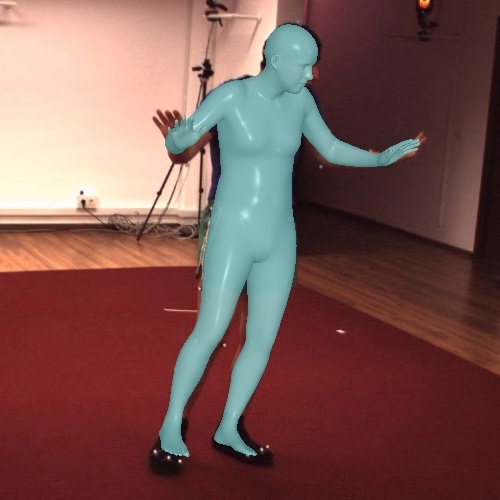}
        \includegraphics[width=0.32\linewidth]{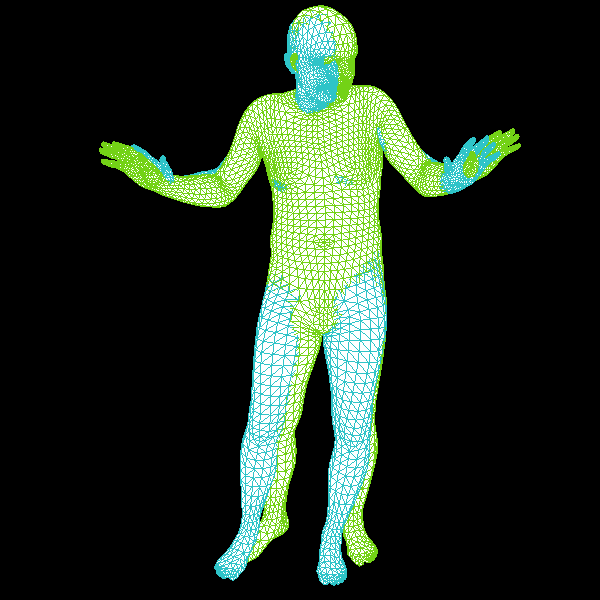}\\
        \includegraphics[width=0.32\linewidth]{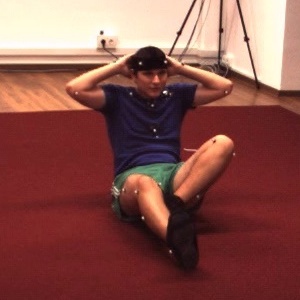}
        \includegraphics[width=0.32\linewidth]{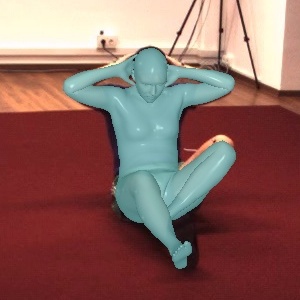}
        \includegraphics[width=0.32\linewidth]{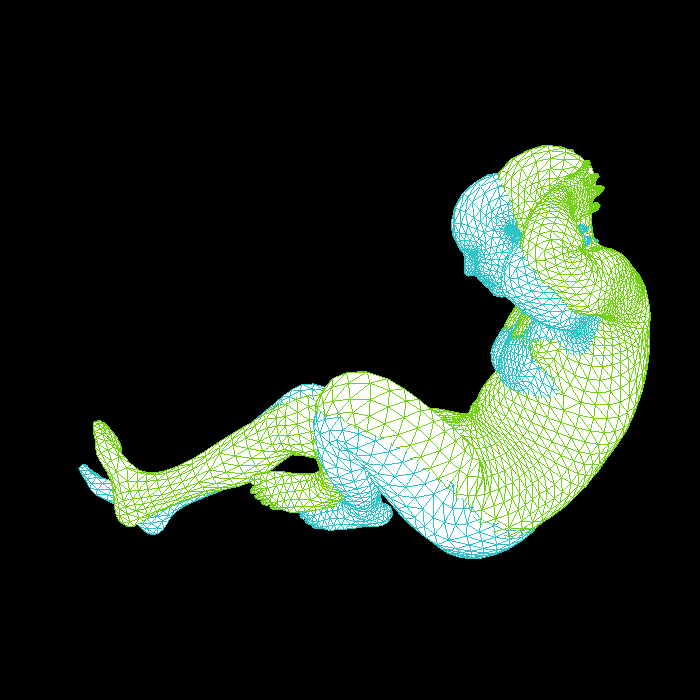}\\
    \end{minipage}
    \centering
    \caption{\textbf{Failure cases.} The projected mesh matches with the target 2D human body exactly, but there is misalignment in the 3D space. Green meshes represent ground-truth.}
    \label{fig:failurecase}
\end{figure}


\begin{figure}[h]
    \centering
    \begin{minipage}{0.45\textwidth}
        \centering
        \includegraphics[width=0.49\textwidth,height=7.05cm]{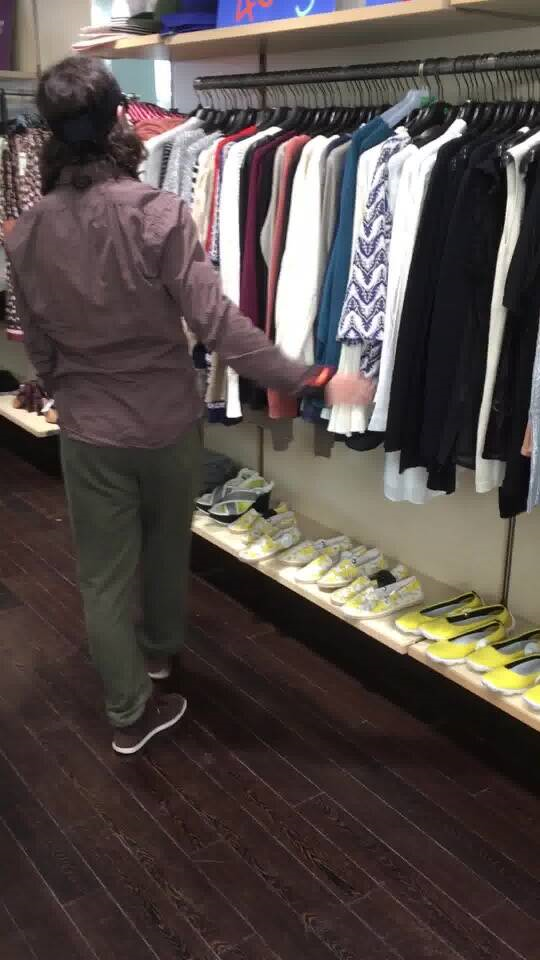}
        \includegraphics[width=0.49\textwidth,height=7.05cm]{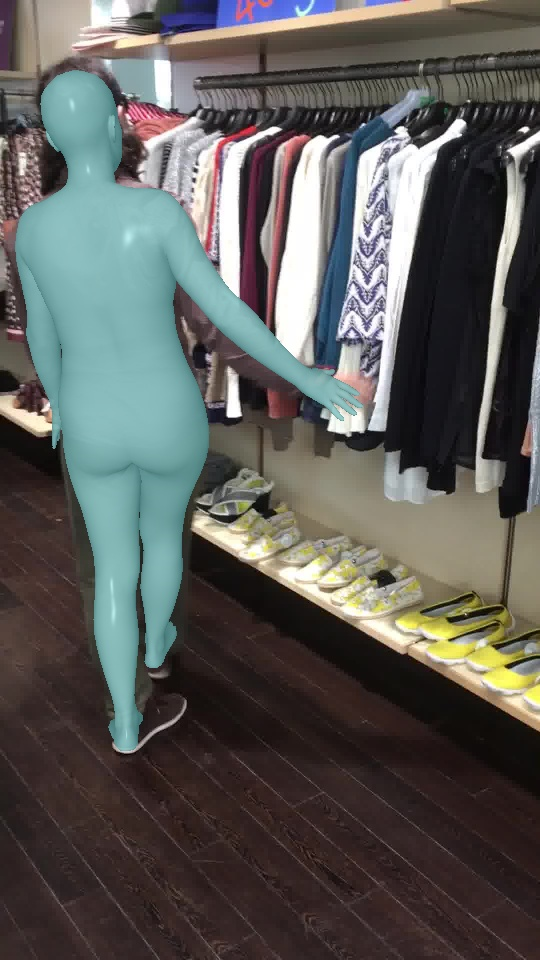}
        \caption{\textbf{Failure case}. The SMPL model's limited flexibility likely causes poor 3D mesh projection in the foot region.}
        \label{fig:failure-case-3}
    \end{minipage}
    \hspace{0.01\textwidth}
    \begin{minipage}{0.45\textwidth}
        \centering
        \includegraphics[width=0.99\textwidth]{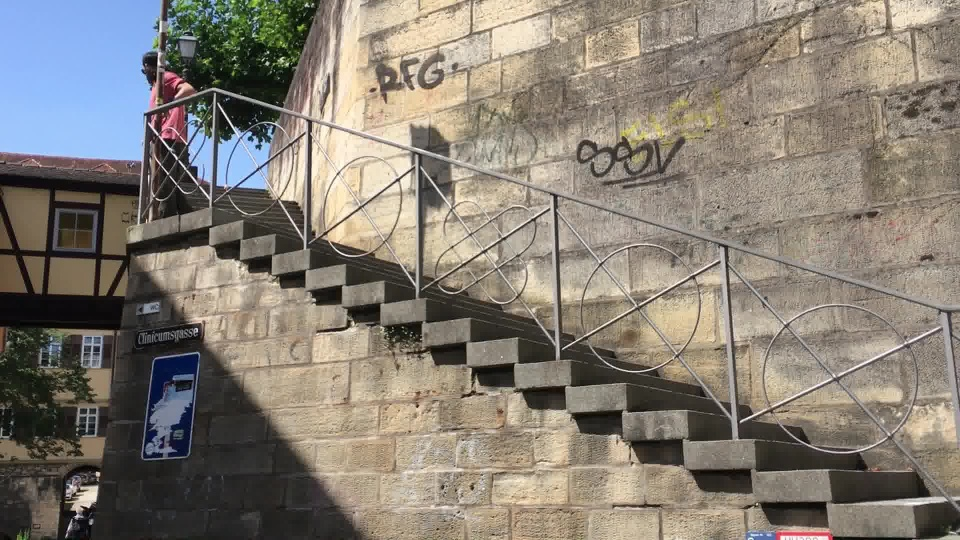} \\
        \includegraphics[width=0.99\textwidth]{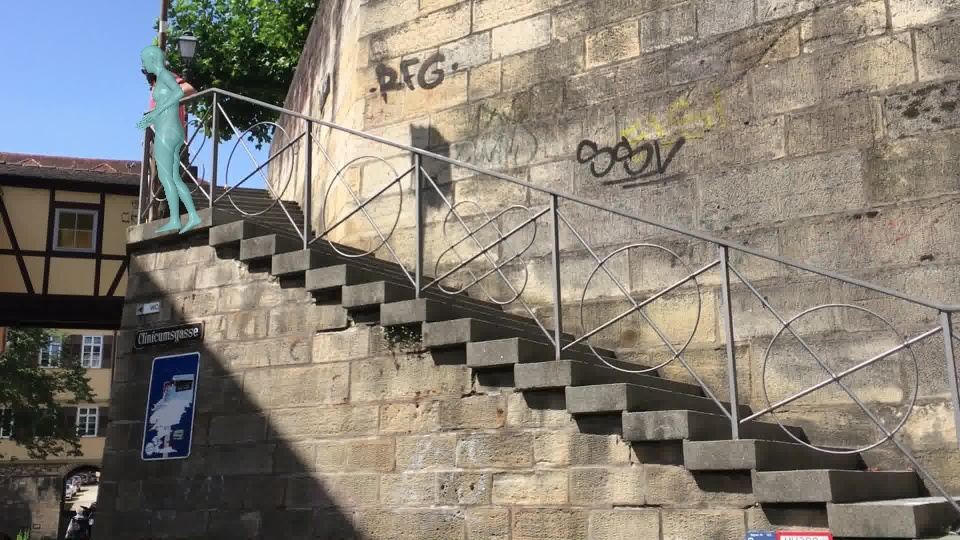}
        \caption{\textbf{Failure case}. Severe occlusion or far camera distance results in a low-quality human mesh estimation.}
        \label{fig:failure-case-4}
    \end{minipage}
\end{figure}

\subsection{Failure Cases}
\label{appendix:Failure Cases}
One kind of failure cases of our method are shown in Figure~\ref{fig:quantitive1} and Figure~\ref{fig:quantitive2}, where there is still slight misalignment between projected mesh and the target 2D evidence, though the misalignment is smaller than that of previous approaches. For example in column 1 of Figure~\ref{fig:quantitive1}, the feet of Our$\dagger$ and Our$\ast$ do not exactly match with the target 2D feet in the image. We observe that the ground-truth meshes in training datasets after projection also show such artifacts. To tackle this problem, one may need to provide more accurate human annotations.

Figure \ref{fig:failurecase} shows another kind of failure cases. In the example of row 1, our reconstructed mesh and the target 2D person are well-aligned. However, from a novel perspective, there is a misalignment between our 3D mesh and the ground truth. This discrepancy arises from the inherent ill-posedness of inferring a 3D mesh from 2D information in a monocular image. In the second row of Figure \ref{fig:failurecase}, we showcase a person with partial occlusion. Please notice the left foot of the person, where the person exhibits self-occlusion. In such a scenario, the accuracy of the corresponding 2D joints is not high, posing a challenge to our method.

Figure \ref{fig:failure-case-3} shows that the 3D mesh projected onto the 2D image performs poorly in the foot region, probably because the SMPL model itself is not flexible enough to capture the large distortion of the two legs.

Due to the severe occlusion or since the distance from the person to the camera is too far in Figure~\ref{fig:failure-case-4}, the quality of the estimated human mesh is poor, not well fitted with the target person.

\end{document}